\documentclass[twoside]{article}

\usepackage[accepted]{aistats2022}
\usepackage[round]{natbib}

\usepackage{amsfonts}
\usepackage{amsmath}
\usepackage{amsthm}
\usepackage{amsbsy}
\usepackage{amssymb} % NOT COMATIBLE WITH svjour3

\usepackage{nicefrac}
\usepackage{mathtools}

\usepackage{xcolor}
\usepackage{color}
\usepackage{graphicx} % Commented out pdftex option since it was clashing with tikz-cd
\usepackage{mdframed} 
\usepackage{thmtools}

\definecolor{shadecolor}{gray}{0.90}
\declaretheoremstyle[
headfont=\normalfont\bfseries,
notefont=\mdseries, notebraces={(}{)},
bodyfont=\normalfont,
postheadspace=0.5em,
spaceabove=1pt,
mdframed={
  skipabove=8pt,
  skipbelow=8pt,
  hidealllines=true,
  backgroundcolor={shadecolor},
  innerleftmargin=4pt,
  innerrightmargin=4pt}
]{shaded}

\declaretheorem[style=shaded,within=section]{definition}
\declaretheorem[style=shaded,sibling=definition]{theorem}
\declaretheorem[style=shaded,sibling=definition]{proposition}
\declaretheorem[style=shaded,sibling=definition]{assumption}
\declaretheorem[style=shaded,sibling=definition]{corollary}

\declaretheorem[style=shaded,sibling=definition]{lemma}

\usepackage[hypertexnames=false,backref=page]{hyperref}
\renewcommand*{\backref}[1]{}
\renewcommand*{\backrefalt}[4]{({%
    \ifcase #1 Not cited.%
          \or Cited on page~#2.%
          \else Cited on pages #2.%
    \fi%
    })}
\usepackage{textcomp}
\usepackage{enumitem}
\usepackage{cases}  % should be loaded after amsmath
\usepackage{silence}
\newcommand{\R}{\mathbb{R}} % Reals
\newcommand{\N}{\mathbb{N}} % Naturals

\newcommand{\cA}{{\cal A}}

\newcommand{\cN}{{\cal N}}
\newcommand{\cO}{{\cal O}}
\newcommand{\cP}{{\cal P}}

\newcommand{\cR}{{\cal R}}
\newcommand{\cS}{{\cal S}}

\newcommand{\mH}{{\bf H}}
\newcommand{\mI}{{\bf I}}

\newcommand{\ones}{{\bf 1}}

\usepackage{tikz}
\newcommand*\circled[1]{\tikz[baseline=(char.base)]{
            \node[shape=circle,draw,inner sep=.5pt] (char) {#1};}}

\usepackage[colorinlistoftodos,bordercolor=orange,backgroundcolor=orange!20,linecolor=orange]{todonotes}

\newcommand{\ruiline}[1]{{#1}}

\newcommand{\eqdef}{\overset{\text{def}}{=}} 
\newcommand{\dotprod}[1]{\left< #1\right>} % product
\newcommand{\norm}[1]{ \left\| #1 \right\|}      % norm 

\newcommand{\Diag}[1]{\mathbf{Diag}\left( #1\right)}

\newcommand{\E}[1]{\mathbb{E}\left[#1\right] } 
\newcommand{\EE}[2]{\mathbb{E}_{#1}\left[#2\right] } 
\newcommand{\colvec}[1]{\begin{bmatrix}#1\end{bmatrix}}

\usepackage{bbm}
\usepackage{nameref}
\usepackage{url}
\usepackage{pifont}% http://ctan.org/pkg/pifont
\usepackage{multirow}% http://ctan.org/pkg/multirow
\usepackage{makecell}
\usepackage[usestackEOL]{stackengine}
\usepackage{booktabs}
\usepackage{tabularx}
\usepackage{hhline}
\usepackage{colortbl}
\usepackage{appendix}
\usepackage{wrapfig}
\usepackage{tikz-cd}
\usepackage{xspace}
\usepackage{algorithm}
\usepackage{algorithmic}
\newcommand\mlnode[1]{\fbox{\begin{tabular}{@{}c@{}}#1\end{tabular}}}

\newcommand{\hnabla}{{\widehat{\nabla}}}

\newcommand{\xmark}{\ding{55}}%

\definecolor{pearThree}{HTML}{E74C3C}
\definecolor{pearcomp}{HTML}{B97E29}
\definecolor{pearDark}{HTML}{2980B9}
\definecolor{pearDarker}{HTML}{1D2DEC}
\hypersetup{
	colorlinks,
	citecolor=pearDark,
	linkcolor=pearThree,
	breaklinks=true,
	urlcolor=pearDarker}

\definecolor{HighlightColor}{gray}{0.97}
\definecolor{aliceblue}{rgb}{0.94, 0.97, 1.0}
\definecolor{palecornflowerblue}{rgb}{0.67, 0.8, 0.94}
\definecolor{paleaqua}{rgb}{0.74, 0.83, 0.9}

\definecolor{linen}{rgb}{0.98, 0.94, 0.9}
\definecolor{magnolia}{rgb}{0.97, 0.96, 1.0}
\definecolor{mistyrose}{rgb}{1.0, 0.89, 0.88}
\definecolor{piggypink}{rgb}{0.99, 0.87, 0.9}
\colorlet{colorast}{red!80!black}

\usepackage{minitoc}
\begin{document}

\runningtitle{A general sample complexity analysis of vanilla policy gradient}

\runningauthor{Rui Yuan , Robert M.~Gower ,  Alessandro Lazaric}

\twocolumn[

\aistatstitle{A general sample complexity analysis of vanilla policy gradient}

\aistatsauthor{ Rui Yuan \And Robert M.~Gower \And  Alessandro Lazaric }

\aistatsaddress{ Meta AI \\ LTCI, T\'el\'ecom Paris \\ Institut Polytechnique de Paris \And  CCM, Flatiron Institute, New York \\ LTCI, T\'el\'ecom Paris \\ Institut Polytechnique de Paris \And Meta AI } ]

%\pagestyle{empty}

% For TOC in appendix (https://tex.stackexchange.com/a/419290)
\doparttoc % Tell to minitoc to generate a toc for the parts
\faketableofcontents % Run a fake tableofcontents command for the partocs

\begin{abstract}
%% Rob; Leave this sort of thing for the intro. Best to get straight to the point!
%Policy gradient (PG) is one of the most popular methods for solving reinforcement learning (RL) problems. However, a solid theoretical understanding of even the ``vanilla'' PG has remained elusive.
We adapt recent tools developed for the analysis of Stochastic Gradient Descent (SGD) in non-convex optimization to obtain convergence and sample complexity guarantees for the vanilla policy gradient (PG). % -- REINFORCE and GPOMDP.
Our only assumptions are that the expected return is smooth w.r.t.\ the policy parameters, that its $H$-step truncated gradient is close to the exact gradient, and a certain \emph{ABC assumption}. This assumption requires the second moment of the estimated gradient to be bounded by $A\geq 0$ times the suboptimality gap,  $B \geq 0$ times the norm of the full batch gradient and an additive constant $C \geq 0$, or any combination of aforementioned. 
%We show that the ABC assumption is more general as compared to the typical bounded gradient assumptions currently in use. 
We show that the ABC assumption is more general than the commonly used assumptions on the policy space to prove convergence to a stationary point. 
%Indeed, all current assumptions used to prove convergence to a stationary point are special cases of our ABC assumption, as we carefully highlight in a hierarchy diagram.
We provide a single convergence theorem that recovers the $\widetilde{\mathcal{O}}(\epsilon^{-4})$ sample complexity of PG to a stationary point. 
Our results also affords greater flexibility in the choice of hyper parameters such as the step size and the batch size $m$, including the single trajectory case (i.e., $m=1$).
When an additional \emph{relaxed weak gradient domination} assumption is available, we establish a novel global optimum convergence theory of PG with $\widetilde{\mathcal{O}}(\epsilon^{-3})$ sample complexity.
 We then instantiate our theorems in different settings, where we both recover existing results and obtain improved sample complexity, e.g., $\widetilde{\mathcal{O}}(\epsilon^{-3})$ sample complexity for the convergence to the global optimum for Fisher-non-degenerated parametrized policies. 
%****
% Not only is the ABC assumption more general, but we also recover the 
%the well known $\widetilde{\mathcal{O}}(\epsilon^{-4})$ sample complexity under the ABC assumption. Thus despite are gains in generality, our convergence analysis remains equally as tight...
%*****
% under smoothness assumption on the objective function and weak conditions on the second moment of the norm of the estimated gradient. When instantiated under common assumptions on the policy space, our general result immediately recovers existing $\widetilde{\mathcal{O}}(\epsilon^{-4})$ sample complexity guarantees, but for wider ranges of parameters (e.g., step size and batch size $m$) with respect to previous literature. Notably, our result includes the single trajectory case (i.e., $m=1$) and it provides
%a more accurate analysis of the dependency on
%problem-specific parameters by fixing previous
%results available in the literature. 
%We believe that the generality of the ABC assumption may provide theoretical guarantees for PG to a much broader range of problems that have not been previously considered.
%We believe that the integration of the state-of-the-art tools from non-convex optimization may lead to identify a much broader range of problems where PG methods enjoy strong theoretical guarantees.
\end{abstract}

%\rui{Abstract: 1. add the ``truncation'' assumption; 2. add the new sample complexity results for Fisher-non-degenerate parametrized policy.}

\section{Introduction}
\label{sec:intro}

\begin{small}
\begin{table*}[t]
	\caption{Overview of different convergence results for vanilla PG methods. The darker cells contain our new results. The light cells contain previously known results that we recover as special cases of our analysis, and extend the permitted parameter settings. White cells contain existing results that we could not recover under our general analysis.
		%The unshaded cells contain results known in the the literature that is not covered in our paper. 
	}
	\label{tab:full picture}
	%\centering
	\begin{small}
	\begin{center}
		\begin{tabular}{|c|c|c|c|m{50mm}|}
			\hline
			\textbf{Guarantee$^{*}$ }  &\textbf{Setting$^{**}$ }  &\Centerstack{\textbf{Reference}\\ (our results in bold)}  &\textbf{Bound }  
			& \multicolumn{1}{c|}{\textbf{Remarks} } \tabularnewline
			\hhline{|=|=|=|=|=|}
			\multirow[b]{3}{*}{\Centerstack{Sample \\ complexity of \\ stochastic PG \\ for FOSP}}
			&\cellcolor{paleaqua}\ref{eq:ABC}         
			&\cellcolor{paleaqua}\Centerstack{\textbf{Thm.~\ref{pro:ABC}} }        
			&\cellcolor{paleaqua}$\widetilde{\cO}(\epsilon^{-4})$         &\cellcolor{paleaqua}Weakest asm.  \tabularnewline
			\hhline{|~|-|-|-|-}
			& \cellcolor{aliceblue} \nameref{E-LS}         
			&\cellcolor{aliceblue} \Centerstack{ \citet{papini2020safe} \\ \textbf{Cor.~\ref{cor:sample_complexity}} }         
			&\cellcolor{aliceblue} $\widetilde{\cO}(\epsilon^{-4})$         & \cellcolor{aliceblue} Weaker asm.; \newline Wider range of parameters; \newline Recover $\cO(\epsilon^{-2})$  for exact PG; \newline Improved smoothness constant   \tabularnewline
			%\hhline{~-|-|-|-}
			%         &\ref{eq:lipschitz_smooth_policy}         &$\widetilde{\cO}(\frac{1}{\epsilon^4})$         
			%&\Centerstack{\textbf{Refs:} \citet{zhang2020global} \\ \textbf{Ours:} Cor.~\ref{cor:sample_complexity} }         
			%&Same remarks as for~\nameref{E-LS}   \\
			\hhline{|=|=|=|=|=|} 
			\multirow[b]{3}{*}{
			\Centerstack{Sample \\ complexity of \\ stochastic PG \\ for GO}
			}
			&\cellcolor{paleaqua}\ref{eq:ABC} + \ref{eq:PL}         
			&\cellcolor{paleaqua}\Centerstack{\textbf{Thm.~\ref{pro:PL}} }         
			&\cellcolor{paleaqua}$\widetilde{\cO}(\epsilon^{-1})$         &\cellcolor{paleaqua}Recover linear convergence  for the exact PG\tabularnewline
			\hhline{|~|-|-|-|-}
			&\cellcolor{paleaqua} \Centerstack{\ref{eq:ABC} + \eqref{eq:weak}}         
			&\cellcolor{paleaqua} \Centerstack{\textbf{Thm.~\ref{pro:weak}} }
			&\cellcolor{paleaqua} $\widetilde{\cO}(\epsilon^{-3})$         &\cellcolor{paleaqua} Recover $\cO(\epsilon^{-1})$ for the exact PG 
			\tabularnewline
			\hhline{|~|-|-|-|-}
			&\cellcolor{paleaqua} \Centerstack{\nameref{E-LS} + \ref{eq:FI} + \\  \ref{eq:compatible}}         
			&\cellcolor{paleaqua} \Centerstack{\textbf{Cor.~\ref{cor:FI}} }
			&\cellcolor{paleaqua} $\widetilde{\cO}(\epsilon^{-3})$         &\cellcolor{paleaqua} Improved by $\epsilon$ compared to Cor.~\ref{cor:sample_complexity} \tabularnewline
			\hhline{|=|=|=|=|=|}
			\multirow{5}{*}{\Centerstack{Sample \\ complexity of \\ stochastic PG \\ for AR}}
			& \cellcolor{paleaqua} \Centerstack{\ref{eq:ABC} + \eqref{eq:weak}}         
			& \cellcolor{paleaqua} \Centerstack{\textbf{Cor.~\ref{cor:regret}} }
			& \cellcolor{paleaqua} $\widetilde{\cO}(\epsilon^{-4})$         &\cellcolor{paleaqua}Weakest asm. \tabularnewline
			\hhline{|~|-|-|-|-}
			&\cellcolor{aliceblue} \Centerstack{\nameref{E-LS} + \ref{eq:FI} + \\ \ref{eq:compatible}}
			&\cellcolor{aliceblue} \Centerstack{\citet{liu2020animproved} \\ \textbf{Cor.~\ref{cor:FI_regret}} }
			&\cellcolor{aliceblue} $\widetilde{\cO}(\epsilon^{-4})$         &\cellcolor{aliceblue} Weaker asm.; \newline Wider range of parameters \tabularnewline
			\hhline{|~|-|-|-|-}
			&\cellcolor{aliceblue} \Centerstack{Softmax + \\ log barrier~\eqref{eq:barrier}}         
			&\cellcolor{aliceblue} \Centerstack{\citet{zhang2020sample} \\ 
				\textbf{Cor.~\ref{cor:regret_log}} }        &\cellcolor{aliceblue} $\widetilde{\cO}(\epsilon^{-6})$         & \cellcolor{aliceblue}Constant step size; \newline Wider range of parameters; \newline Extra phased learning step unnecessary  \tabularnewline
			%\hhline{|=|=|=|=|=|}
			%\multicolumn{1}{c}{ $(\star)$}         &\makecell[l]{Softmax + \\ log barrier~\eqref{eq:barrier}}        &$\widetilde{\cO}(\frac{1}{\delta^2\epsilon^4})$         
			%&Cor.~\ref{cor:log:concentration}         &Recover $\cO(\frac{1}{\epsilon^2})$ for the exact case  \\
			\hhline{|=|=|=|=|=|}
			\multirow{9}{*}{\Centerstack{Iteration \\ complexity of \\ the exact PG \\ for GO}}
			&\cellcolor{aliceblue} \Centerstack{Softmax + \\ log barrier~\eqref{eq:barrier}}         
			&\cellcolor{aliceblue} \Centerstack{\citet{agarwal2021theory} \\ \cellcolor{aliceblue} \textbf{Cor.~\ref{cor:log:concentration}} }         
			&\cellcolor{aliceblue} $\cO(\epsilon^{-2})$         &\cellcolor{aliceblue} Improved by $1-\gamma$  \tabularnewline
			\hhline{|~|-|-|-|-}
			&\cellcolor{aliceblue} Softmax~\eqref{eq:softmax_main}         
			&\cellcolor{aliceblue} \Centerstack{\citet{mei2020ontheglobal} \\ \cellcolor{aliceblue} \textbf{Thm.~\ref{pro:weak}} }         
			&\cellcolor{aliceblue} $\cO(\epsilon^{-1})$         & \cellcolor{aliceblue}  \tabularnewline
			\hhline{|~|-|-|-|-}         
			&\cellcolor{aliceblue} \Centerstack{Softmax + \\ entropy~\eqref{eq:entropy}}         
			&\cellcolor{aliceblue} \Centerstack{ \citet{mei2020ontheglobal} \\  \cellcolor{aliceblue} \textbf{Thm.~\ref{pro:PL}} }         
			&\cellcolor{aliceblue} linear         &\cellcolor{aliceblue}  \tabularnewline
			\hhline{|~|-|-|-|-}
			&\Centerstack{\ref{eq:lipschitz_smooth_policy} + bijection \\ + PPG}         
			& \Centerstack{\citet{zhang2020variational} }  
			& $\cO(\epsilon^{-1})$    & \tabularnewline%Direct parametrized policy satisfies this setting \\
			\hhline{|~|-|-|-|-}
			&Tabular + PPG         
			& \Centerstack{\citet{xiao2022convergence} }         
			& $\cO(\epsilon^{-1})$         & \tabularnewline
			\hhline{|~|-|-|-|-}
			&  LQR         
			& \Centerstack{\citet{fazel2018global} }         
			& linear         & \tabularnewline
			\hline
		\end{tabular} 
	\end{center}
\end{small}
	{$^{*}$ \footnotesize \textbf{Type of convergence.} \textit{PG}: policy gradient; \textit{FOSP}: first-order stationary point; 
	\textit{GO}: global optimum; \textit{AR}: average regret to the global optimum.}\\ %($J^* - \frac{1}{T}\sum_{t=0}^{T-1}\E{J(\theta_t)} \leq \epsilon$).} \\ 
	% $(\star)$ = sample complexity of stochastic PG for high probability convergence guarantee for the global optimum ($\mathbb{P}\left(J^* - J(\theta_t) \geq \epsilon\right) \leq \delta$).} \\
	{$^{**}$ \footnotesize \textbf{Setting.} 
	%\textit{FI}: Asm.~2.1 in~\citet{liu2020animproved} on Fisher information; \textit{compatible}: Asm.~4.4 in~\citet{liu2020animproved} on function approximation error; 
	\textit{bijection}: Asm.1 in~\citet{zhang2020variational} about occupancy distribution; \textit{PPG}: analysis also holds for the projected PG; \textit{Tabular}: direct parametrized policy; \textit{LQR}: linear-quadratic regulator. } %\\
%	{$^{***}$ \footnotesize This column is for additional noteworthy observations from our work compared to others.}
\end{table*}
\end{small}

Policy gradient (PG) is one of the most popular reinforcement learning (RL) methods for computing policies that maximize long-term rewards~\citep{williams1992simple,sutton2000policy,baxter2001infinite}. The success of PG methods is due to their simplicity and versatility, as they can be readily implemented to solve a wide range of problems (including non-Markov and partially-observable environments) and they can be effectively paired with other techniques to obtain more sophisticated algorithms such as the actor-critic~\citep{konda2000actor,volodymyr2016asynchronous}, natural PG~\citep{kakade2001anatural}, natural actor-critic~\citep{peters2008natural,bhatnagar2009natural}, policy mirror descent~\citep{tomar2022mirror,vaswani2022general}, trust-region based variants~\citep{trpo,schulman2017proximal,shani2020adaptive}, and variance-reduced methods~\citep{svrpg,shen2019hessian,xu2020sample,yuan2020stochastic,huang2020momentum,pham2020hybrid,yang2021policy,huang2022bregman}.
%In practice, PG methods are known to be sample-inefficient due to the high variance affecting the gradient estimates and many variants have been proposed to alleviate this issue, e.g., using baselines, variance-reduction techniques, natural gradient variants. Despite its success, 
%
Unlike value-based methods, a solid theoretical understanding of even the ``vanilla'' PG has long been elusive. Recently, a more complete theory of PG has been derived by leveraging the RL structure of the problem together with tools from convex and non-convex optimization (see App.~\ref{sec:related_work} for a thorough review).
%\rui{High level literature reviews here. Also bring Table~\ref{tab:full picture} in the intro.}
%Due to space constraints, we defer a thorough review of recent results to App.~\ref{sec:related_work}.

In this paper,  we first focus 
%\ruiline{our first contribution is} 
on the sample complexity of PG for reaching a FOSP (first-order stationary point). 
We show how PG can be analysed under a very general assumption on the second moment of the estimated gradient called the \emph{ABC} assumption, which includes most of the bounded gradient type assumptions as a special case. Our first contribution is convergence guarantees and sample complexity for both REINFORCE~\citep{williams1992simple} and GPOMDP~\citep{sutton2000policy,baxter2001infinite} under the ABC and assumptions on the smoothness of the expected return and on its truncated gradient. Our sample complexity analysis recovers both the well known $\cO(\epsilon^{-2})$ iteration complexity of exact PG and the $\widetilde{\cO}(\epsilon^{-4})$ sample complexity of REINFORCE and GPOMDP under weaker assumptions than had previously been explored~\citep{zhang2020global,liu2020animproved,xiong2021non-asymptotic}. Furthermore, our analysis is less restrictive when it comes to the hyper-parameter choices. In fact, our results allow for a wide range of step sizes and place almost no restriction on the batch size $m$, even allowing for single trajectory sampling $(m=1)$, which is uncommon in the literature.
The generality of our assumption allows us to unify much of the fragmented results in the literature under one guise. Indeed, we show that the analysis of Lipschitz and smooth policies, Gaussian polices, softmax tabular polices with or without a log barrier or an entropy regularizer are all special cases of our general analysis (see hierarchy diagram further down in Figure~\ref{fig:hierarchy}).

%. To further understand how these different settings are connected, and special cases of the ABC assumption, we provide a hierarchy diagram further down in Figure~\ref{fig:hierarchy}. 

%we apply recent tools developed for the analysis of stochastic gradient descent (SGD) in non-convex optimization~\citep{khaled2020better} to obtain FOSP convergence guarantees for both REINFORCE and GPOMDP under a smoothness assumption on the objective function and a very general assumption on the second moment of the estimated gradient. 
%% Not clear at all. Need to clarify, single theorem under ABC, that ABC includes as special case most all the assumptions used in proving FOSP convergence. This includes many standard assumptions on the policy space, such as .......
%When instantiated under common assumptions on the policy space, our general result  recovers the $\widetilde{\cO}(\epsilon^{-4})$ sample complexity guarantees~\citep{liu2020animproved}, but for wider ranges of parameters (e.g., step size and batch size) with respect to previous literature. Notably, our result includes the single trajectory case (i.e., $m=1$) and it provides a more accurate analysis of the dependency on
%problem-specific parameters by getting tighter bounds of the smoothness constant and the variance of the gradient estimator.
%%fixing previous results available in the literature. 

\ruiline{Recently, there has also been much work on establishing the convergence of PG to a global optimum (i.e., the best-in-class policy). This usually requires more restrictive assumptions~\citep{zhang2020variational,zhang2021convergence}, specific RL settings (e.g., linear-quadratic regulator~\citep{fazel2018global}, tabular~\citep{agarwal2021theory} and softmax tabular policy~\citep{mei2020ontheglobal}), and it is often limited to exact PG. Inspired by the sample complexity analysis of the stochastic PG for the global optimum in~\citet{liu2020animproved} and~\citet{ding2021global}, our second contribution is to establish a novel global optimum convergence theory of PG when an additional \emph{relaxed weak gradient domination} assumption is available.
Our sample complexity analysis recovers the well known $\cO(\epsilon^{-1})$ iteration complexity of the exact PG with the softmax tabular policy~\citep{mei2020ontheglobal} as a special case and obtains a new improved $\widetilde{\cO}(\epsilon^{-3})$ sample complexity compared to $\widetilde{\cO}(\epsilon^{-4})$ in~\citet{liu2020animproved}, with the Fisher-non-degenerate parametrized policy~\citep{liu2020animproved,ding2021global} as a special case. 
We also establish even faster global optimum convergence theory when replacing the relaxed weak gradient domination assumption by gradient domination in App.~\ref{sec:PL}. As a special case, we recover the well known linear convergence rate of the exact PG with the softmax tabular policy with entropy regularization~\citep{mei2019onprincipled} in App.~\ref{sec:PL}.}
 Table~\ref{tab:full picture} provides a complete overview of our results.%, including cases where our general analysis would not apply.

%\rui{Remove the following paragraph ? This is already mentioned in Section~\ref{sec:discussion}.}
%We believe that the generality of the ABC assumption may provide theoretical guarantees for PG for a broader range of problems that have not been previously considered, and help unify our current understanding of PG and many assumptions currently in use.

%We believe that the integration of state-of-the-art tools from non-convex optimization may lead to identify a much broader range of problems where PG methods enjoy strong theoretical guarantees.

%Here again we match the best known sample complexity of ?? given by ??, but under the much more general ABC assumption.
%\rob{Fill in details please Rui?}
%and obtain the improved sample complexity 
%
% also establish a 
%
%However, our convergence theory is so general that it only implies the FOSP convergence without further assumptions. Notice that~
%
% establish global optimum convergence theories under different RL settings and different assumptions Additionally, when the (weak) gradient domination assumption is available for a specific RL problem, we also establish the global optimum convergence theory and obtain the improved sample complexity in App.~\ref{sec:PL} compared to Thm.~\ref{pro:ABC}.

% !TEX root = summary.tex

\section{Preliminaries}

\textbf{Markov decision process (MDP).} 
We consider a MDP $ M=\{\cS, \cA, \cP, \cR, \gamma, \rho\}$, where $\cS$ is a state space; $\cA$ is an action space; $\cP$ is a Markovian transition model, where $\cP(s' \mid s, a)$ is  the transition density from state $s$ to $s'$ under action $a$; $\cR$ is the reward function, where $\cR(s, a) \in [-\cR_{\max}, \cR_{\max}]$ is the bounded reward for state-action pair $(s, a)$
% $\cR(s, a) \eqdef \EE{s'\sim\cP(\cdot\mid s, a)}{\cR(s, a, s')} \in [-\cR_{\max}, \cR_{\max}]$ 
; $\gamma \in [0, 1)$ is the discounted factor; and $\rho$ is the initial state distribution. The agent's behaviour is modelled as a policy $\pi \in \Delta(\cA)^\cS$, where $\pi(a\mid s)$ is the density of the distribution over actions at state $s\in\cS$.
We consider the infinite-horizon discounted setting. 
%However, we can never execute an infinite-horizon trajectory for estimation in experiments, we approximate it by truncated trajectories with a fixed length $H$.
%We consider episodic MDPs with effective horizon $H$. This is different to the finite-horizon setting where the optimal policy is non-stationary. We can thus limit the attention to trajectories of length $H$. 
%A trajectory $\tau$ is a sequence of states and actions $(s_0, a_0, s_1, a_1, \cdots, s_{H-1}, a_{H-1})$ observed by following a stationary policy, where $s_0 \sim \rho$.

%Let $p(\tau \mid \pi)$ be the distribution induced by the policy $\pi$ on the set $\cT$ of all possible trajectories,  that is
%% Rob: Removed on time notation $\cT$ and re-written the above commented out
Let $p(\tau \mid \pi)$ be the probability density of a single trajectory $\tau$ being sampled from $\pi$,  that is
\begin{equation} \label{eq:p}
p(\tau\mid\pi) = \rho(s_0)\prod_{t=0}^\infty\pi(a_t\mid s_t)\cP(s_{t+1}\mid s_t,a_t).
\end{equation}
With a slight abuse of notation, let $\cR(\tau) = \sum_{t=0}^\infty\gamma^t\cR(s_t, a_t)$ be the total discounted reward accumulated along trajectory $\tau$. We define the expected return of $\pi$ as
\vspace{-.1cm}
\begin{align} \label{eq:J}
J(\pi) \eqdef \EE{\tau \sim p(\cdot\mid\pi)}{\cR(\tau)}.
\end{align}
%\rob{ I removed $= \EE{\tau \sim p(\cdot\mid\pi, \cM)}{\cR(\tau)} $  because $p(\cdot\mid\pi, \cM)$ is not even defined. }
%which is also called the performance function. 
%To compute the policy maximizing $J(\pi)$, one can use policy gradient methods.

\textbf{Policy gradient.} 
% designed to compute the policy maximizing the total reward $J(\pi)$ by gradient ascent.
%% \rob{I removed the ``one time'' notation $\Pi_{\theta}$}
  We introduce a set of parametrized policies $ \{\pi_{\theta} : \theta \in \mathbb{R}^d\}$, with the assumption that $\pi_{\theta}$ is  differentiable w.r.t.\ $\theta$.    We denote $J(\theta) = J(\pi_{\theta})$ and $p(\tau\mid\theta) = p_{\theta}(\tau) = p(\tau\mid\pi_{\theta})$.  In  general, $J(\theta)$ is a non-convex function.
  The PG methods  use gradient ascent in the space of $\theta$
to find the policy that maximizes the expected return, i.e., $\theta^* \in \arg\sup_{\theta \in \R^d}  J(\theta)$. We denote the \emph{optimal expected return} as $J^* \eqdef J(\theta^*).$
%  For simplicity, we consider $\Theta \subseteq \mathbb{R}^d$. 
%We also define $J^* = \sup_{\theta \in \R^d} J(\theta)$ the optimal expected total reward and the parameter of the optimal policy.
 % of the $\theta$ parameter.

The gradient $\nabla J(\theta)$ of the expected return has the following structure 
\begin{align} \label{eq:GD}
\nabla J(\theta) &= \int\cR(\tau)\nabla p(\tau\mid\theta)d\tau \\ 
&= \int\cR(\tau) \left(\nabla p(\tau\mid\theta) / p(\tau\mid\theta)\right) p(\tau\mid\theta)d\tau \nonumber \\
&= \EE{\tau\sim p(\cdot\mid\theta)}{\cR(\tau)\nabla\log p(\tau\mid\theta)} \nonumber \\ 
&\overset{\eqref{eq:p}}{=} \EE{\tau}{\sum_{t=0}^\infty\gamma^t\cR(s_t, a_t)\sum_{t'=0}^\infty\nabla_{\theta}\log\pi_{\theta}(a_{t'} \mid s_{t'})}.  \nonumber
\end{align}
In practice, we cannot compute this full gradient, since computing the above expectation requires averaging over all possible trajectories $\tau\sim p(\cdot\mid\theta)$. We  
%Since it is not possible to execute all possible trajectories up to infinity to compute the full gradient $\nabla J(\theta)$, one has to
 resort to an empirical estimate of the gradient by sampling $m$ truncated trajectories $\tau_i = \left(s_0^i, a_0^i, r_0^i, s_1^i, \cdots, s_{H-1}^i, a_{H-1}^i, r_{H-1}^i\right)$ with $r_t^i = \cR(s_t^i, a_t^i)$ obtained by executing $\pi_{\theta}$ for a given fixed horizon $H \in \N$. The resulting  gradient estimator is 
\begin{align} \label{eq:REINFORCE}
& \hnabla_m J(\theta) = \nonumber \\
& \; \frac{1}{m}\sum_{i=1}^m\sum_{t=0}^{H-1}\gamma^t\cR(s_t^i, a_t^i)\cdot \sum_{t'=0}^{H-1}\nabla_{\theta}\log\pi_{\theta}(a_{t'}^i \mid s_{t'}^i).
\end{align}
The estimator~\eqref{eq:REINFORCE} is known as the REINFORCE gradient estimator~\citep{williams1992simple}.

The REINFORCE estimator can be simplified by leveraging the fact that future actions do not depend on past rewards. This leads to the alternative formulation of the full gradient
\begin{align} 
& \nabla J(\theta) = \nonumber \\
&\; \EE{\tau}{\sum_{t=0}^\infty\left(\sum_{k=0}^t\nabla_{\theta}\log\pi_{\theta}(a_k \mid s_k)\right)\gamma^t\cR(s_t, a_t)}, \label{eq:GD2*}
\end{align}
%
%From~\eqref{eq:GD2}, one can suggest the gradient estimator $\hnabla_m J(\theta)$ as
%\begin{align} \label{eq:PGT}
%\hnabla_m J(\theta) &= \frac{1}{m}\sum_{i=1}^m\sum_{t=0}^{H-1}\nabla_{\theta}\log\pi_{\theta}(a_t^i \mid s_t^i) \nonumber \\ 
%&\quad \cdot\sum_{t'=t}^{H-1}\gamma^{t'}\cR(s_{t'}^i,a_{t'}^i),
%\end{align}
%known as policy gradient theorem (PGT)~\citep{sutton2000policy}.
which leads to the following estimate of the gradient known as  GPOMDP~\citep{baxter2001infinite}
\begin{align} \label{eq:GPOMDP}
& \hnabla_m J(\theta) =  \nonumber \\ 
&\; \frac{1}{m}\sum_{i=1}^m\sum_{t=0}^{H-1} \left(\sum_{k=0}^t\nabla_{\theta}\log\pi_{\theta}(a_k^i \mid s_k^i)\right)\gamma^t\cR(s_t^i, a_t^i).
\end{align}
%which is known as GPOMDP~\citep{baxter2001infinite}.
%It has been shown in~\citep{peters2008reinforcement} that PGT~\eqref{eq:PGT} is equivalent to PGOMDP~\eqref{eq:GPOMDP}. Due to their equivalence, we refer to them interchangeably.
A derivation of~\eqref{eq:GD2*} is provided in Appendix~\ref{sec:auxiliary} (Lemma~\ref{lem:vanilla_PG_derivation}) for completeness.

Both REINFORCE and GPOMDP are the truncated versions of unbiased gradient estimators and they are unbiased estimates of the gradient of the truncated expected return
$J_H(\theta) \eqdef  \EE{\tau}{\sum_{t=0}^{H-1}\gamma^t\cR(s_t, a_t)}$
\footnote{We allow $H$ to be infinity so that $J_{\infty}(\cdot) = J(\cdot)$. }. 
%\begin{eqnarray*}
%J_H(\theta) &\eqdef& \EE{\tau}{\sum_{t=0}^{H-1}\gamma^t\cR(s_t, a_t)}.
%\end{eqnarray*}

Equipped with gradient estimators, vanilla policy gradient  updates the policy parameters as follows
\begin{eqnarray} \label{eq:GA}
\theta_{t+1} \; = \; \theta_t + \eta_t\hnabla_m J(\theta_t)
\end{eqnarray}
where $\eta_t > 0$ is the step size  at the $t$-th iteration.% (see also Algorithm~\ref{alg:pg} in App.~\ref{sec:related_work}).

%All the three gradient estimators mentioned above are unbiased~\citep{peters2008reinforcement}.
%We refer to  them as vanilla policy gradient methods. In the rest of the paper, we provide a better convergence theory of them.

% !TEX root = summary.tex

\section{Non-convex optimization under ABC assumption} \label{sec:ABC}

\subsection{First-order stationary point convergence} \label{sec:fosp}

We use $\hnabla_m J(\theta)$ to denote 
the unbiased policy gradient estimator of $\nabla J_H(\theta)$ used in~\eqref{eq:GA}. It can be the exact gradient $\nabla J(\theta)$ when $H=m = \infty$,  
or the truncated gradient estimators in~\eqref{eq:REINFORCE} or~\eqref{eq:GPOMDP}. All our forthcoming analysis relies on the following common assumptions.%: the smoothness  and truncation assumption.
\begin{assumption}[Smoothness] \label{ass:smooth}
There exists $L>0$ such that, for all $\theta, \theta' \in \R^d$, we have
\begin{eqnarray}
\left|J(\theta') - J(\theta) - \dotprod{\nabla J(\theta), \theta'-\theta}\right| \leq \frac{L}{2}\norm{\theta'-\theta}^2.
\end{eqnarray}
\end{assumption}
%In order to apply this result to our case, we need an additional assumption to bound the error due to the truncation of the horizon as follows.
\begin{assumption}[Truncation] \label{ass:trunc}
There exists $D, D' > 0$ such that, for all $\theta\in\R^d$, we have
\begin{align}
\left|\dotprod{\nabla J_H(\theta), \nabla J_H(\theta) - \nabla J(\theta)}\right| \; &\leq \; D\gamma^H, \label{eq:trunc} \\
\norm{\nabla J_H(\theta) - \nabla J(\theta)} \; &\leq \; D'\gamma^H. \label{eq:trunc2}
\end{align}
\end{assumption}
%% Rob: Consider removing this whole paragraph? Not very clear, and not really useful.
%While we need~\eqref{eq:trunc} and~\eqref{eq:trunc2} to hold as an assumption, we notice that they are reasonable since we have
%\begin{eqnarray}
%\left|J(\theta) - J_H(\theta)\right| &\overset{\eqref{eq:J}+\eqref{eq:J_H}}{=}& \E{\sum_{t=H}^\infty\gamma^t\cR(s_t,a_t)} \nonumber \\
%&\leq& \frac{\cR_{\max}}{1-\gamma}\gamma^H.
We recall that given the boundedness of the reward function, we have 
$\left|J(\theta) - J_H(\theta)\right| \leq \frac{\cR_{\max}}{1-\gamma}\gamma^H$
by the definition of $J(\cdot)$ and $J_H(\cdot)$. As such, when $H$ is large, the difference between $J(\theta)$ and $J_H(\theta)$ is negligible. However, Asm.~\ref{ass:trunc} is still necessary, since in our analysis we first prove that $\norm{\nabla J_H(\theta)}^2$ is small, and then rely on~\eqref{eq:trunc2} to show that  $\norm{\nabla J(\theta)}^2$ is also small. 

% are built on the first-order stationary point (FOSP). Once we find a stationary point $\theta$ such that $\norm{\nabla J_H(\theta)}$ is close to $0$, we need~\eqref{eq:trunc2} to claim the FOSP of $J(\theta)$.
We also make use of the recent\footnote{After publishing this paper, thanks to Francesco Orabona who pointed out that this ABC assumption already appeared in 1973, see just above equation (3.1) in~\citet{polyakABC1973}, where $M$ is used to denote expectation.} \textit{ABC} assumption~\citep{khaled2020better}\footnote{While \citet{khaled2020better} refer to this assumption as \textit{expected smoothness}, we prefer the alternative name ABC to avoid confusion with the smoothness of $J$.}  which bounds the second moment of the norm of the gradient estimators using the norm of the truncated full gradient, the suboptimality gap and an additive constant. 
\begin{assumption}[ABC] \label{ass:ABC}
There exists $A,B, C \geq 0$ such that the  policy gradient estimator satisfies
\begin{equation} \label{eq:ABC}
\E{\norm{\hnabla_m J(\theta)}^2} \leq 2A(J^*-J(\theta)) + B\norm{\nabla J_H(\theta)}^2 + C, \tag{ABC}
\end{equation}
for all $\theta\in\R^d$.
\end{assumption}
The ABC assumption effectively summarizes a number of popular and more restrictive assumptions commonly used in non-convex optimization. Indeed, the bounded variance of the stochastic gradient assumption~\citep{ghadimi2013Sstochastic}, the gradient confusion assumption~\citep{sankararaman2020impact}, the sure-smoothness assumption~\citep{lei2020stochastic}, the convex expected smoothness assumption~\citep{gower2019sgd,gower2021stochastic} and different variants of strong growth assumptions proposed by~\citet{schmidt2013fast,vaswani2019fast} and~\citet{bottou2018optimization} can all be seen as specific cases of Asm.~\ref{ass:ABC}. %have been to derive complexity guarantees for stochastic gradient descent (SGD).
The ABC assumption has been shown to be %currently 
the weakest among all existing assumptions to provide convergence guarantees for SGD for the minimization of non-convex smooth functions.
%Indeed, the ABC assumption combined with the smoothness of $J(\cdot)$ was used to establish the convergence of SGD in~\citep{khaled2020better} for the minimization of nonconvex expectations.
% as the weakest assumption for SGD to converge in the nonconvex optimization problems. % with the following guarantees.
A more detailed discussion of the assumption for non-convex optimization convergence theory can be found in Thm.~1 in~\citet{khaled2020better}.

We state our main  convergence theorem, that we will then develop into several corollaries.
%% I didn't really understand this next sentence. So I re-wrote and moved it.
%The generality of the ABC assumption also sheds light on the sample complexity analysis of RL optimization. This is the main assumption of the paper that allows us to establish a better understanding of the performance of the vanilla PG as we show next. We adapt Thm.~2 in~\citet{khaled2020better} and obtain the following FOSP convergence guarantee.
\begin{theorem} \label{pro:ABC}
Suppose that Asm.~\ref{ass:smooth},~\ref{ass:trunc} and~\ref{ass:ABC} hold. Consider the iterates $\theta_t$ of the PG method~\eqref{eq:GA} with stepsize $\eta_t = \eta \in \left(0, \frac{2}{LB}\right)$ where $B=0$ means
that $\eta \in (0, \infty) $.
Let $\delta_0 \eqdef J^* - J(\theta_0)$. It follows that
%If $A>0$, then PG defined in~\eqref{eq:GA} satisfies
\vspace{-.1in}
\begin{align}\label{eq:min.performance}
\min_{0\leq t\leq T-1}\E{\norm{\nabla J(\theta_t)}^2} \leq \frac{2\delta_0(1+L\eta^2A)^T}{\eta T(2-LB\eta)} \quad \\
  + \frac{LC\eta}{2-LB\eta}  + \left(\frac{2D(3-LB\eta)}{2-LB\eta} + D'^2\gamma^{H}\right)\gamma^H.\nonumber
\end{align}
In particular if $A=0$, we have
\begin{align} \label{eq:A=0}
&\E{\norm{\nabla J(\theta_U)}^2} \leq \frac{2\delta_0}{\eta T(2-LB\eta)} \quad\quad \\
& + \frac{LC\eta}{2-LB\eta} + \left(\frac{2D(3-LB\eta)}{2-LB\eta} + D'^2\gamma^{H}\right)\gamma^H,\nonumber
\end{align}
where $\theta_U$ is uniformly sampled from $\{\theta_0, \cdots, \theta_{T-1}\}$.
\end{theorem}
%We give the proof of Thm.~\ref{pro:ABC} in App.~\ref{sec:proof ABC}. While Thm.~\ref{pro:ABC} is based on Thm.~2 in~\citet{khaled2020better}, our proof has to take care of the specific structure of PG estimators, notably the bias due to the truncation error.
Thm.~\ref{pro:ABC} provides a general characterization of the convergence of PG as a function of all the constants involved in the assumptions on the problem and the policy gradient estimator. \ruiline{Refer to App.~\ref{sec:related_work_ABC} for a discussion comparing the technical aspects of this result compared to~\citet{khaled2020better}.}
From~\eqref{eq:min.performance} we derive the sample complexity as follows.
\begin{corollary} \label{cor:ABC}
Consider the setting of Thm.~\ref{pro:ABC}.
Given  $\epsilon > 0$, let $\eta = \min\big\{\frac{1}{\sqrt{LAT}}, \frac{1}{LB}, \frac{\epsilon}{2LC}\big\}$ and the horizon $H = \cO(\log\epsilon^{-1})$. If the number of iterations $T$ satisfies
\begin{align}
%    \eta &= \min\left\{\frac{1}{\sqrt{LAT}}, \frac{1}{LB}, \frac{\epsilon}{2LC}\right\}, \nonumber \\
    T \; \geq \; \frac{12\delta_0L}{\epsilon^2}\max\left\{B, \frac{12\delta_0A}{\epsilon^2}, \frac{2C}{\epsilon^2}\right\}, \label{eq:T} %\\
%    H &= \cO(\log\epsilon^{-1}), \nonumber
\end{align}
then $\min_{0\leq t\leq T-1}\E{\norm{\nabla J(\theta_t)}^2} = \cO(\epsilon^2)$. 
%the step size is set to $\eta = \min\{\frac{1}{\sqrt{LAT}}, \frac{1}{LB}, \frac{\epsilon}{2LC}\}$, $T \geq \frac{12\delta_0L}{\epsilon^2}\max\{B, \frac{12\delta_0A}{\epsilon^2}, \frac{2C}{\epsilon^2}\} $ and $H = \cO(\log\epsilon^{-1})$, we have $\min_{0\leq t\leq T-1}\E{\norm{\nabla J(\theta_t)}^2} \leq \epsilon$.
\end{corollary}

Despite the generality of the ABC assumption, Cor.~\ref{cor:ABC} recovers the best known iteration complexity  for vanilla PG in several well-known  cases.

First,~\eqref{eq:T} recovers the $\cO(\epsilon^{-2})$ iteration complexity of the exact gradient method as a special case. To see this, let $H=m = \infty$  and $\hnabla_m J(\theta) = \nabla J(\theta)$ in~\eqref{eq:GA},  thus Asm.~\ref{ass:trunc} and~\ref{ass:ABC} hold automatically with $A=C=D=D'=0$ and $B=1$. By~\eqref{eq:T}, this shows that for any policy and MDP that satisfy the smoothness property (Asm.~\ref{ass:smooth}), the exact full PG converges to a $\epsilon$-FOSP in $T = \cO(\epsilon^{-2})$ iterations. This is the state-of-the-art convergence rate for the exact gradient descent on non-convex objectives without any other assumptions~\citep{beck2017first}.

Second, we recover sample complexity for stochastic vanilla PG. %As we can %rarely compute the exact gradient, in general $A, C, D, D'$ are not all $0$. 
From Cor.~\ref{cor:ABC}, notice that there is no restriction on the batch size $m$. By choosing $m = \cO(1)$, Eq.~\eqref{eq:T} shows that with $TH=\widetilde{\cO}(\epsilon^{-4})$ samples (i.e., single-step interaction with the environment and single sampled trajectory per iteration), the vanilla PG either with updates~\eqref{eq:REINFORCE} or~\eqref{eq:GPOMDP} is guaranteed to converge to an $\epsilon$-stationary point.
Our sample complexity matches the results of~\citet{papini2020safe,zhang2020global,liu2020animproved,xiong2021non-asymptotic}, but improve upon them in generality, i.e., by recovering the exact PG analysis, providing wider range of parameter choices and using the weaker ABC assumption (see Sec.~\ref{sec:expected} for more details). %Our result in Cor.~\ref{cor:ABC} also matches the $\cO(\epsilon^{-4})$  iteration complexity of SGD in  the smooth and non-convex setting~\citep{ghadimi2013Sstochastic,lei2020stochastic,shapiro2021lectures}.
%
%In short, for both the exact and stochastic PG, we recover the state-of-the-art dependency on $\epsilon$ under the ABC assumption.
 
%***
%
%****
%As a consequence of the generality of the ABC assumption, we then show how Thm.~\ref{pro:ABC} can be instantiated in several corollaries.  Each corollary was either know and was developed in it's own paper .... ****\rob{Thinking about re-write.}
%
% capturing a particular state-of-the-art re
%******
%
%Since the ABC assumption is not easily interpretable, we show in the next section that it is a general assumption and Thm.~\ref{pro:ABC} can be instantiated in specific cases as ``corollaries''.
%In particular, even though Thm.~\ref{pro:ABC} is general, we show under special cases that, it recovers a large portion of the existing results with the same convergence rates and provides wider range of parameter choices while remaining the optimal convergence rates. 
%

\subsection{\ruiline{Global optimum convergence under relaxed weak gradient domination}}
\label{sec:weak}

%As mentioned in Sec.~\ref{sec:intro}, the global optimum convergence of PG is an important topic for RL literatures. 
In this section, we present a global optimum convergence of the vanilla PG when the relaxed weak gradient domination assumption is available, in addition to the ~(\nameref{ass:ABC}) assumption.
%We first define this assumption and then establish the convergence of vanilla PG for $J$ satisfying it and~(\nameref{ass:ABC}) assumption.
%As~\citet{agarwal2021theory,mei2020ontheglobal,xiao2022convergence} did for the exact policy gradient update, relying on the following gradient domination assumption, we establish a global optimum convergence guarantee and the sample complexity analysis for the stochastic vanilla PG.

\begin{assumption}[Relaxed weak gradient domination] \label{ass:weak}
 We say that $J$ satisfies the relaxed weak gradient domination condition if for all $\theta\in\R^d$, 
 there exists 
 $\mu > 0$ and $\epsilon' \geq 0$ such that
 %$\mu > 0$ such that
 \begin{align} \label{eq:weak}
 \epsilon' + \norm{\nabla J_H(\theta)} \; \geq \; 2\sqrt{\mu}\left(J^* - J(\theta)\right). %\tag{weak PL}
 \end{align}
\end{assumption}

The relaxed weak gradient domination is an extension of weak gradient domination\footnote{The weak gradient domination is the special case of the Kurdyka-{\L}ojasiewicz (KL) condition with KL exponent $1$~\citep{kurdyka1998gradients}.}~\citep{agarwal2021theory,mei2020ontheglobal,mei2021leveraging} where $\epsilon' = 0$.
Equipped with this assumption, we obtain an average regret convergence as a direct consequence of Cor.~\ref{cor:ABC} (see Cor.~\ref{cor:regret} in App.~\ref{sec:regret}). With the same assumption, we also obtain a new global optimum convergence guarantee (see Thm.~\ref{pro:weak} in App.~\ref{sec:pro_weak} for the full details).
%From Thm.~\ref{pro:weak}, we derive the following sample complexity of PG.
% the sample complexity is $\widetilde{\cO}(\epsilon^{-3})$. \ruiline{To be continued ...}
\begin{corollary} \label{cor:weak}
Consider the setting of Thm.~\ref{pro:weak}. Given $\epsilon > 0$, let the horizon $H = \cO(\log \epsilon^{-1})$. If $\epsilon' = 0$, we choose the number of iterations $T = \cO(\epsilon^{-3})$; if $\epsilon' > 0$, we choose $T = \cO((\epsilon')^{-2}\epsilon^{-1})$. Then $\min\limits_{t \in \{0, 1, \cdots, T\}} J^* - \E{J(\theta_t)} \leq \cO(\epsilon) + \cO(\epsilon').$
\end{corollary}
\ruiline{Consequently, when $\epsilon' = \Theta(\epsilon)$ we have that the complexity of PG to reach a global optimum is $\cO(\epsilon^{-3})$. Thus the   relaxed weak gradient domination has afforded us a factor of $\epsilon^{-1}$ improvement as compared to the $\cO(\epsilon^{-4})$ complexity in Corollary~\ref{cor:ABC}. The relaxed weak gradient domination is an assumption that is unique to PG methods. In Sec.~\ref{sec:FI}, we show that the Fisher-non-degenerate parametrized policy satisfies this assumption.}

% !TEX root = summary.tex

\section{Applications}
\label{sec:app}

In this section we show how the ABC assumption can be used to unify many of the current assumptions used in the literature. %Indeed, the ABC includes many of assumptions in use as a special case. 
In Figure~\ref{fig:hierarchy} we collect all these special cases in a hierarchy tree. Then for each special case we give the sample complexity  of PG as a corollary of Thm~\ref{pro:ABC}. Each of our corollaries match the best known results in these special cases, while also providing a wider range of parameter choices and, in some cases, improving the dependency on some terms in the bound (e.g., the discount factor $\gamma$). Finally, we show that the relaxed weak gradient domination assumption holds for  Fisher-non-degenerate parametrized policies, thus leading to new improved sample complexity result for this setting.

%\subsection{Convergence under the expected Lipschitz and smooth policy assumptions}

\subsection{Expected Lipschitz and smooth policies}
\label{sec:expected}

%In this section, we instantiate the general statement of Thm.~\ref{pro:ABC} under (more restrictive) common assumptions on the policy space and we recover existing results for a wide range of the parameter selections. We also provide a less restrictive dependency on the discount factor $\gamma$. %for problem-specific parameters.
%% Rob: Not at all clear at this point, and a side track
% more accurate dependencies of $(1-\gamma)^{-1}$ on problem-specific parameters. %commonly used to recover existing sample complexity guarantees.%provide a sufficient, but common condition that satisfies Assumption~\ref{ass:ABC} and the smoothness of $J(\cdot)$.

%\subsubsection{ABC assumption as the weakest assumption}

We consider the {\bf expected Lipschitz and smooth policy} (E-LS) assumptions proposed by~\citet{papini2019smoothing}\footnote{While \citet{papini2019smoothing} refers to this assumption as \textit{smoothing policy}, we prefer the alternative name expected Lipschitz and smooth policy, as they not only induce the smoothness of $J$ (see Lemma~\ref{lem:smoothJ}), but also the Lipschitzness (see Lemma~\ref{lem:lipschitzJ}). In~\citet{papini2019smoothing}, they also assume that $\EE{a\sim\pi_\theta(\cdot\mid s)}{\norm{\nabla_\theta\log\pi_\theta(a \mid s)}}$ is bounded, while it is a direct consequence of~\eqref{eq:G2} by Cauchy-Schwarz inequality.}.
\begin{assumption}[E-LS\label{E-LS}] \label{ass:lipschitz_smooth_policy}
There exists constants $G, F > 0$ such that for every state $s\in\cS$, the expected gradient and Hessian of $\log\pi_\theta(\cdot\mid s)$ satisfy 
\begin{align}
\EE{a\sim\pi_\theta(\cdot\mid s)}{\norm{\nabla_\theta\log\pi_\theta(a \mid s)}^2} \; &\leq \; G^2, \label{eq:G2} \\
\EE{a\sim\pi_\theta(\cdot\mid s)}{\norm{\nabla^2_\theta\log\pi_\theta(a\mid s)}} \; &\leq \; F. \label{eq:F} 
\end{align}
\end{assumption}
We call the above \emph{Expected} Lipschitz and Smooth (E-LS), due to the expectation of $a\sim\pi_\theta(\cdot\mid s)$, in contrast to the more restrictive {\bf Lipschitz and smooth policy} (LS) assumption 
\begin{align}
\norm{\nabla_\theta\log\pi_\theta(a \mid s)} \leq G \ \ \mbox{ and } \ \ \norm{\nabla^2_\theta\log\pi_\theta(a\mid s)} \leq F, \label{eq:lipschitz_smooth_policy} \tag{LS}
\end{align}
for all $(s, a) \in \cS \times \cA$.
The (LS) assumption is widely adopted in the analysis of vanilla PG~\citep{zhang2020global} and variance-reduced PG methods, e.g.~\citet{shen2019hessian,xu2020animproved,xu2020sample,yuan2020stochastic,huang2020momentum,pham2020hybrid,liu2020animproved,zhang2021convergence}. It is also a relaxation of the element-wise boundness of $\left|\frac{\partial}{\partial \theta_i}\log\pi_\theta(a \mid s)\right|$ and $\left|\frac{\partial^2}{\partial \theta_i \partial \theta_j}\log\pi_\theta(a \mid s)\right|$
assumed by~\citet{pirotta2015policy} and~\citet{svrpg}
% one in~\citet{pirotta2015policy,svrpg}, which assume that $\left|\frac{\partial}{\partial \theta_i}\log\pi_\theta(a \mid s)\right|$ and $\left|\frac{\partial^2}{\partial \theta_i \partial \theta_j}\log\pi_\theta(a \mid s)\right|$ are bounded element-wise.

\subsubsection{Expected Lipschitz and smooth policy is a special case of ABC}
\label{sec:weakest}

In the following lemma we show that (E-LS) implies the ABC assumption. 
%\begin{figure}
%\centering
%\begin{tikzcd}
%\eqref{eq:lipschitz_smooth_policy} \arrow[r] & (\mbox{\nameref{E-LS}}) \arrow[r] & \eqref{eq:ABC}
%\end{tikzcd}
%\caption{A hierarchy between the assumptions we present throughout the paper. An arrow indicates an implication.}
%\label{fig:hierarchy}
%\end{figure}
\begin{figure*}[t]
\centering
\begin{tikzcd}
 \boxed{\mbox{Softmax with log barrier~\eqref{eq:barrier}}}  \arrow[r]  & \boxed{\colorbox{blue!20}{\ref{eq:ABC}}} & \boxed{\mbox{Softmax with entropy~\eqref{eq:entropy}}} \arrow[l] \\
 \mlnode{Gaussian policy~\eqref{eq:gauss} \\ (unbounded action space)}
 %\boxed{\mbox{Gaussian~\eqref{eq:gauss} (unbounded action space)}}  
 \arrow[r]   & \boxed{\mbox{\nameref{E-LS}}}  \arrow[u] &  \\ %\boxed{\mbox{LQR?}}  \arrow[lu] \\
 \mlnode{Gaussian policy~\eqref{eq:gauss} \\ (bounded action space)}
 %\boxed{\mbox{Gaussian~\eqref{eq:gauss} (bounded action space)}}  
 \arrow[r]    & \boxed{\mbox{\ref{eq:lipschitz_smooth_policy}}} \arrow[u] &  \boxed{\mbox{Softmax~\eqref{eq:softmax_main}}} \arrow[l] %\arrow[lu]
\end{tikzcd}
\caption{A hierarchy between the assumptions we present throughout the chapter. An arrow indicates an implication.}
\label{fig:hierarchy}
\end{figure*}
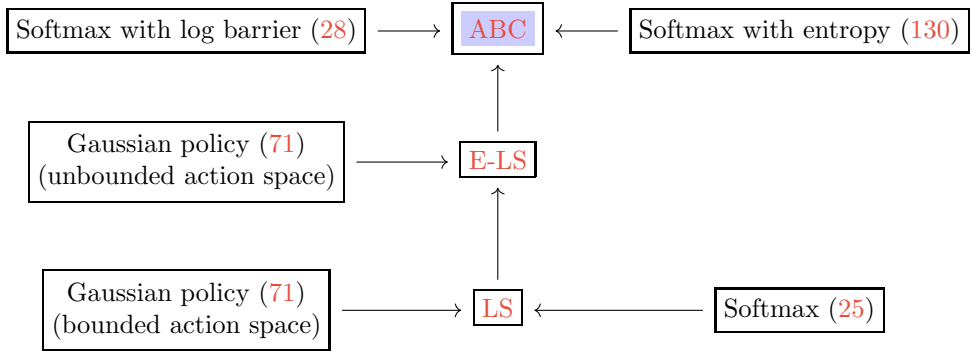
%From Lemma 16 in~\citep{papini2019smoothing}, softmax log-linear policy defined in~\eqref{} also satisfies this assumption.
\begin{lemma} \label{lem:ABC}
Under Asm.~\ref{ass:lipschitz_smooth_policy}, consider a truncated gradient estimator defined either in~\eqref{eq:REINFORCE} or~\eqref{eq:GPOMDP}. Asm.~\ref{ass:ABC} holds with $A=0, B=1-\frac{1}{m}$ and $C = \frac{\nu}{m}$, that is,
\begin{eqnarray} \label{eq:nu}
\E{\norm{\hnabla_m J(\theta)}^2} \leq \left(1-\frac{1}{m}\right)\norm{\nabla J_H(\theta)}^2 + \frac{\nu}{m},
\end{eqnarray}
where $m$ is the mini-batch size, and $\nu = \frac{HG^2\cR_{\max}^2}{(1-\gamma)^2}$ when using REINFORCE gradient estimator~\eqref{eq:REINFORCE} or $\nu = \frac{G^2\cR_{\max}^2}{(1-\gamma)^3}$ when using GPOMDP gradient estimator~\eqref{eq:GPOMDP}.
\end{lemma}

\paragraph*{Bounded variance of the gradient estimator.} Interestingly, from~\eqref{eq:nu} we immediately obtain
\begin{eqnarray} \label{eq:vargrad}
\mathbb{V}\mbox{ar}\left[\hnabla_m J(\theta)\right]
& = & \E{\norm{\hnabla_m J(\theta)}^2} - \norm{\nabla J_H(\theta)}^2 \nonumber \\ 
& \overset{\eqref{eq:nu}}{\leq} & \frac{\nu - \norm{\nabla J_H(\theta)}^2}{m}
\; \leq \; \frac{\nu}{m}, 
\end{eqnarray}
which was used as an assumption by~\citet{svrpg,xu2020animproved,xu2020sample,yuan2020stochastic,huang2020momentum,liu2020animproved}. Yet~\eqref{eq:vargrad} needs not to be an additional assumption since it is a direct consequence of Asm.~\ref{ass:lipschitz_smooth_policy}. 

%A similar\footnote{\rob{Similar how so?}} result to
%Eq.~\eqref{eq:vargrad} was given in Lemma 17 and 18 in~\citet{papini2019smoothing} for both REINFORCE and GPOMDP. \rob{``Similar'' is a bit too vague. Either say exactly what you mean, or don't say anything?}
%% \rob{I've commented this out, because you already detail this in the appendix.}

The \eqref{eq:lipschitz_smooth_policy}
and  (\nameref{E-LS}) form the backbone of our hierarchy of assumptions in Figure~\ref{fig:hierarchy}. In particular, 
~\eqref{eq:lipschitz_smooth_policy}  implies~(\nameref{E-LS}), and thus ABC is  the weaker (and most general) assumption of the three.
%making the ABC assumption the weakest (and most general) 
%. Consequently,~\eqref{eq:ABC} is the weakest, and hence the most general, among~\eqref{eq:lipschitz_smooth_policy},~(\nameref{E-LS}) and~\eqref{eq:ABC} (see Figure~\ref{fig:hierarchy}). 
%We  formalize  this statement in  Cor.~\ref{cor:weakest}.

\begin{corollary} \label{cor:weakest}
The~\eqref{eq:ABC} assumption is the weakest condition compared to~\eqref{eq:lipschitz_smooth_policy} and~(\nameref{E-LS}).
\end{corollary}

%\rui{Need to provide examples s.t. the implications are strict.}

\subsubsection{Sample complexity analysis for stationary point convergence}
%\label{sec:sufficient}

Of independent interest to the ABC assumption, Asm.~\ref{ass:lipschitz_smooth_policy} also implies the smoothness of $J(\cdot)$ and the truncated gradient assumptions as reported in the following lemmas.
\begin{lemma} \label{lem:smoothJ}
Under Asm.~\ref{ass:lipschitz_smooth_policy}, $J(\cdot)$ is $L$-smooth, namely $\norm{\nabla^2 J(\theta)} \leq L$ for all $\theta$ which is a sufficient condition of Asm.~\ref{ass:smooth}, with
\begin{eqnarray}\label{eq:LsmoothJ}
L &=& \frac{\cR_{\max}}{(1-\gamma)^2}\left(G^2 + F\right).
\end{eqnarray}
\end{lemma}
The smoothness constant~\eqref{eq:LsmoothJ} is tighter by a factor of $1-\gamma$ as compared to the  smoothness constant proposed in~\citet{papini2019smoothing}. 
%When $\gamma$ is close to $1$, we improve the smoothness constant proposed in~\citep{papini2019smoothing} by a factor of $1-\gamma$ by a sharper analysis.
This is the tightest upper bound of $\nabla^2 J(\cdot)$ we are aware of in the existing literature. 
%Interestingly,~\citet[][Lemma B.1]{liu2020animproved} achieves a better smoothness constant $\frac{F\cR_{\max}}{(1-\gamma)^2}$ of $J(\cdot)$ under more restrictive assumption~\eqref{eq:lipschitz_smooth_policy} without bounding the Hessian. See App.~\ref{sec:better_cst} for a detailed discussion on the smoothness constant and the Hessian of $J(\cdot)$.
%\footnote{After publishing this paper, thanks to Yanli Liu, we are aware of Lemma B.1 in~\citet{liu2020animproved}, where their upper bound $L = \frac{F\cR_{\max}}{(1-\gamma)^2}$ under more restrictive assumption~\eqref{eq:lipschitz_smooth_policy} is slightly tighter than ours. See App.~\ref{sec:better_cst} for more details.} 
(see App.~\ref{sec:better_cst} for more details).
\begin{lemma} \label{lem:trunc}
Under Asm.~\ref{ass:lipschitz_smooth_policy}, Asm.~\ref{ass:trunc} holds with
\begin{align}
D  \; &= \; \frac{D'G\cR_{\max}}{(1-\gamma)^{3/2}}, \\ 
D' \; &= \; \frac{G\cR_{\max}}{1-\gamma}\sqrt{\frac{1}{1-\gamma} + H}. \label{eq:d'}
\end{align}
\end{lemma}

The coefficient $D'$ in~\eqref{eq:d'} got improved and is tighter by a factor of $(1-\gamma)^{1/2}$ as compared to the same term analysed in Lemma B.1 in~\citet{liu2020animproved}.

As a by-product, in Lemma~\ref{lem:lipschitzJ} in the appendix, we also show that $J(\cdot)$ is Lipschitz under Asm.~\ref{ass:lipschitz_smooth_policy} with a tighter Lipschitzness constant, as compared to~\citet{papini2019smoothing,xu2020sample,yuan2020stochastic}. See more details in App.~\ref{sec:lipschitz}.

Now we can establish the sample complexity of vanilla PG for the expected Lipschitz and smooth policy assumptions as a corollary of Thm.~\ref{pro:ABC} and Lemmas~\ref{lem:ABC},~\ref{lem:smoothJ}, and~\ref{lem:trunc}.
\begin{corollary} \label{cor:expected}
Suppose that Asm.~\ref{ass:lipschitz_smooth_policy} is satisfied. Let $\delta_0 \eqdef J^* - J(\theta_0)$. 
The PG method applied in~\eqref{eq:GA} with a mini-batch sampling of size $m$ and constant step size
\vspace{-.1in}
\begin{eqnarray} \label{eq:stepsize}
\eta &\in& \Big(0,  \;\frac{2}{L\big(1-1/m\big)}\Big),
\end{eqnarray}
satisfies
\begin{align} \label{eq:A=0B=1C=g/m}
& \E{\norm{\nabla J(\theta_U)}^2} \leq \frac{2\delta_0}{\eta T\left(2-L\eta\left(1-\frac{1}{m}\right)\right)} \nonumber \\
&\; + \frac{L\nu\eta}{m\left(2-L\eta\left(1-\frac{1}{m}\right)\right)} \nonumber \\ 
&\; + \left(\frac{2D\left(3-L\eta\left(1-\frac{1}{m}\right)\right)}{2-L\eta\left(1-\frac{1}{m}\right)} + D'^2\gamma^{H}\right)\gamma^H,
\end{align}
where $\nu, L$ and $D, D'>0$ are provided in Lemmas~\ref{lem:ABC},~\ref{lem:smoothJ} and~\ref{lem:trunc}, respectively.
\end{corollary}
We first note that Cor.~\ref{cor:expected}  imposes no restriction on the batch size, allowing us to analyse both exact full PG and its stochastic variants REINFORCE and GPOMDP. For exact  PG, i.e., $H=m = \infty$, we recover the $\cO(1/T)$ convergence. This translates to an iteration complexity $T = \cO\left(\frac{1}{\epsilon^2}\right)$ with a constant step size $\eta = \frac{1}{L}$ to guarantee $\E{\norm{\nabla J(\theta_U)}^2} = \cO(\epsilon^2)$. 
%From Cor.~\ref{cor:expected}, we first notice that we impose no restriction on the batch size. When considering the exact full PG, i.e. $m = \infty$, by setting $\frac{1}{m} = 0$, we recover the iteration complexity $T = \cO\left(\frac{1}{\epsilon^2}\right)$ with a constant step size $\eta = \frac{1}{L}$ to guarantee $\E{\norm{\nabla J(\theta_U)}^2} = \cO(\epsilon^2)$. 
On the other extreme, when $m = 1$, by~\eqref{eq:stepsize} we have that $\eta\in(0,\infty)$, i.e., we place no restriction on the step size. In this case, we have that~\eqref{eq:A=0B=1C=g/m} reduces to 
\[ 
\E{\norm{\nabla J(\theta_U)}^2} \; \leq \; \frac{\delta_0}{\eta T} + \frac{L\nu\eta}{2} 
+ \left(3D + D'^2\gamma^{H}\right)\gamma^H.
\]
Thus  the stepsize $\eta$ controls the trade-off between the rate of convergence 
$\frac{1}{\eta T}$ and leading constant term $\frac{L\nu\eta}{2} $. 
%the neighbourhood of convergence (the terms independent of $1/T$) 
%
%This result shows that the average of the gradient norm squared will converge to a neighbourhood around zero with the rate of $\frac{1}{\eta T}$. The size of the neighbourhood is controlled by $\frac{\eta}{m}$, i.e. the step size and the batch size.
Using Cor.~\ref{cor:expected}, next we develop an explicit sample complexity for PG methods.
%As in Thm.~\ref{pro:ABC}, we can derive the explicit sample complexity guarantees as follows. 
\begin{corollary} \label{cor:sample_complexity}
%Suppose that Asm.~\ref{ass:lipschitz_smooth_policy} is satisfied.
Consider the setting of Corollary~\ref{cor:expected}.
 For a given $\epsilon>0$, by choosing the mini-batch size $m$ such that $1 \leq m \leq \frac{2\nu}{\epsilon^2}$, the step size $\eta = \frac{\epsilon^2m}{2L\nu}$, the number of iterations $T$ such that
\begin{align} \label{eq:sample_complexity}
    Tm \geq \frac{8\delta_0L\nu}{\epsilon^4} =  
    \begin{cases}
    \cO\left(\frac{H}{(1-\gamma)^4\epsilon^4}\right) \quad \mbox{for REINFORCE} \\ 
    \cO\left(\frac{1}{(1-\gamma)^5\epsilon^4}\right) \quad \mbox{for GPOMDP}
    \end{cases}
\end{align}
and the horizon $H = \cO\left((1-\gamma)^{-1}\log\left(1/\epsilon\right)\right)$, then $\E{\norm{\nabla J(\theta_U)}^2} = \cO(\epsilon^2)$. 
\end{corollary}
%\vspace{-.3cm}
\paragraph{Remark.} Given the horizon $H = \cO\left((1-\gamma)^{-1}\log\left(1/\epsilon\right)\right)$, we have that 
\eqref{eq:sample_complexity} shows that the sample complexity of GPOMDP is a factor of $\log\left(1/\epsilon\right)$ smaller than that of REINFORCE. 
%\rob{I think we should remove these remarks. Is it not known already that GPOMDP has a better dependency on the Horizon? }
%From~\eqref{eq:sample_complexity} and the horizon $H = \cO\left((1-\gamma)^{-1}\log\left(1/\epsilon\right)\right)$, it shows that the sample complexity $Tm \times H$ of GPOMDP is improved by $\log\left(1/\epsilon\right)$ compared to the one of REINFORCE.

Cor.~\ref{cor:sample_complexity} greatly extends the range of parameters for which PG is guaranteed to converge within the existing literature.
It shows that it is \emph{possible} for vanilla policy gradient methods to converge with a mini-batch size per iteration from $1$ to $\cO(\epsilon^{-2})$ and a constant step size chosen accordingly between $\cO(\epsilon^2)$ and $\cO(1)$, while still achieving the $Tm \times H = \widetilde{\cO}\left(\epsilon^{-4}\right)$ optimal complexity.

%optimal sample complexity $Tm \times H$ remains the same $\widetilde{\cO}\left(\epsilon^{-4}\right)$ complexity.
%% \rob{We keep repeating that it is FOSP convergnece and pointing to the same references multiple times. Maybe we don't need to keep repeating this?}
% for FOSP convergence as known in the literature, i.e., $\widetilde{\cO}\left(\epsilon^{-4}\right)$~\citep{papini2020safe,zhang2020global,xiong2021non-asymptotic}.
  In particular, Cor.4.4 in~\citet{zhang2020global}, Prop.1 in~\citet{xiong2021non-asymptotic} and Thm.E.1 in~\citet{liu2020animproved} establish $\widetilde{\cO}\left(\epsilon^{-4}\right)$ for FOSP convergence by using the more restrictive assumption~\eqref{eq:lipschitz_smooth_policy}.~\citet{papini2020safe} obtain the same results with the weaker assumption~(\nameref{E-LS}), which is also our case. However, we improve upon all of them by recovering the exact full PG analysis, allowing much wider range of choices for the batch size $m$ and the constant step size $\eta$ to achieve the same optimal sample complexity $\widetilde{\cO}\left(\epsilon^{-4}\right)$.
 % 
 % and getting tighter smoothness constant. \rob{Already mentioned recently this, and it's confusing to add it on the end of the sentence here}
%
%
%Combining the results in Cor.~\ref{cor:expected} and~\ref{cor:sample_complexity}, we highlight that this is not trivial. Given Cor.~\ref{cor:expected}, we provide an upper bound of the stationary point performance analysis~\eqref{eq:A=0B=1C=g/m} for almost ``any'' choices of parameters. As for finding the optimal sample complexity, by Cor.~\ref{cor:sample_complexity} it still allows a wide range of parameters.
%When the batch size $m$ is bigger, the step size can be bigger and the number of iterations $T$ can be smaller.
%This shows that the vanilla PG algorithm is in fact not as sensitive to parameters as previous theories may suggest.
%\rob{ It seems to repeat a similar message from before and the next paragraph. Also I don't understand the "we highlight that this is not trivial." What is the "this" ? So I commented it out for now}
%
%
%In terms of the freedom of the hyperparameter choices, our result is novel. \ruiline{
Indeed, to achieve the optimal sample complexity of FOSP,~\citet{papini2020safe,zhang2020global,xiong2021non-asymptotic,liu2020animproved} do not allow a single trajectory sampled per iteration. 
%\rui{Wrong references. Need to update here.}
They require the batch size $m$ to be either $\epsilon^{-1}$ or $\epsilon^{-2}$. 
%Otherwise, when $m = 1$, their analysis would not return the optimal rate of convergence. 
The existing analysis for vanilla PG that allows $m=1$ that we are aware of is~\citet{zhang2020sample},
%However, when $m > \cO(1)$, the analysis of~\citet{huang2020momentum} does not benefit from larger batch sizes and thus fail to match the optimal sample complexity for large batch sizes.
%
%will not benefit from larger batch size sampling, which means they will not match the optimal sample complexities by increasing the batch size. 
which we compare with in Sec.~\ref{sec:log} under the specific setting of softmax tabular policy with log barrier regularization for the average regret analysis. %}

%\rob{I found this page one of the hardest to read. It's filled with small benefits of our work, entwined with comparisons to alternative work, and background on alternative work. I've commented our a few things that were a bit repetitive or non-pertinent, and re-written a bit others.}

% !TEX root = summary.tex

\subsection{Softmax tabular policy} \label{sec:softmax}

In this section, we instantiate the FOSP convergence results of Cor.~\ref{cor:expected} and~\ref{cor:sample_complexity} in the case of the softmax tabular policy. 
Combined with the specific properties of the softmax, our general theory also recovers the average regret of the global optimum convergence analysis for the softmax with log barrier regularization~\citep{zhang2020sample} and brings new insights of the theory by leveraing the ABC assumption analysis.

%In this section, we give a new sample complexity analysis of the softmax tabular policy and provide  improved smoothness constants for the corresponding policy gradient.

Here the state space $\cS$ and the action space $\cA$ are finite. For all $\theta \in \R^{|\cS||\cA|}$ and any state-action pair $(s,a) \in \cS \times \cA$, consider the following softmax tabular policy
\vspace{-.1in}
\begin{eqnarray} \label{eq:softmax_main}
\pi_\theta(a \mid s) \; \eqdef \; \frac{\exp(\theta_{s,a})}{\sum_{a'\in\cA}\exp(\theta_{s,a'})}.
\end{eqnarray}
We show that the softmax tabular policy satisfies~(\nameref{E-LS}) as illustrated in the following lemma.
\begin{lemma} \label{lem:softmax_expected}
The softmax tabular policy satisfies Asm.~\ref{ass:lipschitz_smooth_policy} with $G^2 = 1 - \frac{1}{|\cA|}$ and $F = 1$, that is, for all $s \in \cS$, we have
\begin{align}
\EE{a\sim\pi_\theta(\cdot\mid s)}{\norm{\nabla_\theta\log\pi_\theta(a \mid s)}^2} &\leq 1 - \frac{1}{|\cA|}, \\
\EE{a\sim\pi_\theta(\cdot\mid s)}{\norm{\nabla^2_\theta\log\pi_\theta(a \mid s)}} &\leq 1.
\end{align}
\end{lemma}
\paragraph*{Remark.} The softmax tabular policy also satisfies~\eqref{eq:lipschitz_smooth_policy} but with a bigger constant (see App.~\ref{sec:fosp:softmax}).

Lemma~\ref{lem:softmax_expected} and the results in Section~\ref{sec:expected} immediately imply that all assumptions including the~\eqref{eq:ABC} assumption of Thm.~\ref{pro:ABC} are verified.
% For the resulting
%smoothness constant of $J(\cdot)$  see Lemma~\ref{lem:softmax_smoothness}.
 %In particular, the smoothness constant is shown in the following lemma.
Thus, as a consequence of Cor.~\ref{cor:expected} and~\ref{cor:sample_complexity}, we have the following sample complexity for the softmax tabular policy.\footnote{The exact statement is similar to Cor.~\ref{cor:sample_complexity}. For the sake of space here we report a more compact statement.}
% we obtain the sample complexity to achieve the FOSP for the softmax tabular policy from Cor.~\ref{cor:expected} presented as follows.

\begin{corollary}[Informal] \label{cor:fosp_softmax}
Given $\epsilon>0$, there exists a range of parameter choices for the batch size $m$ s.t. $1 \leq m \leq \cO(\epsilon^{-2})$, the step size $\eta$ s.t. $\cO(\epsilon^2) \leq \eta \leq \cO(1)$, the number of iterations $T$ and the horizon $H$ such that the sample complexity of the vanilla PG (either REINFORCE or GPOMDP) is $Tm \times H = \widetilde{\cO}\left(\frac{1}{(1-\gamma)^6\epsilon^4}\right)$ to achieve $\E{\norm{\nabla J(\theta_U)}^2} = \cO(\epsilon^2)$. 
\end{corollary}
%Next, by leveraging the structure of softmax, our general theory helps to establish its global optimum convergence with log barrier regularization and brings new insights in the next.

\subsubsection{Global optimum convergence of softmax with log barrier regularization}
\label{sec:log}

%\rui{Shorten this section.}

Leveraging the work of~\citet{agarwal2021theory} and our Thm.~\ref{pro:ABC}, we can establish a global optimum convergence analysis for softmax policies with log barrier regularization. 

Log barrier regularization is often used to prevent the policy from becoming deterministic. Indeed, when optimizing the softmax by PG, policies can rapidly become near deterministic and the optimal policy is usually obtained by sending some parameters to infinity. This can result in an extremely slow convergence of PG. 
%The motivation of such problem is that, when optimizing the softmax by PG, policies can rapidly become near deterministic and the optimal policy is usually obtained by sending some parameters to infinity due to the exponential scaling with parameters $\theta$. Thus PG can result in extremely slow convergence.
~\citet{li2021softmax} show that PG can even take exponential time to converge. To prevent the parameters from becoming too large and to ensure enough exploration, 
an entropy-based regularization term is commonly used to keep the probabilities from getting too small~\citep{williams1991function,volodymyr2016asynchronous,nachum2017bridging,haarnoja2018soft,mei2019onprincipled}
%a log barrier regularization term is commonly used to keep the probabilities from getting too small~\citep{agarwal2021theory}
. Here we study stochastic gradient ascent on a relative entropy regularized objective, softmax with log barrier regularization, which is defined as
\begin{align} \label{eq:barrier}
L_{\lambda}(\theta) &\eqdef 
J(\theta) - \lambda\EE{s\sim\mbox{Unif}_\cS}{\mbox{KL}(\mbox{Unif}_\cA,\pi_\theta(\cdot \mid s))} \nonumber \\ 
& = J(\theta) + \frac{\lambda}{|\cA||\cS|}\sum_{s,a}\log\pi_\theta(a \mid s) + \lambda\log|\cA|,
\end{align}
%\rob{You could remove the first line of the above equations, and just keep the second. This will save a lot of space, since then you do not need to define $\mbox{KL}(p,q) \eqdef \EE{x \sim p}{- \frac{\log q(x)}{\log p (x)}}$ or $\mbox{Unif}_{\chi}$.  }
where the relative entropy for distributions $p$ and $q$ is defined as $\mbox{KL}(p,q) \eqdef \EE{x \sim p}{- \frac{\log q(x)}{\log p (x)}}$, $\mbox{Unif}_{\chi}$ denotes the uniform distribution over a set $\chi$ and $\lambda > 0$ determines the strength of the penalty.
%Let $\hnabla_m L_\lambda(\theta)$ be the stochastic gradient estimator of $L_{\lambda}(\theta)$ using REINFORCE or GPOMDP with batch size $m$. 
%We show that $\hnabla_m L_\lambda(\theta)$ satisfies~\eqref{eq:ABC} in Lemma~\ref{lem:ABC:log} and $L_{\lambda}(\theta)$ is smooth in Lemma~\ref{lem:smooth:log}.
%Similar to Cor.~\ref{cor:fosp_softmax}, from Thm.~\ref{pro:ABC}, Lemma~\ref{lem:ABC:log} and~\ref{lem:smooth:log}, we have $\{\theta_t\}_{t\geq0}$ converges to a FOSP of $L_\lambda(\cdot)$. We postpone the formal statement of this result to App.~\ref{sec:fosp:log} for the sake of space.

Let $\hnabla_m L_\lambda(\theta)$ be the stochastic gradient estimator of $L_{\lambda}(\theta)$ using REINFORCE or GPOMDP with batch size $m$ (see the closed form of $\hnabla_m L_\lambda(\theta)$ in~\eqref{eq:barrier-nabla}). Thus $\hnabla_m L_\lambda(\theta)$ is an unbiased estimate of the gradient of the truncated function 
\begin{align} \label{eq:barrier_trunc}
L_{\lambda,H}(\theta) \eqdef J_H(\theta) + \frac{\lambda}{|\cA||\cS|}\sum_{s,a}\log\pi_\theta(a \mid s) + \lambda\log|\cA|.
\end{align}
We show in the following that $\hnabla_m L_\lambda(\theta)$ satisfies the~\eqref{eq:ABC}.
\begin{lemma} \label{lem:ABC:log}
Consider $\hnabla_m L_\lambda(\theta)$ by using either RIENFORCE~\eqref{eq:REINFORCE} or GPOMDP~\eqref{eq:GPOMDP}, Asm.~\ref{ass:ABC} holds with $A=0, B=1-\frac{1}{m}$ and $C = \frac{\nu}{m}$, that is,
\begin{align} 
\E{\norm{\hnabla_m L_\lambda(\theta)}^2} \leq \left(1-\frac{1}{m}\right)\norm{\nabla L_{\lambda,H}(\theta)}^2 + \frac{\nu}{m},
\end{align}
 where 
 $\nu =
 2\left(1-\frac{1}{|\cA|}\right)\left(\frac{H\cR_{\max}^2}{(1-\gamma)^2} + \frac{\lambda^2}{|\cS|}\right)
 $ when using REINFORCE or
 $\nu =
 2\left(1-\frac{1}{|\cA|}\right)\left(\frac{\cR_{\max}^2}{(1-\gamma)^3} + \frac{\lambda^2}{|\cS|}\right)
 $ when using GPOMDP.
\end{lemma}

Similar to the softmax case, we show in App.~\ref{sec:fosp:log} that $L_\lambda(\theta)$ is also smooth and verifies Asm.~\ref{ass:trunc}. Thus from Thm.~\ref{pro:ABC}, we have $\{\theta_t\}_{t\geq0}$ converges to a FOSP of $L_\lambda(\cdot)$. We postpone the formal statement of this result to App.~\ref{sec:fosp:log} for the sake of space.
%\begin{corollary}[Informal] \label{cor:fosp:log}
%For any accuracy level $\epsilon$, there exists a range of parameter choices for the batch size $m$ s.t. $1 \leq m \leq \cO(\epsilon^{-2})$, the step size $\eta$, the number of iterations $T$ and the horizon $H$ such that the sample complexity of the vanilla PG (either REINFORCE or GPOMDP) is $Tm \times H = \widetilde{\cO}\left(\frac{1}{(1-\gamma)^6\epsilon^4}\right)$ for $\E{\norm{\nabla L_\lambda(\theta_U)}^2} = \cO(\epsilon^2)$. 
%\end{corollary}

Besides, thanks to Thm.~5.2 in~\citet{agarwal2021theory}, the FOSP of $L_\lambda(\cdot)$ is approximately the global optimal solution of $J(\cdot)$ when the regularization parameter $\lambda$ is sufficiently small. 
%From~\citep{agarwal2021theory} and Cor.~\ref{cor:fosp:log}, we establish the high probability global optimum convergence with sample complexity $\cO(\delta^{-2}\epsilon^{-4})$, i.e. with probability greater than $1- \delta$, we need $\cO(\delta^{-2}\epsilon^{-4})$ samples to achieve $J^* - J(\theta_t) \leq \epsilon$. Since the result is very different to others presented in the main paper, for the sake of homogeneity, see details in Section~\ref{sec:high_proba} in Appendix.
As a by-product, we can also establish a high probability global optimum convergence analysis (App.~\ref{sec:high_proba}). %Since the result is very different to others presented in the main paper, for the sake of homogeneity, see details in .

In the following corollary, we show that we can leverage the versatility of Thm.~\ref{pro:ABC} to derive yet another type of result: a guarantee on the average regret w.r.t.\ the global optimum. 
%focus on an alternative type of result Interestingly,  is versatile that, combined with Thm.~5.2 in~\citep{agarwal2021theory}, we also establish the sample complexity of the average regret of the global optimum as follows.
\begin{corollary} \label{cor:regret_log}
Given $\epsilon > 0$, consider the batch size $m$ such that $1 \leq m \leq \frac{1}{(1-\gamma)^6\epsilon^3}$, the step size $\cO(\epsilon^3) \leq \eta = \frac{(1-\gamma)^3\epsilon^3m}{2L\nu} \leq \cO(1)$ with $L, \nu$ in the setting of Cor.~\ref{cor:fosp:log}
%, the horizon $H = \cO\left(\frac{\log(1/\epsilon)}{1-\gamma}\right)$ and the number of iterations $T$ such that
. If the horizon $H = \cO\left(\frac{\log(1/\epsilon)}{1-\gamma}\right)$ and the number of iterations $T$ is such that
\[
%Tm \; \geq \; \cO\left(\frac{|\cS|^4|\cA|^4\norm{\frac{d_\rho(\theta^*)}{\rho}}_{\infty}^4}{(1-\gamma)^{11}\epsilon^6}\right),
Tm \times H \; \geq \; \widetilde{\cO}\left(\frac{1}{(1-\gamma)^{12}\epsilon^6}\right),
\]
we have
%\vspace{-.3cm}
%\begin{align} \label{eq:regret}
$J^* - \frac{1}{T}\sum_{t=0}^{T-1}\E{J(\theta_t)} \; = \; \cO(\epsilon).$
%\end{align}
\end{corollary}
This result recovers the sample complexity $\widetilde{\cO}(\epsilon^{-6})$ of~\citet{zhang2020sample}. However,~\citet{zhang2020sample} do not study the vanilla policy gradient. Instead, they add an extra phased learning step to enforce the exploration of the MDP and use a decreasing step size.
%Moreover, their result is restricted to the softmax policy parametrization with a log-barrier regularization, which makes their analysis less general. 
Our result shows that such extra phased learning step is unnecessary and the step size can be constant. We also provide a wider range of parameter choices for the batch size and the step size with the same sample complexity.
%and our convergence theory is satisfied for a much larger class of parametrized policies.

As~\citet{agarwal2021theory} mentioned, the regularizer~\eqref{eq:barrier} is more “aggressive” in penalizing small probabilities than the more commonly utilized entropy regularizer. We also show that entropy regularized softmax satisfies the~\eqref{eq:ABC} and provide its FOSP analysis in App.~\ref{sec:entropy}, again thanks to the versatility of Thm.~\ref{pro:ABC}. Notice that for the FOSP convergence, only an asymptotic result was established in Lemma 4.4 in~\citet{ding2021exact}. Thus all proofs and implications in Fig.~\ref{fig:hierarchy} are provided.

\subsection{\ruiline{Fisher-non-degenerate parameterization}}
\label{sec:FI}

%\rui{Introduce FI + compatible asm. and the implied relaxed weak gradient domination given by~\citet{ding2021global}.}

In this section, we study a general policy class that satisfies the following assumption.
\begin{assumption}[Fisher-non-degenerate, Asm.~2.1 in~\citet{ding2021global}] \label{ass:FI}
For all $\theta \in \R^d$, there exists $\mu_F > 0$ s.t.\ the Fisher information matrix $F_\rho(\theta)$ induced by policy $\pi_\theta$ and initial state distribution $\rho$ satisfies
\begin{align} \label{eq:FI}
F_\rho(\theta) &\eqdef \EE{(s,a) \sim v_\rho^{\pi_\theta}}{\nabla_\theta \log \pi_\theta(a \mid s)\nabla_\theta \log \pi_\theta(a \mid s)^\top} \nonumber \\ 
&\geq \mu_F \mI_d, \tag{FI}
\end{align}
where $v_\rho^{\pi_\theta}$ is the state-action visitation measure defined as
\[v_\rho^{\pi_\theta}(s,a) \eqdef (1-\gamma)\mathbb{E}_{s_0 \sim \rho}\sum_{t=0}^\infty\gamma^t\mathbb{P}(s_t = s, a_t = a | s_0, \pi_\theta).\]
\end{assumption}
This assumption is commonly used in the literatures~\citep{liu2020animproved,ding2021global}. Similar conditions of the Fisher-non-degeneracy is also required in other global optimum convergence framework (Asm. 6.5 in~\citet{agarwal2021theory} on the relative condition number). This assumption is satisfied by a wide families of policies, including the Gaussian policy~\eqref{eq:gauss} and certain neural policy. We refer to Sec.~B.2 in~\citet{liu2020animproved} and Sec.~8 in~\citet{ding2021global} for more discussions on the Fisher-non-degenerate setting.

We also need the following \emph{compatible function approximation error} assumption\footnote{We defer the definition of the advantage function $A^{\pi_\theta}$ in App.\ref{sec:proof_FI}.}.
\begin{assumption}[Compatible, Asm.~4.6 in~\citet{ding2021global}] \label{ass:compatible}
For all $\theta \in \R^d$, there exists $\epsilon_{bias} > 0$ s.t.\ the \emph{transferred compatible function approximation error} with $(s,a) \sim v_\rho^{\pi_{\theta^*}}$ satisfies
\begin{align} \label{eq:compatible}
\E{(A^{\pi_\theta}(s,a) - (1-\gamma)u^{*\top}\nabla_\theta \pi_\theta(a \mid s))^2} \leq \epsilon_{bias}, \tag{compatible}
\end{align}
where $v_\rho^{\pi_{\theta^*}}$ is the state-action distribution induced by an optimal policy $\pi_{\theta^*}$, $u^* = (F_\rho(\theta))^\dagger\nabla J(\theta)$.
\end{assumption}
This is also a common assumption~\citep{Wang2020Neural,agarwal2021theory,liu2020animproved,ding2021global}. In particular, when $\pi_\theta$ is a softmax tabular policy~\eqref{eq:softmax}, $\epsilon_{bias}$ is $0$~\citep{ding2021global}; when $\pi_\theta$ is a rich neural policy, $\epsilon_{bias}$ is  small~\citep{Wang2020Neural}.

Combining Asm.~\eqref{eq:FI},~\eqref{eq:compatible} with Asm.~\nameref{E-LS}, by Lemma~4.7 in~\citet{ding2021global}, we know that $J(\cdot)$ satisfies the relaxed weak gradient domination property~\eqref{eq:weak} with $\epsilon' = \frac{\mu_F \sqrt{\epsilon_{bias}}}{(1-\gamma)G}$ and $\mu = \frac{\mu_F^2}{4G^2}$. Consequently, we recover the average regret convergence result $\cO(\epsilon^{-4})$ of~\citet{liu2020animproved} in Cor.~\ref{cor:FI_regret} in App.~\ref{sec:average} with weaker assumption and allowing wider range of parameter choices. We also have the following new global optimum convergence result for the Fisher-non-degenerate parametrized policy.
\begin{corollary} \label{cor:FI}
If the policy $\pi_\theta$ satisfies Asm.~\ref{ass:lipschitz_smooth_policy},~\ref{ass:FI} and~\ref{ass:compatible}, consider the setting of Cor.~\ref{cor:weak} with $\epsilon' = \frac{\mu_F \sqrt{\epsilon_{bias}}}{(1-\gamma)G}$ and $\mu = \frac{\mu_F^2}{4G^2}$. Then $\min\limits_{t \in \{0, 1, \cdots, T\}} J^* - \E{J(\theta_t)} \leq \cO(\epsilon) + \cO(\sqrt{\epsilon_{bias}})$ and the sample complexity $T \times H = \widetilde{\cO}(\epsilon^{-3})$ when $\epsilon_{bias} = 0$ or $T \times H = \widetilde{\cO}((\epsilon_{bias} \cdot \epsilon)^{-1})$ when $\epsilon_{bias} > 0$.
\end{corollary}

\section{Discussion}
\label{sec:discussion}

We believe the generality of Thm.~\ref{pro:ABC} opens the possibility to identify a broader set of configurations (i.e., MDP and policy space) for which PG is guaranteed to converge. In particular, we notice that Asm.~\ref{ass:lipschitz_smooth_policy} despite being very common, is somehow restrictive, as general policy spaces defined by e.g., a multi-layer neural network, may not satisfy it, unless some restriction on the parameters is imposed. Other interesting venues of investigation include whether it is possible to extend the analysis to projected PG, identify counterparts of the ABC assumption for variance-reduced versions of PG and for the improved analysis of~\cite{zhang2021convergence} leveraging composite optimization tools.

\section*{Acknowledgment}

We are indebted to Matteo Papini, who suggested us to consider the more general Assumption~(\nameref{E-LS}) instead of~\eqref{eq:lipschitz_smooth_policy} on an early draft of this work. This suggestion helped us significantly improve all our results in Section~\ref{sec:expected}.

After publishing this paper, we gratefully acknowledge Francesco Orabona who pointed out that the~\ref{eq:ABC} assumption already appeared in 1973 in~\citet{polyakABC1973} and that we did not cite it properly in the previous version. 

%We would also like to thank Yanli Liu who found an incorrect comment in a previous version of this paper stating that~\citet[][Lemma B.1]{liu2020animproved} has the dependency of $(1-\gamma)^{-2}$ for the smoothness constant of the value function due to a recurring mistake in a crucial step in bounding the Hessian. He gave an extended proof of Lemma B.1 in~\citet{liu2020animproved} which is correct. In particular, such proof does not involve the Hessian.

We would also like to thank the anonymous reviewers for their helpful comments.

\bibliography{references}
\bibliographystyle{plainnat}

% Supplementary material: To improve readability, you must use a single-column format for the supplementary material.
\onecolumn

%%%%%%%%%%%%%%%%%%%%%%%%%%%%%%%%%%%%%%%%%%%%%%%%%%%%%%%%%%%%
%\newpage

%\tableofcontents

\newpage

\appendix

\part{Appendix}

\parttoc

Here we provide the related work discussion, the missing proofs from the main paper and some additional noteworthy observations made in the main paper. %Each proposition and lemma have a respect section with its proof.

%Compared to the original submission, we have added a few references in the supplementary material, 
%especially in Section~\ref{sec:related_work}. There were also two typos in the main paper after the original submission.
%\begin{itemize}
%	\item In Corollary~\ref{cor:ABC}, it should be $\min_{0\leq t\leq T-1}\E{\norm{\nabla J(\theta_t)}^2} = \cO(\epsilon^2)$, 
%	not $\min_{0\leq t\leq T-1}\E{\norm{\nabla J(\theta_t)}^2} = \cO(\epsilon^{-2})$.
%	\item The definition of the softmax tabular policy in~\eqref{eq:softmax_main} should be $\pi_\theta(a \mid s)$, 
%	not $\pi_\theta(s \mid a)$.
%\end{itemize}

\section{Related work}
\label{sec:related_work}

We provide an extended discussion for the context of our work, including \ruiline{a discussion comparing the technical novelty of the paper to the finite sum minimization result in~\citet{khaled2020better},} 
a comparison of the convergence theories of vanilla PG and the problem dependent constants. We refer to Algorithm~\ref{alg:pg} as the vanilla PG with $\hnabla_m J(\theta_t)$ defined as either the exact full gradient~\eqref{eq:GD} and~\eqref{eq:GD2*} or the stochastic PG estimator~\eqref{eq:REINFORCE} or~\eqref{eq:GPOMDP}. 
%For clarity, a resume of our contributions compared to the existing literature is presented in Table~\ref{tab:full picture}. 
Furthermore, we discuss future work to extend our general sample complexity analysis to other policy gradient methods and other RL settings.

\begin{algorithm}[tb]
	\caption{Vanilla policy gradient}
	\label{alg:pg}
	\begin{algorithmic}
		\STATE {\bfseries Input:} initialized $\theta_0$, mini-batch size $m$, step size $\eta_0$
		\FOR{$t=0$ {\bfseries to} $T-1$}
		\STATE Sample $m$ trajectories following policy $\pi_{\theta_t}$ from the MDP % from $\cM$, 
		\STATE Compute the policy gradient estimator $\hnabla_m J(\theta_t)$
		\STATE Update $\theta_{t+1} = \theta_t + \eta_t \hnabla_m J(\theta_t)$ and $\eta_t$
		\ENDFOR
	\end{algorithmic}
\end{algorithm}

%\rui{Explain the ``delta'' compared to the ABC paper.}

\subsection{\ruiline{Technical contribution and novelty compared to~\citet{khaled2020better}}}
\label{sec:related_work_ABC}

Our technical novelty compared to~\citet{khaled2020better} is threefold. First, Theorem~\ref{pro:ABC} is not a direct application of Theorem~2 in~\citet{khaled2020better},
%We give the proof of Thm.~\ref{pro:ABC} in App.~\ref{sec:proof ABC}. While Thm.~\ref{pro:ABC} is based on Thm.~2 in~\citet{khaled2020better}, our proof has to take care of the specific structure of PG estimators, notably the bias due to the truncation error.
which requires unbiased estimators of the gradient.  Yet in PG methods, we have to deal with biased estimators due to the truncation of the trajectories. The first technical challenge was to adapt the proof technique to allow for biased gradients and a truncation error. This also explains the need of Assumption~\ref{ass:trunc}. 
Similarly, we need to handle the same challenge for the proof of Theorem~\ref{pro:PL} when adapting the proof of Theorem~3 in~\citet{khaled2020better}.
Second, when considering the results we derived in specific cases in Section~\ref{sec:app}, the difference between our work and~\citet{khaled2020better} is even more significant. All cases studied in~\citet{khaled2020better} (e.g., finite-sum structure) are not applicable for PG methods and we had to derive specific analysis for our specialized settings (soft-max with different regularizers, expected Lipschitz and smooth policies, Fisher-non-degenerate parametrized policies). Furthermore, our focus is on deriving explicit sample complexity, whereas the results in~\citet{khaled2020better} are concerned with convergence rates in terms of number of iterations. These dimensions are where most of the technical work was done. Without this work of developing sample complexity and studying specific cases found in PG literatures, it was not clear at all that the~\eqref{eq:ABC} assumption proposed in~\citet{khaled2020better} would be relevant in RL. Finally, 
% It is true that Thm.~3.4 and Cor.~3.5 are direct adaptations of their paper, and so is Thm.~G.2 when the gradient domination (Asm.~G.1) is available. The difficulty in adapting their results is that in the PG setting, we do not have access to unbiased estimates of the gradient due to the truncation (see comments after Thm.~3.4). This is why we need Asm.~3.2 in extra. \\
% Need to improve the writing to surprise Reviewer 1&6, make the second point sound louder. But I also want to be honest to the response.
%2) However, the objective of the paper is twofold. First, we provide an unified analysis that covers much of the fragmented results in the literature (see detailed discussion below in {\bf R4}'s answer). Second, [1] either focuses on iteration complexity convergence or on specific instances (i.e. finite sum cases) that do not apply to PG setting. We take a significant step forward w.r.t. to them for the sample complexity analysis. 
%When specialized in finite sum minimization, [1] requires $O(1/\epsilon^2)$ for the mini batch size to achieve the optimal rate (see Sec.6 in [1]).
%However,
%Cor.~4.7 and the rest of corollaries (e.g. Cor.~4.11, E.2, E.4, E.5) provide a better parameter range understanding. It shows that, leading to "optimal rate", the bounds still hold for big range of parameters which suggests that the algorithm is not as sensitive to parameters as previous theory may suggest. This result is important and new compared to [1].  \\
we also consider the setting where the relaxed weak gradient domination holds (Assumption~\ref{ass:weak} and Theorem~\ref{pro:weak}). This is an assumption that is unique to PG methods and had not been considered in~\citet{khaled2020better}.  Technically speaking, the proof of Theorem~\ref{pro:weak} is unique and required a different approach (see the arguments following~\eqref{eq:rtweak}).

\subsection{Sample complexity analysis of the vanilla policy gradient }

Despite the success of PG methods in practice, a comprehensive theoretical understanding was lacking until recently. 

%Under special cases,~\citet{agarwal2021theory} also establishes a $\cO(\epsilon^{-2})$ rate of convergence for the exact projected PG in the constrained tabular parametrized policy. 
%By leveraging the hidden convex structure using composite optimization tools with additional assumptions where the constrained tabular parametrized policy satisfies,~\citet{zhang2020variational} and~\citet{zhang2021convergence} obtain an improved convergence rate $\cO(\epsilon^{-1})$ for the exact full gradient.~\citet{mei2020ontheglobal} also establishes the same convergence rate $\cO(\epsilon^{-1})$ for the true gradient in the softmax policy by using the weak gradient domination property (Lojasiewicz inequality). 
%Similarly,~\citet{fazel2018global} establishes a linear convergence rate for the exact full gradient in the linear-quadratic regulators by using the gradient domination property (Polyak-Lojasiewicz condition).
%However, these convergence rates are only conceptual, as we can rarely access the exact full gradient for the update in practice.

\paragraph*{Global optimum convergence of vanilla PG with the exact full gradient.}
We refer to global optimum convergence as an analysis that guarantees that $J^* - J(\theta_T) \leq \epsilon$ after $T$ iterations.
The global optimum convergence results of PG with the exact full gradient have been developed under a number of different specific settings. 

By using a gradient domination property of the expected return, which is also referred to as the Polyak-Lojasiewicz~\eqref{eq:PL} condition \citep{polyak1963gradient,lojasiewicz1963une},~\citet{fazel2018global} show that the linear-quadratic regulator (LQR) converges linearly to the global optimum for PG with the exact full gradient. However, in the LQR setting the function $J$ is not smooth, and thus does not fit into the general setting we considered in this paper. 
Notice that such~\eqref{eq:PL} condition is widely explored by~\citet{bhandari2019global} to identify more general MDP settings. When such~\eqref{eq:PL} condition holds,~\citet{bhandari2019global} show that any stationary point of the policy gradient of the expected return is a global optimum.
More recently, \citet{agarwal2021theory} leveraged a \emph{weak} gradient domination property, also called the weak Polyak-Lojasiewicz condition which is exactly our condition~\eqref{eq:weak} with $\epsilon' = 0$, to show that the projected PG converges to the global optimum with a $\cO(\epsilon^{-2})$ convergence rate in tabular MDPs with tabular policies, also called direct policy parameterization. 
In later work,~\citet{xiao2022convergence} improve this result by a factor of $\epsilon$, i.e., they establish a $\cO(\epsilon^{-1})$ convergence rate for the projected PG in the tabular setting when the exact full gradient is available. At the moment, we could not adapt our general ABC structure to analyze and derive a sample complexity guarantee for the projected PG.
The same convergence rate $\cO(\epsilon^{-1})$ is developed by~\citet{zhang2020variational} by leveraging the hidden convex structure of the cumulative reward and consequently showing that all local optima (i.e., stationary points) are in fact global optima under certain bijection assumptions based on the occupancy measure space (Assumption 1 in~\citet{zhang2020variational}). Notice that the assumptions proposed by~\citet{zhang2020variational} are satisfied in the specific case of the tabular setting. We do not cover this specific assumption in our current analysis.

The global optimum convergence analysis with exact PG is also investigated in the case of softmax tabular policy with or without regularization.~\citet{agarwal2021theory} first provide an asymptotic convergence for the softmax tabular without regularization and a $\cO(\epsilon^{-2})$ convergence rate for the softmax tabular with log barrier regularization. Even though the gradient domination property (\eqref{eq:PL} or \eqref{eq:weak}) is not globally satisfied for the softmax tabular,~\citet{mei2020ontheglobal} prove that it is available by following the path of the iterations with the exact full gradient updates. Such a property is called the non-uniform Lojasiewicz inequality. Consequently,~\citet{mei2020ontheglobal} show a $\cO(\epsilon^{-1})$ convergence rate for the softmax tabular without regularization by the weak gradient domination condition and a linear convergence rate for the softmax tabular with entropy regularization by the gradient domination condition. Finally, \citet{li2021softmax} recently showed that the result of~\citet{mei2020ontheglobal} for softmax tabular policies may actually contain a term that is exponential in the discount factor, thus showing that exact PG may take an exponential time to converge.
  \\[0.2cm]
\noindent \emph{Our Contributions.} We provide a general sample complexity analysis which, when instantiated using  specific settings given in the literature, recovers the same or even slightly improved convergence rates. Indeed, from Corollary~\ref{cor:log:concentration} we recover the $\cO(\epsilon^{-2})$ convergence rate of~\citet{agarwal2021theory} for the softmax tabular with log barrier regularization and improve the rate by a factor of $1-\gamma$ through a better analysis of the smoothness constant.
By leveraging the (relaxed weak) gradient domination properties which hold under the path of the iterations~\citep{mei2020ontheglobal}, we recover their results. That is, we recover the $\cO(\epsilon^{-1})$ convergence rate for the softmax tabular without regularization in Theorem~\ref{pro:weak} and the linear convergence rate for the softmax tabular with entropy regularization in Theorem~\ref{pro:PL}. %However, the results in~\citep{mei2020ontheglobal} depend on constants that are unclear with respect to the parameters of the MDP. We are aware from~\citep{li2021softmax} that it is possible for some settings that the softmax tabular with the exact policy gradient updates take exponential time to converge.

\paragraph*{Sample complexity for FOSP convergence.} The  convergence rates derived for exact PG are representative of the behavior of the algorithm but do not take into account the additional errors due to the stochastic nature of the actual algorithm used in practice. In this paper we mostly focus on the sample complexity of the stochastic vanilla PG for FOSP convergence.
The well known sample complexity for REINFORCE is $\widetilde{\cO}(\epsilon^{-4})$ s.t. $\frac{1}{T}\sum_{t=0}^{T-1}\E{\norm{\hnabla_m J(\theta_t)}^2} \leq \epsilon^2$ after $T$ iterations. 
However, as~\citet{papini2020safe} mentioned, \emph{``formal proofs of this result are surprisingly hard to find both in the policy optimization and in the nonconvex optimization literature.''}~\citet{papini2020safe} give a proof of the result under the expected Lipschitz and smooth policy assumption~(\nameref{E-LS}) in Theorem 7.1. 
When an estimate of the Q-function is available,~\citet{zhang2020global} also establish the same dependency on $\epsilon$ for the sample complexity of FOSP convergence for the policy gradient theorem~\citep{sutton2000policy} under more restrictive Lipschitz and smooth policy assumption~\eqref{eq:lipschitz_smooth_policy}. By adding an additional uniform ergodicity assumption~\citep{mitrophanov2005sensitivity},~\citet{xiong2021non-asymptotic} improve the sample complexity of~\citep{zhang2020global} by some factors of $1-\gamma$ but still has the same dependency on $\epsilon$. \\[0.2cm]
\noindent \emph{Our Contributions.} We establish the sample complexity analysis for the vanilla PG -- REINFORCE~\eqref{eq:REINFORCE} and GPOMDP~\eqref{eq:GPOMDP}. We improve the results of~\citet{papini2020safe,zhang2020global,xiong2021non-asymptotic}  by using weaker assumptions and allowing much wider range of hyper parameters (the batch size $m$ and the constant step size $\eta$) to achieve the optimal sample complexity. Overall, for both the exact and stochastic PG, our general sample complexity analysis recovers the state-of-the-art dependency on $\epsilon$ under the ABC assumption.
%Our analysis also recovers the exact PG analysis as a special case.

\paragraph*{Sample complexity for global optimum convergence.} We refer to sample complexity of global optimum convergence as an analysis that  guarantees that $J^* - \E{J(\theta_T)} \leq \epsilon$ after $T$ iterations. To the best of our knowledge, there is no existing analysis that considers this type of convergence result for the stochastic vanilla PG. As for variance-reduced PG, by using Assumption 1 in~\citet{zhang2020variational} about occupancy distribution,~\citet{zhang2021convergence} establish a $\widetilde{\cO}(\epsilon^{-2})$ sample complexity to achieve the global optimum.\\[0.2cm]
\noindent \emph{Our Contributions.} 
Under the ABC assumption, the smoothness and an additional gradient domination type assumptions~\eqref{eq:weak} and~\eqref{eq:PL}, we establish the faster sample complexity analysis for the global optimum convergence in Section~\ref{sec:weak} and Section~\ref{sec:PL}. More precisely, when the relaxed weak gradient domination assumption~\eqref{eq:weak} is available, we establish $\widetilde{\cO}(\epsilon^{-3})$ sample complexity in Theorem~\ref{pro:weak}. We also show that one wide family of policies, the Fisher-non-degenerate parametrized policies, satisfy this relaxed weak gradient domination assumption. When the gradient domination assumption~\eqref{eq:PL} is available, we establish $\widetilde{\cO}(\epsilon^{-1})$ sample complexity for the global optimum in Theorem~\ref{pro:PL}. It remains an open question whether softmax or softmax with entropy still satisfy the (weak) gradient domination type of assumptions for the stochastic PG updates based on the exact PG analysis of~\citet{mei2020ontheglobal}.
%More precisely, when the gradient domination assumption~\eqref{eq:PL} is available, we establish $\widetilde{\cO}(\epsilon^{-1})$ sample complexity for the global optimum in Theorem~\ref{pro:PL}; when the weak gradient domination assumption~\eqref{eq:weak} is available, we establish $\widetilde{\cO}(\epsilon^{-3})$ sample complexity in Theorem~\ref{pro:weak}. It remains an open question whether softmax or softmax with entropy still satisfy the gradient domination type of assumptions for the stochastic PG updates.
%, which act the same as in the case of exact full gradient updates. 

\paragraph*{Sample complexity for the average regret convergence.} We refer to the sample complexity for average regret as an analysis that guarantees that $J^* - \frac{1}{T}\sum_{t=0}^{T-1}\E{J(\theta_t)} \leq \epsilon$.~\citet{zhang2020sample} show that with sample complexity $\widetilde{\cO}(\epsilon^{-6})$, PG methods can converge to the average regret optimum by using as little as a single sampled trajectory per iteration (i.e., mini-batch size $m=1$) for softmax with log barrier regularization. However, their setting does not use ``vanilla'' PG but a modified version with re-projection meant to guarantee a sufficient level of policy randomization.~\citet{liu2020animproved} obtain faster sample complexity $\widetilde{\cO}(\epsilon^{-4})$ by assuming in addition a Fisher-non-degenerate parameterization, i.e. the Fisher information matrix is strictly lower bounded (Assumption~\ref{ass:FI}), and the compatible function approximation assumption (Assumption~\ref{ass:compatible}). Notice that the softmax with log barrier regularization does not satisfy all these assumptions and they require large batch sizes per iteration. \\[0.2cm]
\noindent \emph{Our Contributions.} We recover the sample complexity for the average regret convergence $\widetilde{\cO}(\epsilon^{-6})$ of~\citet{zhang2020sample} in the softmax with log barrier regularization with the vanilla PG setting . Compared to their results,
we show that the extra phased learning step is unnecessary and the step size can be constant instead of using a decreasing step size. We also provide a wider range of parameter choices for the batch size and the step size with the same sample complexity. For the Fisher-non-degenerate parametrized policy, we also recover the sample complexity for the average regret convergence $\widetilde{\cO}(\epsilon^{-4})$ of~\citet{liu2020animproved} in Corollary~\ref{cor:FI_regret}. Compared to their results, we improve upon them by using weaker assumption~\nameref{E-LS},
allowing much wider range of choices for the batch size $m \in \left[1;\frac{2\nu}{\epsilon^2}\right]$ and the corresponded constant step size $\eta$ to achieve the same optimal sample complexity $\widetilde{\cO}\left(\epsilon^{-4}\right)$.

%~\citet{liu2020animproved} and \citet{zhang2021convergence} unify the lines of work on global optimum convergence and variance reduction in PG. However, these works require either the exact full gradient updates or large batch sizes per iteration. Using a regret minimization analysis,~\citet{zhang2020sample} showed that 
%PG methods can converge with using as little as a single sampled trajectory per iteration  (i.e., mini-batch size $m=1$).
%it is possible to allow a single sampled trajectory (i.e., mini-batch size $m=1$) for the convergence. 
%However, their setting is restricted to softmax policy and does not use ``vanilla'' PG but a modified version with re-projection meant to guarantee a sufficient level of policy randomization.

\subsection{Better analysis of the problem dependent constants}
\label{sec:better_cst}

Throughout the paper, we also provided tighter bounds on the smoothness constants, Lipschitzness constants, and the variance of the gradient estimators under Assumption~(\nameref{ass:lipschitz_smooth_policy}). Notice that the smoothness and Lipschitz constants we consider here are properties of the expected return $J(\cdot)$ in~\eqref{eq:J} or the regularized expected return $L_\lambda(\cdot)$ in~\eqref{eq:barrier}. They depend only on the assumptions and are independent to the specific PG algorithm.
For this reason, below we compare our bounds with work that studies variants of PG other than vanilla PG, where the bounds on the smoothness and Lipschitz constants are also needed. 
% so we do compare with works that may eventually study variants of vanilla PG.
On the other hand, for the variance of the gradient estimators, we only consider the vanilla gradient estimators REINFORCE~\eqref{eq:REINFORCE} and GPOMDP~\eqref{eq:GPOMDP} with batch size $m$. 
A resume of the improved problem dependent constants -- smoothness and Lipschitzness constants, is provided in Table~\ref{tab:constant}.

\begin{table}
 \caption{E-LS constants $G,F$ (Assumption~\ref{ass:lipschitz_smooth_policy}), smoothness constant $L$ and Lipschitzness constant $\Gamma$ for Gaussian and (regularized) Softmax tabular policies, where $\varphi$ is an upper bound on the euclidean norm of the feature function for the Gaussian policy, $R_{\max}$ is the maximum absolute-valued reward, $\gamma$ is the discount factor, $\sigma$ is the standard deviation of the Gaussian policy.}
 \label{tab:constant}
 \centering
 \begin{tabular}{cccc} \\
  \toprule
            & {\bf Gaussian$^*$}              & {\bf Softmax}            & {\bf Softmax with log barrier} %& {\bf Softmax with entropy} 
             \\
  \midrule
  $G^2$     & $\frac{\varphi^2}{\sigma^2}$   & $1 - \frac{1}{|\cA|}$ & \xmark$^{**}$     \\ %& \xmark  \\
  $F$       & $\frac{\varphi^2}{\sigma^2}$   & $1$                   & \xmark    \\ %& \xmark  \\
  $L$       & $\frac{2\cR_{\max}\varphi^2}{(1-\gamma)^2\sigma^2}$    & $\frac{\cR_{\max}}{(1-\gamma)^2}\left(2 - \frac{1}{|\cA|}\right)$ 
  & $\frac{\cR_{\max}}{(1-\gamma)^2}\left(2 - \frac{1}{|\cA|}\right) + \frac{\lambda}{|\cS|}$    \\ %&  \\
  $\Gamma$  & $\frac{\cR_{\max}\varphi}{(1-\gamma)^{3/2}\sigma}$       & $\frac{\cR_{\max}}{(1-\gamma)^{3/2}}\sqrt{1 - \frac{1}{|\cA|}}$
  & $\sqrt{2\left(1-\frac{1}{|\cA|}\right)\left(\frac{\cR_{\max}^2}{(1-\gamma)^3} + \frac{\lambda^2}{|\cS|}\right)}$   \\ %&  \\
  \bottomrule
 \end{tabular} \\
 {$^{*}$\footnotesize The (\nameref{E-LS}) constants $G^2$ and $F$ are provided in Lemma 23 in~\citet{papini2019smoothing}.} \\ 
 {$^{**}$\footnotesize When there is a ``\xmark'', it means this is not applicable directly in such setting.}
\end{table}

\paragraph*{Smoothness constant.}
The smoothness constant $L = \frac{\cR_{\max}}{(1-\gamma)^2}\left(G^2+F\right)$ in~\eqref{eq:LsmoothJ} provided in Lemma~\ref{lem:smoothJ} is tighter as compared to~\citet[][Lemma 6]{papini2019smoothing} under Assumption~(\nameref{ass:lipschitz_smooth_policy}), and is also tighter as compared to~\citet[][Proposition~4.2~(2)]{xu2020sample} and~\citet[][Lemma B.1]{liu2020animproved} under more restrictive assumptions~\eqref{eq:lipschitz_smooth_policy}. Compared to existing bounds, our result shows that when $\gamma$ is close to 1, the smoothness constant~\eqref{eq:LsmoothJ} depends on $(1-\gamma)^{-2}$ instead of $(1-\gamma)^{-3}$ as derived in~\citet{papini2019smoothing},~\citet{xu2020sample} and~\citet{liu2020animproved}.
Consequently, the smoothness constant for softmax derived in Lemma~\ref{lem:softmax_smoothness} and~\ref{lem:smooth:log} are also tighter than the one derived in Lemma 7 in~\citet{mei2020ontheglobal} and Lemma D.2 in~\citet{agarwal2021theory}, which both have the dependency of $(1-\gamma)^{-3}$.
Finally, compared to the smoothness constant in~\citet{shen2019hessian} and~\citet{xu2020animproved}, our result is independent to the horizon $H$
%, same as in~\citep{papini2019smoothing} and~\citep{xu2020sample}
.

Recent works, such as Proposition~1 in~\citet{huang2020momentum}
%, Lemma B.1 in~\citet{liu2020animproved}
%\footnote{After publishing this paper, thanks to Yanli Liu who pointed out that Lemma B.1 in~\citet{liu2020animproved} is correct.} 
and equation (17) in~\citet{yuan2020stochastic}, have the dependency of $(1-\gamma)^{-2}$ for the smoothness constant under assumptions~\eqref{eq:lipschitz_smooth_policy}. However, this is due to a recurring mistake in a crucial step in bounding the Hessian.\footnote{In a previous version of the proof in Sect.~C, \citet{xu2020sample} rely on the identity $\nabla_\theta^2J(\theta) = \EE{\tau}{\nabla_\theta g(\tau\mid\theta)}$, which is  incorrect since the operators $\nabla_\theta$ and $\E{\cdot}$ are not commutative in this case as the density $p(\cdot\mid\theta)$ of $\E{\cdot}$ depends on $\theta$ as well. This error is recently fixed by~\citet{xu2020sample} on \url{https://arxiv.org/pdf/1909.08610.pdf} in their original paper.} 

%Thus, we obtain the tighter smoothness constant by establishing the tightest upper bound of the Hessian $\nabla^2 J(\cdot)$ that we are aware of. It is worth noting that~\citet[][Lemma B.1]{liu2020animproved} has a slightly better smoothness constant $\frac{F\cR_{\max}}{(1-\gamma)^2}$ than ours $\frac{\cR_{\max}}{(1-\gamma)^2}\left(G^2+F\right)$ with the stronger assumption~\eqref{eq:lipschitz_smooth_policy}. Their proof does not involve the Hessian, but aims to prove directly the following inequality
%\[\norm{\nabla J(\theta) - \nabla J(\theta')}^2 \leq \left(\frac{F\cR_{\max}}{(1-\gamma)^2}\right)^2\norm{\theta-\theta'}^2,\]
%with any $\theta, \theta' \in \R^d$. We refer to Sec.~B.1 in~\citet{liu2020animproved} for the proof of their Lemma B.1 for more details.

\paragraph*{Lipschitzness constant.}
The improved Lipschitzness constant under Assumption~(\nameref{ass:lipschitz_smooth_policy}) is provided in Lemma~\ref{lem:lipschitzJ}~\ref{itm:J-lipschitz} in Section~\ref{sec:lipschitz}. 
Compared to the existing bounds, our result shows that when $\gamma$ is close to 1, the Lipschitzness constant $\Gamma$ depends on $(1-\gamma)^{-3/2}$ instead of $(1-\gamma)^{-2}$ derived in the proof of Lemma 6 in~\citet{papini2019smoothing} under the same Assumption~(\nameref{ass:lipschitz_smooth_policy}).

\paragraph*{Upper bound of the variance of the gradient estimators.}
As for the result in Lemma~\ref{lem:ABC}, our bounds~\eqref{eq:vargrad} on the variance of the gradient estimators REINFORCE and GPOMDP are slightly tighter than the one in Lemma 17 and 18 in~\citet{papini2019smoothing}, see more details in Section~\ref{sec:proof_lem:ABC}.~\citet{shen2019hessian} and~\citet{pham2020hybrid} also showed that the variance of the vanilla gradient estimator with batch size $m=1$ is bounded under more restrictive assumptions~\eqref{eq:lipschitz_smooth_policy}. While their bounded variance depends on $(1-\gamma)^{-4}$ and they only consider the GPOMDP gradient estimator, ours~\eqref{eq:vargrad} depends on $(1-\gamma)^{-3}$ for GPOMDP or $\frac{H}{(1-\gamma)^2}$ for REINFORCE which is tighter in both cases.

\subsection{Future work}

%\rui{Mention that the empirical results will be the future direction; point the LQR case as well.}

%\begin{itemize}
%
%	\item Extend the work to projected PG
%	\item Investigate the (weak) PL in softmax with or without entropy regularization for stochastic PG updates
%	\item Investigate the ``almost'' smoothness property of LQR in our general sample complexity analysis
%	\item Identify a much broader range of problems that have not been previously considered and that satisfy~\eqref{eq:ABC} assumption
%	\item Consider variance reduced PG, actor-critic or natural PG type algorithms
%
%\end{itemize}

%As mentioned in Section~\ref{sec:discussion}, 
%\ale{Rephrase this paragraph. It is not clear if this is something we can already do or it's an open question....}
%\rui{It is an open question and I add a potential consequence of that question.}
The main focus of this paper was the theoretical analysis of vanilla variants of the PG method. 
%Our paper is indeed a theoretical investigation of the properties of PG. We agree with the reviewer that 
The results we have obtained open up several experimental questions related to parameter settings for PG. We leave such questions as an important future work to further support our theoretical findings.  %of equal length.
%an important future work is to design an extensive empirical campaign to further support our theoretical findings. 
% This paper mainly consider the vanilla PG.

One natural open question is whether the ABC assumption and the associated analysis can be extended to the projected PG. 
%thus will include the case of direct parametrized policy and other families of policy space. 
If the answer is positive, this might improve the sample complexity analysis of the direct policy parameterization setting in the stochastic case. Indeed, knowing that the direct policy parameterization satisfies a variant of~\eqref{eq:weak} condition~\citep{agarwal2021theory,xiao2022convergence} under the proximal framework, if the ABC assumption and the associated analysis can be extended, from Theorem~\ref{pro:weak} which also uses the~\eqref{eq:weak} condition, then it might be possible to establish the $\widetilde{\cO}(\epsilon^{-3})$ sample complexity 
%\rui{We obtained the $O(\epsilon^{-3})$ sample complexity by using the bijection assumption. This result is not publicly available. }
for the global optimum convergence for the direct policy parameterization and allow for a wider range of hyperparameter choices.

Similarly, we wonder if the ABC assumption and the associated analysis can be extended to the LQR setting. The challenge here will be the smoothness assumption and whether the ABC assumption is satisfied by the LQR when doing the stochastic PG updates. Indeed, the LQR only has an ``almost'' smoothness property~\citep{fazel2018global}. One needs to investigate how this will affect the current ABC analysis by extending the smoothness property to the ``almost'' smoothness property.

Recently, variance reduced methods used to decrease the variance of SGD, such as SVRG~\citep{rie2013accelerating}, SARAH~\citep{nguyen2017sarah}, SPIDER~\citep{Spider}, STORM~\citep{cutkosky2019momentum}
 and more~\citep{tran-dinh2019hybrid},
have been applied to PG methods, such as SVRPG~\citep{svrpg}, SRVR-PG~\citep{xu2020sample}, STORM-PG~\citep{yuan2020stochastic}, ProxHSPGA~\citep{pham2020hybrid}, VRMPO~\citep{yang2021policy} and VR-BGPO~\citep{huang2022bregman}. Leveraging these variance reduction techniques has led to an overall improved sample complexity of reaching a first-order stationary point (FOSP). 
%% Rui's version. Rob found it hard to parse: 
%To improve sample efficiency,~\citet{svrpg,xu2020sample,yuan2020stochastic,pham2020hybrid} introduce stochastic variance reduced gradient techniques~\citep{rie2013accelerating,nguyen2017sarah,Spider,cutkosky2019momentum,tran-dinh2019hybrid} to policy optimization, and they have studied the sample complexity of PG methods to achieve a first-order stationary point (FOSP).
However, all these works require either the exact full gradient updates or large batch sizes per iteration. It is interesting to understand whether the ABC assumption analysis can be applied to these algorithms and possibly allow for a wider range of hyperparameter choices, including the batch size. Furthermore, when the gradient domination type assumptions are available, it will be interesting to see if we can obtain faster sample complexity as we did for the vanilla PG.

Another interesting venue of investigation might be whether the ABC assumption analysis can be extended to the sample complexity analysis of (natural) actor-critic~\citep{yang2019provably,kumar2021sample,xu2020improving} or natural policy gradient algorithms~\citep{agarwal2021theory,liu2020animproved,Wang2020Neural}.

Finally we believe that the generality of Theorem~\ref{pro:ABC} opens the possibility to identify a broader set of configurations (i.e., MDP and policy space) for which PG is guaranteed to converge, notably thinking about settings such that the constant $A$ in Assumption~\eqref{eq:ABC} is \emph{non-zero}, using additional assumptions such as the bijection assumptions based on the occupancy measure space~\citep{zhang2020variational} to not only get improved sample complexity for the global optimum convergence, but also allow a wider range of hyperparameter choices for the convergence.

\section{Auxiliary Lemmas} \label{sec:auxiliary}

%\begin{lemma} \label{lem:youngs}
%For all $v,w \in \R^d$ and positive scalar $a>0$ we have that
%\[2|\dotprod{v,w}|  \leq a \norm{v}^2 + \frac{1}{a}\norm{w}^2.\]
%\end{lemma}
%\begin{proof}
%The proof follows by expanding the squares of $\norm{a v + \frac{1}{a}w}^2 \geq 0$ and  $\norm{a v - \frac{1}{a}w}^2 \geq 0.$
%\end{proof}

\begin{lemma} \label{lem:sum_of_gamma}
For all $\gamma \in [0,1)$ and any strictly positive integer $H$, we have that
\[\sum_{t=0}^{H-1}(t+1)\gamma^t \; \leq \; \sum_{t=0}^{\infty}(t+1)\gamma^t \; = \; \frac{1}{(1-\gamma)^2}.\]
\end{lemma}
\begin{proof}
The first part of the inequality is trivial. We now prove the second part of the inequality. Let
\begin{eqnarray*}
S &\eqdef& \sum_{t=0}^{\infty}(t+1)\gamma^t.
\end{eqnarray*}
We have
\begin{eqnarray*}
\gamma S \; = \; \sum_{t=0}^{\infty}(t+1)\gamma^{t+1} \; = \; \sum_{t=1}^{\infty}t\gamma^t.
\end{eqnarray*}
Subtracting of the above two equations gives
\begin{eqnarray*}
(1-\gamma)S \; = \; \sum_{t=0}^{\infty}(t+1)\gamma^t - \sum_{t=1}^{\infty}t\gamma^t
\; = \; 1 + \sum_{t=1}^{\infty}(t+1-t)\gamma^t
\; = \; \sum_{t=0}^{\infty}\gamma^t
\; = \; \frac{1}{1-\gamma}.
\end{eqnarray*}
Finally, the proof follows by dividing $1-\gamma$ on both hand side.
\end{proof}

\begin{lemma} \label{lem:sum_of_gamma2}
For all $\gamma \in [0,1)$ and any strictly positive integer $H$, we have that
\[\sum_{t=0}^{\infty}(t+1)^2\gamma^t \; \leq \; \frac{2}{(1-\gamma)^3}.\]
\end{lemma}

\begin{proof}
Let
\begin{eqnarray*}
S &\eqdef& \sum_{t=0}^{\infty}(t+1)^2\gamma^t.
\end{eqnarray*}
We have
\begin{eqnarray*}
\gamma S \; = \; \sum_{t=0}^{\infty}(t+1)^2\gamma^{t+1} \; = \; \sum_{t=1}^{\infty}t^2\gamma^t.
\end{eqnarray*}
Thus, the subtraction of the above two equations gives
\begin{eqnarray*}
(1-\gamma)S &=& \sum_{t=0}^{\infty}(t+1)^2\gamma^t - \sum_{t=1}^{\infty}t^2\gamma^t \nonumber \\
&=& 1 + \sum_{t=1}^{\infty}((t+1)^2-t^2)\gamma^t \nonumber \\
&=& 1 + \sum_{t=1}^{\infty}(2t+1)\gamma^t \nonumber \\
&=& \sum_{t=0}^{\infty}(2t+1)\gamma^t \nonumber \\
&=& 2\sum_{t=0}^{\infty}(t+1)\gamma^t - \sum_{t=0}^{\infty}\gamma^t \nonumber \\
&\overset{\mbox{Lemma~\ref{lem:sum_of_gamma}}}{=}& \frac{2}{(1-\gamma)^2} - \frac{1}{1-\gamma} \nonumber \\
&\leq& \frac{2}{(1-\gamma)^2}.
\end{eqnarray*}
Finally, the proof follows by dividing $1-\gamma$ on both hand side.
\end{proof}

\begin{lemma} \label{lem:vanilla_PG_derivation}
The full policy gradient \eqref{eq:GD} can be re-written as the following expressions
\begin{align}
\nabla J(\theta) &= \EE{\tau}{\sum_{k=0}^\infty\gamma^k\cR(s_k, a_k)\sum_{t=0}^\infty \nabla_{\theta}\log\pi_{\theta}(a_t \mid s_t)} \nonumber \\
&= \EE{\tau}{\sum_{t=0}^\infty\left(\sum_{k=0}^t\nabla_{\theta}\log\pi_{\theta}(a_k \mid s_k)\right)\gamma^t\cR(s_t, a_t)} \nonumber \\
&= \EE{\tau}{\sum_{t=0}^\infty\nabla_{\theta}\log\pi_{\theta}(a_t \mid s_t)\sum_{t'=t}^\infty\gamma^{t'}\cR(s_{t'},a_{t'})}. \label{eq:GD2}
\end{align}
\end{lemma}

\begin{proof}
To simplify \eqref{eq:GD}, we notice that future actions do not depend on past rewards. That is, for $0 \leq k < l$ among terms of the two sums in equation \eqref{eq:GD}, we have
\begin{align*}
\EE{\tau}{\nabla_{\theta}\log\pi_{\theta}(a_l \mid s_l)\gamma^k\cR(s_k, a_k)} 
&= \EE{s_{0:l}, a_{0:l}}{\nabla_{\theta}\log\pi_{\theta}(a_l \mid s_l)\gamma^k\cR(s_k, a_k)} \nonumber \\
&= \EE{s_{0:l}, a_{0:(l-1)}}{\gamma^k\cR(s_k, a_k)\EE{a_l}{\nabla_{\theta}\log\pi_{\theta}(a_l \mid s_l) \ \bigg| \ s_{0:l}, a_{0:(l-1)}}} \nonumber \\
&= \EE{s_{0:l}, a_{0:(l-1)}}{\gamma^k\cR(s_k, a_k) \int \pi_{\theta}(a_l \mid s_l)\nabla_{\theta}\log\pi_{\theta}(a_l \mid s_l) \ da_l} \nonumber \\
&= \EE{s_{0:l}, a_{0:(l-1)}}{\gamma^k\cR(s_k, a_k) \int \nabla_{\theta}\pi_{\theta}(a_l \mid s_l) \ da_l} \nonumber \\
&= \EE{s_{0:l}, a_{0:(l-1)}}{\gamma^k\cR(s_k, a_k) \nabla_{\theta}\underbrace{\int\pi_{\theta}(a_l \mid s_l) \ da_l}_{=1}} = 0.
\end{align*}
Plugging the above property into \eqref{eq:GD} yields the lemma's claim.
\end{proof}

\paragraph*{Remark.} Equation~\eqref{eq:GD2} is known as the policy gradient theorem (PGT)~\citep{sutton2000policy}.
From~\eqref{eq:GD2}, one can suggest the gradient estimator $\hnabla_m J(\theta)$ as
\begin{align} \label{eq:PGT}
\hnabla_m J(\theta) \;=\; \frac{1}{m}\sum_{i=1}^m\sum_{t=0}^{H-1}\nabla_{\theta}\log\pi_{\theta}(a_t^i \mid s_t^i) \cdot \sum_{t'=t}^{H-1}\gamma^{t'}\cR(s_{t'}^i,a_{t'}^i),
\end{align}
It has been shown by~\citet{peters2008reinforcement} that PGT~\eqref{eq:PGT} is equivalent to GPOMDP~\eqref{eq:GPOMDP}.

\begin{lemma} \label{lem:E_tau_nabla}
Under Assumption~\ref{ass:lipschitz_smooth_policy}, for all non negative integer $t$ and any state-action pair $(s_t, a_t) \in \cS \times \cA$ at time $t$ of a trajectory $\tau\sim p(\cdot\mid\theta)$ sampled under the parametrized policy $\pi_\theta$, we have that
\begin{eqnarray}
\EE{\tau\sim p(\cdot\mid\theta)}{\norm{\nabla_\theta\log\pi_\theta(a_t \mid s_t)}^2} &\leq& G^2, \label{eq:E_tau_G2} \\
\EE{\tau\sim p(\cdot\mid\theta)}{\norm{\nabla^2_\theta\log\pi_\theta(a_t \mid s_t)}} &\leq& F. \label{eq:E_tau_F}
\end{eqnarray}
\end{lemma}

\begin{proof}
For $t > 0$ and $(s_t, a_t) \in \cS \times \cA$, we have
\begin{align*} 
\EE{\tau}{\norm{\nabla_{\theta}\log\pi_{\theta}(a_t \mid s_t)}^2}
\; = \; \EE{s_t}{\EE{a_t\sim \pi_\theta(\cdot\mid s_t)}{\norm{\nabla_{\theta}\log\pi_{\theta}(a_t \mid s_t)}^2 \big | s_t}}
\; \overset{\eqref{eq:G2}}{\leq} \; G^2,
\end{align*}
where the first equality is obtained by the Markov property.

Similarly, we have
\begin{align*} 
\EE{\tau}{\norm{\nabla_{\theta}^2\log\pi_{\theta}(a_t \mid s_t)}}
\; = \; \EE{s_t}{\EE{a_t\sim \pi_\theta(\cdot\mid s_t)}{\norm{\nabla_{\theta}^2\log\pi_{\theta}(a_t \mid s_t)} \big | s_t}}
\; \overset{\eqref{eq:F}}{\leq} \; F.
\end{align*}
\end{proof}

\begin{lemma} \label{lem:cross}
For all non negative integers $0 \leq h < h'$, and any state-action pairs $(s_h, a_h), (s_{h'}, a_{h'}) \in \cS \times \cA$ at time $h$ and $h'$ respectively of the same trajectory $\tau\sim p(\cdot\mid\theta)$ sampled under the parametrized policy $\pi_\theta$, we have
\begin{eqnarray}
\EE{\tau}{\left(\nabla_{\theta}\log\pi_{\theta}(a_h \mid s_h)\right)^\top\nabla_{\theta}\log\pi_{\theta}(a_{h'} \mid s_{h'})} &=& 0. \label{eq:cross}
\end{eqnarray}
\end{lemma}

\begin{proof}
For $0 \leq h < h'$, we have
\begin{align*}
&\quad \ \EE{\tau}{\left(\nabla_{\theta}\log\pi_{\theta}(a_h \mid s_h)\right)^\top\nabla_{\theta}\log\pi_{\theta}(a_{h'} \mid s_{h'})} \nonumber \\
&= \EE{a_h, s_h, s_{h'}}{\EE{a_{h'}}{\left(\nabla_{\theta}\log\pi_{\theta}(a_h \mid s_h)\right)^\top\nabla_{\theta}\log\pi_{\theta}(a_{h'} \mid s_{h'}) \bigg | s_h, a_h, s_{h'}}} \nonumber \\
&= \EE{a_h, s_h, s_{h'}}{\left(\nabla_{\theta}\log\pi_{\theta}(a_h \mid s_h)\right)^\top\EE{a_{h'}}{\nabla_{\theta}\log\pi_{\theta}(a_{h'} \mid s_{h'}) \bigg | s_h, a_h, s_{h'}}} \nonumber \\ 
&= \EE{a_h, s_h, s_{h'}}{\left(\nabla_{\theta}\log\pi_{\theta}(a_h \mid s_h)\right)^\top\int_{a_{h'}}\pi_{\theta}(a_{h'} \mid s_{h'})\nabla_{\theta}\log\pi_{\theta}(a_{h'} \mid s_{h'})da_{h'}} \nonumber \\
&= \EE{a_h, s_h, s_{h'}}{\left(\nabla_{\theta}\log\pi_{\theta}(a_h \mid s_h)\right)^\top\int_{a_{h'}}\nabla_{\theta}\pi_{\theta}(a_{h'} \mid s_{h'})da_{h'}} \nonumber \\ 
&= \mathbb{E}_{a_h, s_h, s_{h'}}\bigg [\left(\nabla_{\theta}\log\pi_{\theta}(a_h \mid s_h)\right)^\top\nabla_{\theta}\underbrace{\int_{a_{h'}}\pi_{\theta}(a_{h'} \mid s_{h'})da_{h'}}_{=1} \bigg ] = 0,
\end{align*}
where the first and second equality is obtained by the Markov property.
\end{proof}

\begin{lemma} \label{lem:Etnorm2}
For all non negative integers $0 \leq t$, and any state-action pairs $(s_h, a_h) \in \cS \times \cA$ at time $0 \leq h \leq t$ of the same trajectory $\tau\sim p(\cdot\mid\theta)$ sampled under the parametrized policy $\pi_\theta$, we have
\begin{eqnarray}
\EE{\tau}{\norm{\sum_{h=0}^t\nabla_{\theta}\log\pi_{\theta}(a_h \mid s_h)}^2} &=& \sum_{h=0}^t\EE{\tau}{\norm{\log\pi_{\theta}(a_h \mid s_h)}^2}. \label{eq:Etnorm2}
\end{eqnarray}
\end{lemma}

\begin{proof}
For $0 \leq t$, we have
\begin{eqnarray*}
 \EE{\tau}{\norm{\sum_{h=0}^t\nabla_{\theta}\log\pi_{\theta}(a_h \mid s_h)}^2} 
&=& \sum_{h=0}^t\EE{\tau}{\norm{\log\pi_{\theta}(a_h \mid s_h)}^2}  \\
&\quad& \ + \ 2\sum_{h=0}^{t-1}\sum_{h'=h+1}^t\EE{\tau}{\left(\nabla_\theta\log \pi_\theta(a_h\mid\theta_h)\right)^\top\nabla_\theta\log \pi_\theta(a_{h'}\mid\theta_{h'})} \\
&\overset{\eqref{eq:cross}}{=}& \sum_{h=0}^t\EE{\tau}{\norm{\log\pi_{\theta}(a_h \mid s_h)}^2}.
\end{eqnarray*}
\end{proof}

\section{Proof of Section~\ref{sec:ABC}}

\subsection{Proof of Theorem~\ref{pro:ABC}}
\label{sec:proof ABC}

\begin{proof}
We start with $L$-smoothness of $J$ from Assumption~\ref{ass:smooth}, which implies
\begin{eqnarray}
J(\theta_{t+1}) &\geq& J(\theta_t) + \dotprod{\nabla J(\theta_t), \theta_{t+1} - \theta_t} - \frac{L}{2}\norm{\theta_{t+1}-\theta_t}^2 \nonumber \\
&=& J(\theta_t) + \eta\dotprod{\nabla J(\theta_t), \hnabla_m J(\theta_t)} - \frac{L\eta^2}{2}\norm{\hnabla_m J(\theta_t)}^2.
\end{eqnarray}
Taking expectations conditioned on $\theta_t$, we get
\begin{eqnarray}
\EE{t}{J(\theta_{t+1})} &\geq& J(\theta_t) + \eta\dotprod{\nabla J(\theta_t), \nabla J_H(\theta_t)} - \frac{L\eta^2}{2}\EE{t}{\norm{\hnabla_m J(\theta_t)}^2} \nonumber \\
&\overset{\eqref{eq:ABC}}{\geq}& J(\theta_t) + \eta\dotprod{\nabla J_H(\theta_t) + \left(\nabla J(\theta_t) - \nabla J_H(\theta_t)\right), \nabla J_H(\theta_t)} \nonumber \\
&\quad& \ - \frac{L\eta^2}{2}\left(2A(J^*-J(\theta_t)) + B\norm{\nabla J_H(\theta_t)}^2 + C\right) \nonumber \\
&=& J(\theta_t) + \eta\left(1-\frac{LB\eta}{2}\right)\norm{\nabla J_H(\theta_t)}^2 - L\eta^2A(J^*-J(\theta_t)) \nonumber \\
&\quad& \ - \frac{LC\eta^2}{2} + \eta\dotprod{\nabla J_H(\theta_t), \nabla J(\theta_t)-\nabla J_H(\theta_t)} \nonumber \\
&\overset{\eqref{eq:trunc}}{\geq}& J(\theta_t) + \eta\left(1-\frac{LB\eta}{2}\right)\norm{\nabla J_H(\theta_t)}^2 - L\eta^2A(J^*-J(\theta_t)) - \frac{LC\eta^2}{2} - \eta D\gamma^H.
\end{eqnarray}
Subtracting $J^*$ from both sides gives
\begin{eqnarray}
-\left(J^* - \EE{t}{J(\theta_{t+1})}\right) &\geq& -(1+L\eta^2A)(J^*-J(\theta_t)) + \eta\left(1-\frac{LB\eta}{2}\right)\norm{\nabla J_H(\theta_t)}^2 - \frac{LC\eta^2}{2} - \eta D\gamma^H.
\end{eqnarray}
Taking the total expectation and rearranging, we get
\begin{align}
\E{J^* - J(\theta_{t+1})} + \eta\left(1-\frac{LB\eta}{2}\right)\E{\norm{\nabla J_H(\theta_t)}^2} \; \leq \; (1+L\eta^2A)\E{J^*-J(\theta_t)} + \frac{LC\eta^2}{2} + \eta D\gamma^H.
\end{align}
Letting $\delta_t \eqdef \E{J^* - J(\theta_t)}$ and $r_t \eqdef \E{\norm{\nabla J_H(\theta_t)}^2}$, we can rewrite the last inequality as
\begin{eqnarray} \label{eq:introduce_weight}
\eta\left(1-\frac{LB\eta}{2}\right)r_t &\leq& (1+L\eta^2A)\delta_t - \delta_{t+1} + \frac{LC\eta^2}{2} + \eta D\gamma^H.
\end{eqnarray}
We now introduce a sequence of weights $w_{-1}, w_0, w_1, \cdots, w_{T-1}$ based on a technique developed by~\citet{stich2019unified}. Let $w_{-1} > 0$. Define $w_t \eqdef \frac{w_{t-1}}{1+L\eta^2A}$ for all $t\geq0$. Notice that if $A=0$, we have $w_t=w_{t-1}=\cdots=w_{-1}$. Multiplying~\eqref{eq:introduce_weight} by $w_t/\eta$,
\begin{eqnarray}
\left(1-\frac{LB\eta}{2}\right)w_tr_t &\leq& \frac{w_t(1+L\eta^2A)}{\eta}\delta_t - \frac{w_t}{\eta}\delta_{t+1} + \frac{LC\eta}{2}w_t + D\gamma^Hw_t \nonumber \\
&=& \frac{w_{t-1}}{\eta}\delta_t - \frac{w_t}{\eta}\delta_{t+1} + \left( \frac{LC\eta}{2} + D\gamma^H \right) w_t.
\end{eqnarray}
Summing up both sides as $t=0,1,\cdots,T-1$ and using telescopic sum, we have,
\begin{eqnarray} \label{eq:wtrt}
\left(1-\frac{LB\eta}{2}\right)\sum_{t=0}^{T-1}w_tr_t &\leq& \frac{w_{-1}}{\eta}\delta_0 - \frac{w_{T-1}}{\eta}\delta_T + \left( \frac{LC\eta}{2} + D\gamma^H \right)\sum_{t=0}^{T-1}w_t \nonumber \\
&\leq& \frac{w_{-1}}{\eta}\delta_0 + \left( \frac{LC\eta}{2} + D\gamma^H \right)\sum_{t=0}^{T-1}w_t.
\end{eqnarray}
Let $W_T \eqdef \sum_{t=0}^{T-1}w_t$. Dividing both sides by $W_T$, we have,
\begin{eqnarray} \label{eq:min_rt}
\left(1-\frac{LB\eta}{2}\right)\min_{0\leq t\leq T-1}r_t \; \leq \; \frac{1}{W_T}\cdot\left(1-\frac{LB\eta}{2}\right)\sum_{t=0}^{T-1}w_tr_t \; \leq \; \frac{w_{-1}}{W_T}\frac{\delta_0}{\eta} + \frac{LC\eta}{2} + D\gamma^H.
\end{eqnarray}
Note that,
\begin{eqnarray}
W_T \; = \; \sum_{t=0}^{T-1}w_t \geq \sum_{t=0}^{T-1}\min_{0\leq i\leq T-1}w_i \; = \; Tw_{T-1} \; = \; \frac{Tw_{-1}}{(1+L\eta^2A)^T}.
\end{eqnarray}
Using this in~\eqref{eq:min_rt},
\begin{eqnarray} \label{eq:min_rt2}
\left(1-\frac{LB\eta}{2}\right)\min_{0\leq t\leq T-1}r_t &\leq& \frac{(1+L\eta^2A)^T}{\eta T}\delta_0 + \frac{LC\eta}{2} + D\gamma^H.
\end{eqnarray}
However, we have
\begin{eqnarray} \label{eq:difference}
\E{\norm{\nabla J(\theta_t)}^2} &=& \E{\norm{\nabla J(\theta_t) - \nabla J_H(\theta_t) + \nabla J_H(\theta_t)}^2} \nonumber \\
&=& \E{\norm{\nabla J_H(\theta_t)}^2} + 2\E{\dotprod{\nabla J_H(\theta_t), \nabla J(\theta_t) - \nabla J_H(\theta_t)}} + \E{\norm{\nabla J(\theta_t) - \nabla J_H(\theta_t)}^2} \nonumber \\
&\overset{\eqref{eq:trunc}+\eqref{eq:trunc2}}{\leq}& \E{\norm{\nabla J_H(\theta_t)}^2} + 2D\gamma^H + D'^2\gamma^{2H}.
\end{eqnarray}
Substituting $r_t$ in~\eqref{eq:min_rt2} by $\E{\norm{\nabla J(\theta_t)}^2}$ and using~\eqref{eq:difference}, we get
\begin{align*}
\left(1-\frac{LB\eta}{2}\right)\min_{0\leq t\leq T-1}\E{\norm{\nabla J(\theta_t)}^2} \; \leq \; \frac{(1+L\eta^2A)^T}{\eta T}\delta_0 + \frac{LC\eta}{2} + D\gamma^H + \left(1-\frac{LB\eta}{2}\right)\left(2D\gamma^H + D'^2\gamma^{2H}\right).
\end{align*}
Our choice of step size guarantees that no matter $B > 0$ or $B=0$, we have $1-\frac{LB\eta}{2}>0$.
Dividing both sides by $1-\frac{LB\eta}{2}$ and rearranging yields the theorem's claim.

If $A=0$, we know that $\{w_t\}_{t\geq-1}$ is a constant sequence. In this case, $W_T = Tw_{-1}$. Dividing both sides of~\eqref{eq:wtrt} by $W_T$, we have,
\begin{eqnarray} \label{eq:min_rt3}
\left(1-\frac{LB\eta}{2}\right)\frac{1}{T}\sum_{t=0}^{T-1}r_t \; \leq \; \frac{\delta_0}{\eta T} + \frac{LC\eta}{2} + D\gamma^H.
\end{eqnarray}
Similarly, substituting $r_t$ in~\eqref{eq:min_rt3} by $\E{\norm{\nabla J(\theta_t)}^2}$ and using~\eqref{eq:difference}, we get
\begin{eqnarray*}
\left(1-\frac{LB\eta}{2}\right)\E{\norm{\nabla J(\theta_U)}^2} &=& \left(1-\frac{LB\eta}{2}\right)\frac{1}{T}\sum_{t=0}^{T-1}\E{\norm{\nabla J(\theta_t)}^2} \\ 
&\leq& \frac{\delta_0}{\eta T} + \frac{LC\eta}{2} + D\gamma^H + \left(1-\frac{LB\eta}{2}\right)\left(2D\gamma^H + D'^2\gamma^{2H}\right).
\end{eqnarray*}
Dividing both sides by $1-\frac{LB\eta}{2}$ and rearranging yields the theorem's claim.
\end{proof}

\subsection{Proof of Corollary~\ref{cor:ABC}}

\begin{proof}
Given $\epsilon > 0$, from Corollary~1 in~\citet{khaled2020better}, we know that if $\eta = \min\big\{\frac{1}{\sqrt{LAT}}, \frac{1}{LB}, \frac{\epsilon}{2LC}\big\}$ and the number of iterations $T$ satisfies
\[
    T \; \geq \; \frac{12\delta_0L}{\epsilon^2}\max\left\{B, \frac{12\delta_0A}{\epsilon^2}, \frac{2C}{\epsilon^2}\right\},
\]
we have
\[
\frac{2\delta_0(1+L\eta^2A)^T}{\eta T(2-LB\eta)} + \frac{LC\eta}{2-LB\eta} \; \leq \; \epsilon^2.
\]
It remains to show 
\[\left(\frac{2D(3-LB\eta)}{2-LB\eta} + D'^2\gamma^{H}\right)\gamma^H \leq \epsilon^2.\]
Besides, our choice of the step size $\eta \leq \frac{1}{LB}$ implies that $\frac{1}{2 - LB\eta} \leq 1$, thus
\[\left(\frac{2D(3-LB\eta)}{2-LB\eta} + D'^2\gamma^{H}\right)\gamma^H \leq \left(6D + D'^2\gamma^{H}\right)\gamma^H.\]
Finally, it suffices to choose $H$ such that
\[\gamma^H \; \leq \; \epsilon^2 \quad \Longleftrightarrow \quad H \; \geq \; \frac{2\log\epsilon^{-1}}{\log\gamma^{-1}} \; = \; \cO(\log\epsilon^{-1}),\]
to guarantee that $\min_{0\leq t\leq T-1}\E{\norm{\nabla J(\theta_t)}^2} = \cO(\epsilon^{-2})$, which concludes the proof. 
\end{proof}

\paragraph*{Remark.} When $\gamma$ is close to $1$, the horizon has the following property.
\[H \; = \; \frac{2\log\epsilon^{-1}}{\log\gamma^{-1}} \; = \; \cO\left(\frac{\log\epsilon^{-1}}{1-\gamma}\right).\]

\paragraph*{Remark.} When $A=0$, by following the same analysis of Corollary~\ref{cor:ABC} applied to~\eqref{eq:A=0} in Theorem~\ref{pro:ABC}, choosing the parameters proposed in Corollary~\ref{cor:ABC} guarantees that $\E{\norm{\nabla J(\theta_U)}^2} = \cO(\epsilon^2)$.

\subsection{Average regret convergence under the relaxed weak gradient domination assumption}
\label{sec:regret}

When the relaxed weak gradient domination assumption~\eqref{eq:weak} is available, it is straightforward to obtain the average regret to the global optimum convergence under the setting of Corollary~\ref{cor:ABC}.

\begin{corollary} \label{cor:regret}
Suppose that Assumption~\ref{ass:smooth},~\ref{ass:trunc},~\ref{ass:ABC} and~\ref{ass:weak} hold with $A=0$.
Given  $\epsilon > 0$, let $\eta = \min\big\{\frac{1}{LB}, \frac{\epsilon}{2LC}\big\}$ and the horizon $H = \cO(\log\epsilon^{-1})$. If the number of iterations $T$ satisfies
\begin{align}
%    \eta &= \min\left\{\frac{1}{\sqrt{LAT}}, \frac{1}{LB}, \frac{\epsilon}{2LC}\right\}, \nonumber \\
    T \; \geq \; \frac{12\delta_0L}{\epsilon^2}\max\left\{B, \frac{2C}{\epsilon^2}\right\},  %\\
%    H &= \cO(\log\epsilon^{-1}), \nonumber
\end{align}
then $J^* - \frac{1}{T}\sum_{t=0}^{T-1}\E{J(\theta_t)} = \cO(\epsilon) + \cO(\epsilon')$. 
%the step size is set to $\eta = \min\{\frac{1}{\sqrt{LAT}}, \frac{1}{LB}, \frac{\epsilon}{2LC}\}$, $T \geq \frac{12\delta_0L}{\epsilon^2}\max\{B, \frac{12\delta_0A}{\epsilon^2}, \frac{2C}{\epsilon^2}\} $ and $H = \cO(\log\epsilon^{-1})$, we have $\min_{0\leq t\leq T-1}\E{\norm{\nabla J(\theta_t)}^2} \leq \epsilon$.
\end{corollary}

\begin{proof}
From the remark of the proof analysis of Corollary~\ref{cor:ABC} with $A=0$, we know that
\[\E{\norm{\nabla J(\theta_U)}^2} = \frac{1}{T}\sum_{t=0}^{T-1}\E{\norm{\nabla J(\theta_t)}^2} = \cO(\epsilon^2).\]
From Assumption~\ref{ass:trunc}, we get
\begin{align} \label{eq:average}
\frac{1}{T}\sum_{t=0}^{T-1}\E{\norm{\nabla J_H(\theta_t)}^2} = \cO(\epsilon^2).
\end{align}
Besides, from~\eqref{eq:weak}, we obtain that
\begin{align} \label{eq:weak3}
(\epsilon')^2 + \norm{\nabla J_H(\theta)}^2 \geq \frac{\left(\epsilon' + \norm{\nabla J_H(\theta)}\right)^2}{2} \geq 2\mu(J^* - J(\theta))^2.
\end{align}
Thus, by~\eqref{eq:average} and~\eqref{eq:weak3}, we have
\begin{eqnarray*}
(\epsilon')^2 + \frac{1}{T}\sum_{t=0}^{T-1}\E{\norm{\nabla J_H(\theta_t)}^2} 
&\overset{\eqref{eq:average}}{=}& (\epsilon')^2 + \cO(\epsilon^2) 
\; \overset{\eqref{eq:weak3}}{\geq} \; \frac{2\mu}{T}\sum_{t=0}^{T-1}\E{(J^* - J(\theta_t))^2} \\ 
&\geq& 2\mu \E{\left(J^* - \frac{1}{T}\sum_{t=0}^{T-1} J(\theta_t)\right)^2}
\; \ge \; 2\mu \left(J^* - \frac{1}{T}\sum_{t=0}^{T-1} \E{J(\theta_t)}\right)^2,
\end{eqnarray*}
where the last two inequalities are obtained by applying Jensen inequality twice.
By using $(a+b)^2 \geq a^2 + b^2$ with $a, b \geq 0$, we conclude that $J^* - \frac{1}{T}\sum_{t=0}^{T-1}\E{J(\theta_t)} = \cO(\epsilon) + \cO(\epsilon')$.
\end{proof}

\subsection{Global optimum convergence under the relaxed weak gradient domination assumption}
\label{sec:pro_weak}

In this section, we present the new global optimum convergence theory under the relaxed weak gradient domination assumption~\eqref{eq:weak}.

\begin{theorem} \label{pro:weak}
Suppose that Assumption~\ref{ass:smooth},~\ref{ass:trunc},~\ref{ass:ABC} and~\ref{ass:weak} hold. Given $\epsilon > 0$, define $\delta$ s.t.\ if $\epsilon' = 0$, set $\delta = \epsilon$, if $\epsilon' > 0$, set $\delta = \epsilon'$. Suppose that PG defined in~\eqref{eq:GA} is run for $T>0$ iterations with step size $(\eta_t)_t$ chosen as
\begin{eqnarray}
\eta_t =
\begin{cases}
\frac{1}{b} \quad & \mbox{if } T \leq \frac{b}{\mu\delta} \; \mbox{ or } \; t \leq t_0 \\
\frac{2}{2b + \mu\delta(t-t_0)} \quad & \mbox{if } T \geq \frac{b}{\mu\delta} \; \mbox{ and } \; t > t_0
\end{cases}
\end{eqnarray}
with $t_0 = \left[\frac{T}{2}\right]$ and $b = \max\{\frac{2AL}{\mu\delta}, 2BL, \mu\delta\}$. If $J^* - \E{J(\theta_t)} \geq \delta$ for all $t \in \{0, 1, \cdots, T-1\}$, then
\begin{align} \label{eq:weak_bound}
J^* - \E{J(\theta_T)} \leq 16\exp\left(-\frac{\mu\delta(T-1)}{2b}\right)(J^* - J(\theta_0))  + \frac{12LC}{\mu^2\delta^2T} + \frac{26D\gamma^H}{\mu\delta} + \frac{12(\epsilon')^2(2b-LB)}{\mu^2\delta^2T} + \frac{2\epsilon'}{\mu},
\end{align}
otherwise, we have
\[\min\limits_{t \in \{0, 1, \cdots, T-1\}} J^* - \E{J(\theta_t)} \leq \delta.\]
\end{theorem}
%\rui{If we set $\delta = \max\{\epsilon, \epsilon' \}$, we are not favoured when $\delta = \epsilon' < \epsilon$, given the fact that $\min\limits_{t \in \{0, 1, \cdots, T-1\}} J^* - \E{J(\theta_t)} \leq \delta.$}

\paragraph*{Remark.} Similar to the exact full gradient update in Thm.~\ref{pro:ABC}, notice that for the exact full gradient update, we have Asm.~\ref{ass:trunc} and~\ref{ass:ABC} hold with $A=C=D=0$ and $B=1$. Thus under the smoothness and the weak gradient domination assumption (i.e., $\epsilon' = 0$), we have
\begin{eqnarray*}
J^* - \E{J(\theta_T)} \leq 16\exp\left(-\frac{\mu\epsilon(T-1)}{2b}\right)(J^* - J(\theta_0)).
\end{eqnarray*}
With $T = \frac{1}{\epsilon}\log\left(\frac{1}{\epsilon}\right)$, we have $J^* - \E{J(\theta_T)} \leq \epsilon$. Thus
we establish $\widetilde{\cO}(\epsilon^{-1})$ convergence rate for the number of iterations to the global optimal. We recover the same rate for the softmax tabular policy in Theorem 4 in~\citet{mei2020ontheglobal} where the smoothness assumption holds and the weak gradient domination condition~\eqref{eq:weak} holds on the path of the iterates in the exact case.

%In this section, we extend the global optimum convergence guarantee of~\citet{agarwal2021theory,mei2020ontheglobal} from the exact policy gradient to the stochastic vanilla PG. We establish its sample complexity analysis with the weak gradient domination assumption.

%\rob{I didn't understand this sentence, so I commented it out?}
% This condition is in contrast to the gradient domination (Asm.~\ref{ass:PL}). 

%\rui{A proposition for the convergence rate and the sample complexity of the average regret with the global optimum.}
%\rui{
%This result may be used for projected PG applied in direct parametrized policy. However, for projected PG, the weak gradient 
%domination property is different, it should be 
%\[G_{1/\eta}(\theta) \; \geq \; \sqrt{2\mu}(J^* - J(\theta^+)).\]
%It is not available for softmax with log barrier regularization.
%}

\begin{proof}
From~\eqref{eq:weak}, we obtain that
\begin{align} \label{eq:weak2}
&\quad \quad (\epsilon')^2 + \norm{\nabla J_H(\theta)}^2 \geq \frac{\left(\epsilon' + \norm{\nabla J_H(\theta)}\right)^2}{2} \geq 2\mu(J^* - J(\theta))^2 \nonumber \\ 
&\Longrightarrow \norm{\nabla J_H(\theta)}^2 \geq 2\mu(J^* - J(\theta))^2 - (\epsilon')^2.
\end{align}

Let $t \in \{0, 1, \cdots, T-1\}$.
Using the $L$-smoothness of $J$ from Assumption~\ref{ass:smooth},
\begin{eqnarray}
J^* - J(\theta_{t+1}) &\leq& J^* - J(\theta_t) - \dotprod{\nabla J(\theta_t), \theta_{t+1}-\theta_t} + \frac{L}{2}\norm{\theta_{t+1}-\theta_t}^2 \nonumber \\ 
&=& J^* - J(\theta_t) - \eta_t\dotprod{\nabla J(\theta_t), \hnabla_m J(\theta_t)} + \frac{L\eta_t^2}{2}\norm{\hnabla_m J(\theta_t)}^2.
\end{eqnarray}
Taking expectation conditioned on $\theta_t$ and using Assumption~\ref{ass:ABC} and~\ref{ass:weak},
\begin{eqnarray}
\EE{t}{J^* - J(\theta_{t+1})} &\leq& J^* - J(\theta_t) - \eta_t\dotprod{\nabla J(\theta_t), \nabla J_H(\theta_t)} + \frac{L\eta_t^2}{2}\EE{t}{\norm{\hnabla_m J(\theta_t)}^2} \nonumber \\ 
&\overset{\eqref{eq:ABC}}{\leq}& J^* - J(\theta_t) - \eta_t\dotprod{\nabla J_H(\theta_t) + (\nabla J(\theta_t) - \nabla J_H(\theta_t)), \nabla J_H(\theta_t)} + \nonumber \\ 
&\quad& \ + \frac{L\eta_t^2}{2}\left(2A(J^* - J(\theta_t)) + B\norm{\nabla J_H(\theta_t)}^2 +C\right) \nonumber \\ 
&=& (1+L\eta_t^2A)(J^* - J(\theta_t)) - \eta_t\left(1-\frac{LB\eta_t}{2}\right)\norm{\nabla J_H(\theta_t)}^2 + \frac{L\eta_t^2C}{2} \nonumber \\ 
&\quad& \ - \eta_t\dotprod{\nabla J(\theta_t) - \nabla J_H(\theta_t), \nabla J_H(\theta_t)} \nonumber \\ 
&\overset{\eqref{eq:weak2}}{\leq}& \left(1 + L\eta_t^2A\right)(J^* - J(\theta_t)) - \mu\eta_t\left(2 - LB\eta_t\right)(J^* - J(\theta_t))^2 + \eta_t\left(1 - \frac{LB\eta_t}{2}\right)(\epsilon')^2 \nonumber \\
&\quad& \  + \frac{L\eta_t^2C}{2} - \eta_t\dotprod{\nabla J(\theta_t) - \nabla J_H(\theta_t), \nabla J_H(\theta_t)} \nonumber \\ 
&\overset{\eqref{eq:trunc}}{\leq}& \left(1 + L\eta_t^2A\right)(J^* - J(\theta_t)) - \mu\eta_t\left(2 - LB\eta_t\right)(J^* - J(\theta_t))^2 + \eta_t\left(1 - \frac{LB\eta_t}{2}\right)(\epsilon')^2 \nonumber \\ 
&\quad& \ + \frac{L\eta_t^2C}{2} + \eta_tD\gamma^H \nonumber \\ 
&\leq& \left(1 + L\eta_t^2A\right)(J^* - J(\theta_t)) - \frac{3\mu}{2}\eta_t(J^* - J(\theta_t))^2 + \eta_t\left(1 - \frac{LB\eta_t}{2}\right)(\epsilon')^2 \nonumber \\ 
&\quad& \ + \frac{L\eta_t^2C}{2} + \eta_tD\gamma^H, \label{eq:rt2}
\end{eqnarray}
where the last line is obtained by the choice of the step size $\eta_t \leq \frac{1}{b}$ with $b \geq 2LB$.

Taking total expectation and letting $r_t \eqdef \E{J^* - J(\theta_t)}$ on~\eqref{eq:rt2}, we have
\begin{eqnarray} \label{eq:rtweak}
r_{t+1} &\leq& r_t + LA\eta_t^2 r_t - \frac{3\mu}{2}\eta_tr_t^2 + \eta_t\left(1 - \frac{LB\eta_t}{2}\right)(\epsilon')^2 + \frac{LC}{2}\eta_t^2 + \eta_tD\gamma^H.
\end{eqnarray}

%\rob{I added some words here, double check.}
If there exists $t \in \{0, 1, \cdots, T-1\}$ such that $r_t < \delta$, then we are done. Alternatively
if $r_t \geq \delta$ for all $t \in \{0, 1, \cdots, T-1\}$, from~\eqref{eq:rtweak}, we have
\begin{eqnarray}
r_{t+1} &\leq& r_t + LA\eta_t^2 r_t - \frac{3\mu\delta}{2}\eta_tr_t + \eta_t\left(1 - \frac{LB\eta_t}{2}\right)(\epsilon')^2 + \frac{LC}{2}\eta_t^2 + \eta_tD\gamma^H \nonumber \\ 
&\leq& (1- \mu\delta\eta_t)r_t + \eta_t\left(1 - \frac{LB\eta_t}{2}\right)(\epsilon')^2 + \frac{LC}{2}\eta_t^2 + \eta_tD\gamma^H, \label{eq:rtw}
\end{eqnarray}
where the last line is obtained by the choice of the step size $\eta_t \leq \frac{1}{b}$ with $b \geq \frac{2LA}{\mu\delta}$. Here $1- \mu\delta\eta_t \geq 0$ as $\eta_t \leq \frac{1}{b}$ with $b \geq \mu\delta$. We notice that~\eqref{eq:rtw} is similar to~\eqref{eq:recurse}. The rest of the proof is similar to the one of Theorem~\ref{pro:PL}.

If $T \leq \frac{b}{\mu\delta}$, $\eta_t = \frac{1}{b}$. From~\eqref{eq:rtw}, we have
\begin{eqnarray}
r_T &\leq& \left(1 - \frac{\mu\delta}{b}\right)r_{T-1} + \frac{LC}{2b^2} + \frac{D\gamma^H}{b} + \frac{2b - LB}{2b^2} (\epsilon')^2 \nonumber \\ 
&\overset{\eqref{eq:rtw}}{\leq}& \left(1-\frac{\mu\delta}{b}\right)^Tr_0 + \left(\frac{LC}{2b^2} + \frac{D\gamma^H}{b} + \frac{2b - LB}{2b^2} (\epsilon')^2\right)\sum_{i=0}^{T-1}\left(1-\frac{\mu\delta}{b}\right)^i \nonumber \\ 
&\leq& \exp\left(-\frac{\mu\delta T}{b}\right)r_0 + \frac{LC}{2\mu\delta b} + \frac{D\gamma^H}{\mu\delta} + \frac{2b - LB}{2\mu \delta b}(\epsilon')^2 \label{eq:T<w} \\ 
&\overset{T \leq \frac{b}{\mu\delta}}{\leq}& \exp\left(-\frac{\mu\delta T}{b}\right)r_0 + \frac{LC}{2\mu^2\delta^2T} + \frac{D\gamma^H}{\mu\delta} + \frac{2b - LB}{2\mu^2\delta^2 T}(\epsilon')^2. \label{eq:T<w2}
\end{eqnarray}

If $T \geq \frac{b}{\mu\delta}$, as $\eta_t = \frac{1}{b}$ when $t \leq t_0$, from~\eqref{eq:T<w}, we have
\begin{eqnarray}
r_{t_0} &\leq& \exp\left(-\frac{\mu\delta t_0}{b}\right)r_0 + \frac{LC}{2\mu\delta b} + \frac{D\gamma^H}{\mu\delta} + \frac{2b - LB}{2\mu \delta b}(\epsilon')^2 \nonumber \\ 
&\leq& \exp\left(-\frac{\mu\delta(T-1)}{2b}\right)r_0 + \frac{LC}{2\mu\delta b} + \frac{D\gamma^H}{\mu\delta} + \frac{2b - LB}{2\mu \delta b}(\epsilon')^2, \label{eq:rt0}
\end{eqnarray}
where the last line is obtained by $t_0 = \left[\frac{T}{2}\right] \geq \frac{T-1}{2}$.

For $t > t_0$,
\[\eta_t \; = \; \frac{2}{\mu\delta\left(\frac{2b}{\mu\delta} + t - t_0\right)}.\]
From~\eqref{eq:rtw}, we have
\begin{eqnarray}
r_t &\leq& \frac{\frac{2b}{\mu\delta} + t - t_0 - 2}{\frac{2b}{\mu\delta} + t - t_0}r_{t-1} + \frac{2LC}{\mu^2\delta^2\left(\frac{2b}{\mu\delta} + t - t_0\right)^2} + \frac{2D\gamma^H}{\mu\delta\left(\frac{2b}{\mu\delta} + t - t_0\right)} \nonumber \\ 
&\quad& \ + \frac{2(\epsilon')^2}{\mu\delta\left(\frac{2b}{\mu\delta} + t - t_0\right)}\left(1 - \frac{LB}{\mu\delta\left(\frac{2b}{\mu\delta} + t - t_0\right)}\right).
\end{eqnarray}
Multiplying both sides by $\left(\frac{2b}{\mu\delta} + t - t_0\right)^2$, we have
\begin{align}
\left(\frac{2b}{\mu\delta} + t - t_0\right)^2r_t &\leq \left(\frac{2b}{\mu\delta} + t - t_0\right)\left(\frac{2b}{\mu\delta} + t - t_0 - 2\right) r_{t-1} + \frac{2LC}{\mu^2\delta^2} + \frac{2D\gamma^H}{\mu\delta}\left(\frac{2b}{\mu\delta} + t - t_0\right) \nonumber \\ 
&\quad \ + \frac{2(\epsilon')^2}{\mu\delta}\left(\frac{2b-LB}{\mu\delta}+t-t_0\right) \nonumber \\ 
&\leq \left(\frac{2b}{\mu\delta} + t - t_0 - 1\right)^2 r_{t-1} + \frac{2LC}{\mu^2\delta^2} + \frac{2D\gamma^H}{\mu\delta}\left(\frac{2b}{\mu\delta} + t - t_0\right) \nonumber \\ 
&\quad \ + \frac{2(\epsilon')^2}{\mu\delta}\left(\frac{2b-LB}{\mu\delta}+t-t_0\right).
\end{align}
Let $w_t \eqdef \left(\frac{2b}{\mu\epsilon} + t - t_0\right)^2$. We have
\begin{eqnarray}
w_tr_t &\leq& w_{t-1}r_{t-1} + \frac{2LC}{\mu^2\delta^2} + \frac{2D\gamma^H}{\mu\delta}\left(\frac{2b}{\mu\delta} + t - t_0\right) + \frac{2(\epsilon')^2}{\mu\delta}\left(\frac{2b-LB}{\mu\delta}+t-t_0\right).
\end{eqnarray}
Summing up for $t = t_0 + 1, \cdots, T$ and telescoping, we get,
\begin{align}
w_Tr_T &\leq w_{t_0}r_{t_0} + \frac{2LC(T-t_0)}{\mu^2\delta^2} + \frac{2D\gamma^H}{\mu\delta}\sum_{t=t_0+1}^{T}\left(\frac{2b}{\mu\delta} + t - t_0\right) + \frac{2(\epsilon')^2}{\mu\delta}\sum_{t=t_0+1}^{T}\left(\frac{2b-LB}{\mu\delta}+t-t_0\right) \nonumber \\ 
&= \frac{4b^2}{\mu^2\delta^2}r_{t_0} + \frac{2LC(T-t_0)}{\mu^2\delta^2} + \frac{4bD(T-t_0)\gamma^H}{\mu^2\delta^2} + \frac{D\gamma^H}{\mu\delta}(T-t_0)(T-t_0+1) \nonumber \\ 
&\quad \ + \frac{2(\epsilon')^2(2b-LB)(T-t_0)}{\mu^2\delta^2} + \frac{(\epsilon')^2}{\mu\delta}(T-t_0)(T-t_0+1).
\end{align}
Dividing both sides by $w_T$ and using that since
\[w_T \; = \; \left(\frac{2b}{\mu\delta}+T-t_0\right)^2 \; \geq \; (T-t_0)^2,\]
we have
\begin{eqnarray}
r_T &\leq& \frac{4b^2}{\mu^2\delta^2w_T}r_{t_0} + \frac{2LC(T-t_0)}{\mu^2\delta^2w_T} + \frac{4bD(T-t_0)\gamma^H}{\mu^2\delta^2w_T} + \frac{D\gamma^H}{\mu\delta w_T}(T-t_0)(T-t_0+1) \nonumber \\ 
&\quad& \ + \frac{2(\epsilon')^2(2b-LB)(T-t_0)}{\mu^2\delta^2 w_T} + \frac{(\epsilon')^2}{\mu\delta w_T}(T-t_0)(T-t_0+1) \nonumber \\ 
&\leq& \frac{4b^2}{\mu^2\delta^2(T-t_0)^2}r_{t_0} + \frac{2LC}{\mu^2\delta^2(T-t_0)} + \frac{4bD\gamma^H}{\mu^2\delta^2(T-t_0)} + \frac{2D\gamma^H}{\mu\delta} + \frac{2(\epsilon')^2(2b-LB)}{\mu^2\delta^2(T-t_0)} + \frac{2(\epsilon')^2}{\mu\delta}.
\end{eqnarray}
By the definition of $t_0$, we have $T - t_0 \geq \frac{T}{2}$. Plugging this estimate and notice that $\frac{(\epsilon')^2}{\delta} = \epsilon'$ by the definition of $\delta$, we have
\begin{eqnarray}
r_T &\leq& \frac{16b^2}{\mu^2\delta^2T^2}r_{t_0} + \frac{4LC + 8bD\gamma^H}{\mu^2\delta^2T} + \frac{2D\gamma^H}{\mu\delta} + \frac{4(\epsilon')^2(2b-LB)}{\mu^2\delta^2T} + \frac{2\epsilon'}{\mu} \nonumber \\ 
&\overset{T \geq \frac{b}{\mu\delta}}{\leq}& \frac{16b^2}{\mu^2\delta^2T^2}r_{t_0} + \frac{4LC}{\mu^2\delta^2T} + \frac{10D\gamma^H}{\mu\delta} + \frac{4(\epsilon')^2(2b-LB)}{\mu^2\delta^2T} + \frac{2\epsilon'}{\mu}  \nonumber \\ 
&\overset{\eqref{eq:rt0}}{\leq}& \frac{16b^2}{\mu^2\delta^2T^2}\left(\exp\left(-\frac{\mu\delta(T-1)}{2b}\right)r_0 + \frac{LC}{2\mu\delta b} + \frac{D\gamma^H}{\mu\delta} + \frac{(\epsilon')^2(2b-LB)}{2\mu\delta b}\right) \nonumber \\ 
&\quad& \ + \frac{4LC}{\mu^2\delta^2T} + \frac{10D\gamma^H}{\mu\delta} + \frac{4(\epsilon')^2(2b-LB)}{\mu^2\delta^2T} + \frac{2\epsilon'}{\mu}  \nonumber \\ 
&\overset{T \geq \frac{b}{\mu\delta}}{\leq}& 16\exp\left(-\frac{\mu\delta(T-1)}{2b}\right)r_0 + \frac{8LC}{\mu^2\delta^2T} + \frac{16D\gamma^H}{\mu\delta} + \frac{8(\epsilon')^2(2b-LB)}{\mu^2\delta^2 T} \nonumber \\ 
&\quad& \ + \frac{4LC}{\mu^2\delta^2T} + \frac{10D\gamma^H}{\mu\delta} + \frac{4(\epsilon')^2(2b-LB)}{\mu^2\delta^2T} + \frac{2\epsilon'}{\mu} \nonumber \\ 
&=& 16\exp\left(-\frac{\mu\delta(T-1)}{2b}\right)r_0 + \frac{12LC}{\mu^2\delta^2T} + \frac{26D\gamma^H}{\mu\delta} + \frac{12(\epsilon')^2(2b-LB)}{\mu^2\delta^2T} + \frac{2\epsilon'}{\mu}. \label{eq:>t0w}
\end{eqnarray}
It remains to take the maximum of the two bounds~\eqref{eq:T<w2} and~\eqref{eq:>t0w} with $b = \max\{\frac{2AL}{\mu\delta}, 2BL, \mu\delta\}$.
\end{proof}

\subsection{Proof of Corollary~\ref{cor:weak}}

\begin{proof}
From Theorem~\ref{pro:weak}, when $H = \cO(\log \epsilon^{-1})$, the dominant terms in~\eqref{eq:weak_bound} are $\frac{12LC}{\mu^2\delta^2T}$ and $\frac{2\epsilon'}{\mu}$. To guarantee that 
\[
\min\limits_{t \in \{0, 1, \cdots, T\}} J^* - \E{J(\theta_t)} \leq \cO(\epsilon) + \cO(\epsilon'),
\]
it suffices to choose $T = \cO(\delta^{-2}\epsilon^{-1})$ such that $\frac{12LC}{\mu^2\delta^2T} = \cO(\epsilon)$. Thus, by the definition of $\delta$, when $\epsilon' = 0$, we have $T = \cO(\epsilon^{-3})$; when $\epsilon' > 0$, we have $T = \cO((\epsilon')^{-2}\epsilon^{-1})$.
Otherwise, from Theorem~\ref{pro:weak}, notice that $\delta \leq \epsilon + \epsilon'$, we have $\min\limits_{t \in \{0, 1, \cdots, T-1\}} J^* - \E{J(\theta_t)} \leq \cO(\epsilon) + \cO(\epsilon')$, which concludes the proof.
\end{proof}

\section{Proof of Section~\ref{sec:expected}}

\subsection{Proof of Lemma~\ref{lem:ABC}}
\label{sec:proof_lem:ABC}

Note that a similar result to Lemma~\ref{lem:ABC} is given as Lemma 17 and 18 in~\citep{papini2019smoothing}. More precisely, Lemma 17 and 18 in~\citep{papini2019smoothing} provide an upper bound of the variance of the PG estimator similar to the following result
\[\mathbb{V}\mbox{ar}\left[\hnabla_m J(\theta)\right] \; \leq \; \frac{\nu}{m}.\]
We derive a slightly tighter bound
\[\mathbb{V}\mbox{ar}\left[\hnabla_m J(\theta)\right] \; \leq \; \frac{\nu - \norm{\nabla J_H(\theta)}}{m}.\]
This tighter bound is crucial for our work since it results in a tighter bound on
 $\E{\norm{\hnabla_m J(\theta)}^2}$ which still fits the format of \eqref{eq:ABC}. Here is the proof for Lemma~\ref{lem:ABC}.

\begin{proof}
Let $g(\tau\mid\theta)$ be a stochastic gradient estimator of one single sampled trajectory $\tau$. Thus $\hnabla_m J(\theta) = \frac{1}{m}\sum_{i=1}^m g(\tau_i\mid\theta)$. Both $\hnabla_m J(\theta)$ and $g(\tau\mid\theta)$ are unbiased estimators of $J_H(\theta)$. We have

\begin{eqnarray}
\E{\norm{\hnabla_m J(\theta)}^2} &=& \E{\norm{\frac{1}{m}\sum_{i=0}^{m-1} g(\tau_i\mid\theta)}^2} \nonumber \\
&=& \E{\norm{\frac{1}{m}\sum_{i=0}^{m-1} g(\tau_i\mid\theta) - \nabla J_H(\theta) + \nabla J_H(\theta)}^2} \nonumber \\
&=& \norm{\nabla J_H(\theta)}^2 + \E{\norm{\frac{1}{m}\sum_{i=0}^{m-1}\left(g(\tau_i\mid\theta)-\nabla J_H(\theta)\right)}^2} \nonumber \\
&=& \norm{\nabla J_H(\theta)}^2 + \frac{1}{m^2}\sum_{i=0}^{m-1}\E{\norm{g(\tau_i\mid\theta)-\nabla J_H(\theta)}^2} \nonumber \\
&=& \norm{\nabla J_H(\theta)}^2 + \frac{1}{m}\E{\norm{g(\tau_1\mid\theta)-\nabla J_H(\theta)}^2} \nonumber \\
&=& \norm{\nabla J_H(\theta)}^2 + \frac{\E{\norm{g(\tau_1\mid\theta)}^2 - \norm{\nabla J_H(\theta)}^2}}{m}, \label{eq:stochgradbnd}
\end{eqnarray}
where the third, the fourth and the fifth lines are all obtained by using $\nabla J_H(\theta) = \E{g(\tau_i\mid\theta)}$. It remains to show $\EE{\tau}{\norm{g(\tau\mid\theta)}^2}$ is bounded under Assumption~\ref{ass:lipschitz_smooth_policy}.

If $\hnabla_m J(\theta)$ is a REINFORCE gradient estimator, then
\begin{eqnarray}
\EE{\tau}{\norm{g(\tau\mid\theta)}^2} &\overset{\eqref{eq:REINFORCE}}{=}& \EE{\tau}{\norm{\sum_{t'=0}^{H-1}\gamma^{t'}\cR(s_{t'}, a_{t'})\cdot \sum_{t=0}^{H-1}\nabla_{\theta}\log\pi_{\theta}(a_t \mid s_t)}^2} \nonumber \\
&\leq& \frac{\cR_{\max}^2}{(1-\gamma)^2}\EE{\tau}{\norm{\sum_{t=0}^{H-1}\nabla_{\theta}\log\pi_{\theta}(a_t \mid s_t)}^2} \nonumber \\
&\overset{\eqref{eq:Etnorm2}}{=}& \frac{\cR_{\max}^2}{(1-\gamma)^2}\sum_{t=0}^{H-1}\EE{\tau}{\norm{\nabla_{\theta}\log\pi_{\theta}(a_t \mid s_t)}^2} \nonumber \\
&\overset{\eqref{eq:E_tau_G2}}{\leq}& \frac{HG^2\cR_{\max}^2}{(1-\gamma)^2}, \label{eq:tempREINFORCE}
\end{eqnarray}
where the second line is obtained by using $\left|\cR(s_{t'}, a_{t'})\right| \leq \cR_{\max}$.

Finally, the ABC assumption holds with
\begin{eqnarray*}
\E{\norm{\hnabla_m J(\theta)}^2} &\overset{\eqref{eq:stochgradbnd}+\eqref{eq:tempREINFORCE}}{\leq}& 
\left(1-\frac{1}{m}\right)\norm{\nabla J_H(\theta)}^2 + \frac{HG^2\cR_{\max}^2}{m(1-\gamma)^2}.
\end{eqnarray*}

If $\hnabla_m J(\theta)$ is a GPOMDP gradient estimator, then

\begin{eqnarray}
\EE{\tau}{\norm{g(\tau\mid\theta)}^2} &\overset{\eqref{eq:GPOMDP}}{=}& \EE{\tau}{\norm{\sum_{t=0}^{H-1}\gamma^{t/2}\cR(s_t, a_t) \gamma^{t/2}\left(\sum_{k=0}^{t}\nabla_{\theta}\log\pi_{\theta}(a_k \mid s_k)\right)}^2} \nonumber \\
&\leq& \EE{\tau}{\left(\sum_{t=0}^{H-1}\gamma^t\cR(s_t, a_t)^2\right)\left(\sum_{k=0}^{H-1}\gamma^k\norm{\sum_{k'=0}^{k}\nabla_{\theta}\log\pi_{\theta}(a_{k'} \mid s_{k'})}^2\right)} \nonumber \\
&\leq& \frac{\cR_{\max}^2}{1-\gamma}\cdot\sum_{k=0}^{H-1}\gamma^k\EE{\tau}{\norm{\sum_{k'=0}^{k}\nabla_{\theta}\log\pi_{\theta}(a_{k'} \mid s_{k'})}^2} \nonumber \\ 
&\overset{\eqref{eq:Etnorm2}}{=}& \frac{\cR_{\max}^2}{1-\gamma}\cdot\sum_{k=0}^{H-1}\gamma^k\sum_{k'=0}^{k}\EE{\tau}{\norm{\nabla_{\theta}\log\pi_{\theta}(a_{k'} \mid s_{k'})}^2} \nonumber \\ 
&\overset{\eqref{eq:E_tau_G2}}{\leq}& \frac{G^2\cR_{\max}^2}{1-\gamma}\cdot\sum_{k=0}^{H-1}\gamma^k(k+1) \nonumber \\ 
&\leq& \frac{G^2\cR_{\max}^2}{(1-\gamma)^3}, \label{eq:tempGPOMDP}
\end{eqnarray}
where the second line is from the Cauchy-Schwarz inequality, the third line is obtained by using $\left|\cR(s_t, a_t)\right| \leq \cR_{\max}$ and the last line is obtained by Lemma~\ref{lem:sum_of_gamma}. 

The above together with~\eqref{eq:stochgradbnd} imply that ABC assumption holds with
\begin{eqnarray*}
\E{\norm{\hnabla_m J(\theta)}^2} &\overset{\eqref{eq:stochgradbnd}+\eqref{eq:tempGPOMDP}}{\leq}& 
\left(1-\frac{1}{m}\right)\norm{\nabla J_H(\theta)}^2 + \frac{G^2\cR_{\max}^2}{m(1-\gamma)^3}.
\end{eqnarray*}
\end{proof}

\subsection{Proof of Corollary~\ref{cor:weakest}}

\begin{proof}
It is trivial that Assumption~\eqref{eq:lipschitz_smooth_policy} implies~(\nameref{E-LS}). Now we show that~(\nameref{ass:lipschitz_smooth_policy}) is strictly weaker than~\eqref{eq:lipschitz_smooth_policy}.

Consider a scalar-action, fixed-variance, Gaussian policy:
\begin{align} \label{eq:gauss}
\pi_\theta(a \mid s) \; = \; \cN\left(a \mid \theta^\top \phi(s), \sigma^2\right) \; = \; \frac{1}{\sigma\sqrt{2\pi}}\exp\left\{-\frac{1}{2}\left(\frac{a-\theta^\top\phi(s)}{\sigma}\right)^2\right\},
\end{align}
where $\theta \in \R^d$, $\sigma > 0$ is the standard deviation, and $\phi : \cS \rightarrow \cR^d$ is a mapping from the state
space to the feature space. 

From Lemma 23 in~\citet{papini2019smoothing}, the Gaussian policy~\eqref{eq:gauss} under the condition that the state feature vectors are bounded satisfies~(\nameref{ass:lipschitz_smooth_policy}). That is,  under the condition that there exists $\varphi \geq 0$ such that $\sup_{s \in \cS} \norm{\phi(s)} \leq \varphi$. One does not require that the actions are bounded for the Gaussian policy. This is not the case in~\citet{xu2020sample} in Section D under assumptions~\eqref{eq:lipschitz_smooth_policy}.

Besides, from Lemma~\ref{lem:ABC}, we know that Assumption~(\nameref{ass:lipschitz_smooth_policy}) implies~\eqref{eq:ABC}. This concludes the claim of the corollary.
\end{proof}

\subsection{Proof of Lemma~\ref{lem:smoothJ}}

\begin{proof}
We know that
\begin{eqnarray}
\nabla^2 J(\theta) &\overset{\eqref{eq:GD2*}}{=}& \nabla_\theta\EE{\tau}{\sum_{t=0}^\infty\gamma^t\cR(s_t, a_t)\left(\sum_{k=0}^t\nabla_{\theta}\log\pi_{\theta}(a_k \mid s_k)\right)} \nonumber \\
&=& \nabla_\theta\int p(\tau\mid\theta)\sum_{t=0}^\infty\gamma^t\cR(s_t, a_t)\left(\sum_{k=0}^t\nabla_{\theta}\log\pi_{\theta}(a_k \mid s_k)\right)d\tau \nonumber \\
&=& \int\nabla_\theta p(\tau\mid\theta)\left(\sum_{t=0}^\infty\gamma^t\cR(s_t, a_t)\left(\sum_{k=0}^t\nabla_{\theta}\log\pi_{\theta}(a_k \mid s_k)\right)\right)^\top d\tau \nonumber \\
&\quad& \ + \int p(\tau\mid\theta)\sum_{t=0}^\infty\gamma^t\cR(s_t, a_t)\left(\sum_{k=0}^t\nabla_{\theta}^2\log\pi_{\theta}(a_k \mid s_k)\right)d\tau \nonumber \\
&=& \int p(\tau\mid\theta)\nabla_\theta\log p(\tau\mid\theta)\left(\sum_{t=0}^\infty\gamma^t\cR(s_t, a_t)\left(\sum_{k=0}^t\nabla_{\theta}\log\pi_{\theta}(a_k \mid s_k)\right)\right)^\top d\tau \nonumber \\
&\quad& \ + \int p(\tau\mid\theta)\sum_{t=0}^\infty\gamma^t\cR(s_t, a_t)\left(\sum_{k=0}^t\nabla_{\theta}^2\log\pi_{\theta}(a_k \mid s_k)\right)d\tau \nonumber \\
&=& \EE{\tau}{\nabla_\theta\log p(\tau\mid\theta)\left(\sum_{t=0}^\infty\gamma^t\cR(s_t, a_t)\left(\sum_{k=0}^t\nabla_{\theta}\log\pi_{\theta}(a_k \mid s_k)\right)\right)^\top} \nonumber \\
&\quad& \ + \EE{\tau}{\sum_{t=0}^\infty\gamma^t\cR(s_t, a_t)\left(\sum_{k=0}^t\nabla_{\theta}^2\log\pi_{\theta}(a_k \mid s_k)\right)} \nonumber \\
&\overset{\eqref{eq:p}}{=}& \underbrace{\EE{\tau}{\sum_{t'=0}^\infty\nabla_\theta\log \pi_\theta(a_{t'}\mid\theta_{t'})\left(\sum_{t=0}^\infty\gamma^t\cR(s_t, a_t)\left(\sum_{k=0}^t\nabla_{\theta}\log\pi_{\theta}(a_k \mid s_k)\right)\right)^\top}}_{\circled{1}} \nonumber \\
&\quad& \ + \underbrace{\EE{\tau}{\sum_{t=0}^\infty\gamma^t\cR(s_t, a_t)\left(\sum_{k=0}^t\nabla_{\theta}^2\log\pi_{\theta}(a_k \mid s_k)\right)}}_{\circled{2}}. \label{eq:1+2}
\end{eqnarray}
We now bound the above two terms separately. The second term can be bounded easily. That is,
\begin{eqnarray}
\norm{\circled{2}} &\leq& \EE{\tau}{\sum_{t=0}^\infty\gamma^{t}\left|\cR(s_{t},a_{t})\right|\left(\sum_{k=0}^t\norm{\nabla_{\theta}^2\log\pi_{\theta}(a_k \mid s_k)}\right)} \nonumber \\
&\leq& \cR_{\max}\sum_{t=0}^\infty\gamma^t\left(\sum_{k=0}^t\EE{\tau}{\norm{\nabla_{\theta}^2\log\pi_{\theta}(a_k \mid s_k)}}\right) \nonumber \\ 
&\overset{\eqref{eq:E_tau_F}}{\leq}& F\cR_{\max}\sum_{t=0}^\infty\gamma^t(t+1) \nonumber \\
&=& \frac{F\cR_{\max}}{(1-\gamma)^2}, \label{eq:temp2}
\end{eqnarray}
where the second line is obtained by using $\left|\cR(s_t, a_t)\right| \leq \cR_{\max}$ and the last line is obtained by Lemma~\ref{lem:sum_of_gamma}.

To bound the first term, we use the following notation $x_{0:t} \eqdef (x_0, x_1, \cdots, x_t)$ with $\{x_t\}_{t\geq0}$ a sequence of random variables. Similar to the derivation of GPOMDP, we notice that future actions do not depend on past rewards and past actions. That is, for $0 \leq t < t'$ among terms of the two sums in \circled{1}, we have
\begin{align}
&\quad \EE{\tau}{\nabla_\theta\log\pi_\theta(a_{t'}\mid s_{t'})\cdot\gamma^t\cR(s_t,a_t)\left(\sum_{k=0}^t\nabla_\theta\log\pi_\theta(a_k\mid s_k)\right)^\top} \nonumber \\
&= \EE{s_{0:t'}, a_{0:t'}}{\nabla_\theta\log\pi_\theta(a_{t'}\mid s_{t'})\cdot\gamma^t\cR(s_t,a_t)\left(\sum_{k=0}^t\nabla_\theta\log\pi_\theta(a_k\mid s_k)\right)^\top} \nonumber \\
&= \EE{s_{0:t'}, a_{0:(t'-1)}}{\EE{a_{t'}}{\nabla_\theta\log\pi_\theta(a_{t'}\mid s_{t'})\cdot\gamma^t\cR(s_t,a_t)\left(\sum_{k=0}^t\nabla_\theta\log\pi_\theta(a_k\mid s_k)\right)^\top \ \bigg| \ s_{0:t'}, a_{0:(t'-1)}}} \nonumber \\
&= \EE{s_{0:t'}, a_{0:(t'-1)}}{\EE{a_{t'}}{\nabla_\theta\log\pi_\theta(a_{t'}\mid s_{t'}) \ \bigg| \ s_{t'}}\cdot\gamma^t\cR(s_t,a_t)\left(\sum_{k=0}^t\nabla_\theta\log\pi_\theta(a_k\mid s_k)\right)^\top} \nonumber \\
&= \EE{s_{0:t'}, a_{0:(t'-1)}}{\int\pi_\theta(a_{t'}\mid s_{t'})\nabla_\theta\log\pi_\theta(a_{t'}\mid s_{t'})da_{t'} \cdot\gamma^t\cR(s_t,a_t)\left(\sum_{k=0}^t\nabla_\theta\log\pi_\theta(a_k\mid s_k)\right)^\top} \nonumber \\
&= \EE{s_{0:t'}, a_{0:(t'-1)}}{\int\nabla_\theta\pi_\theta(a_{t'}\mid s_{t'})da_{t'} \cdot\gamma^t\cR(s_t,a_t)\left(\sum_{k=0}^t\nabla_\theta\log\pi_\theta(a_k\mid s_k)\right)^\top} \nonumber \\
&= \EE{s_{0:t'}, a_{0:(t'-1)}}{\nabla_\theta\underbrace{\int\pi_\theta(a_{t'}\mid s_{t'})da_{t'}}_{=1} \cdot\gamma^t\cR(s_t,a_t)\left(\sum_{k=0}^t\nabla_\theta\log\pi_\theta(a_k\mid s_k)\right)^\top} \nonumber \\
&= 0, \label{eq:0}
\end{align}
where the third equality is obtained by the Markov property.
Thus, \circled{1} can be simplified. We have
\begin{eqnarray}
\circled{1} &\overset{\eqref{eq:0}}{=}& \EE{\tau}{\sum_{t'=0}^t\nabla_\theta\log \pi_\theta(a_{t'}\mid\theta_{t'})\left(\sum_{t=0}^\infty\gamma^t\cR(s_t, a_t)\left(\sum_{k=0}^t\nabla_{\theta}\log\pi_{\theta}(a_k \mid s_k)\right)\right)^\top} \nonumber \\
&=& \EE{\tau}{\sum_{t=0}^\infty\gamma^t\cR(s_t, a_t)\left(\sum_{t'=0}^t\nabla_\theta\log \pi_\theta(a_{t'}\mid\theta_{t'})\right)\left(\sum_{k=0}^t\nabla_{\theta}\log\pi_{\theta}(a_k \mid s_k)\right)^\top}. \label{eq:temp1}
\end{eqnarray}
Now we can bound \circled{1} easily. That is,
\begin{eqnarray}
\norm{\circled{1}} &\overset{\eqref{eq:temp1}}{\leq}& \EE{\tau}{\sum_{t=0}^\infty\gamma^t\left|\cR(s_t, a_t)\right|\norm{\sum_{t'=0}^t\nabla_\theta\log \pi_\theta(a_{t'}\mid\theta_{t'})}^2} \nonumber \\
%&\leq& \EE{\tau}{\sum_{t=0}^\infty\gamma^t\left|\cR(s_t, a_t)\right|\left(\sum_{t'=0}^t\norm{\nabla_\theta\log \pi_\theta(a_{t'}\mid\theta_{t'})}\right)^2} \nonumber \\
%&\leq& \EE{\tau}{G^2\cR_{\max}\sum_{t=0}^\infty(t+1)^2\gamma^t} \nonumber \\
%&\leq& \frac{2G^2\cR_{\max}}{(1-\gamma)^3}
&\leq& \cR_{\max}\sum_{t=0}^\infty\gamma^t\EE{\tau}{\norm{\sum_{t'=0}^t\nabla_\theta\log \pi_\theta(a_{t'}\mid\theta_{t'})}^2} \nonumber \\ 
&\overset{\eqref{eq:Etnorm2}}{=}& \cR_{\max}\sum_{t=0}^\infty\gamma^t\sum_{t'=0}^t\EE{\tau}{\norm{\nabla_\theta\log \pi_\theta(a_{t'}\mid\theta_{t'})}^2} \nonumber \\
&\overset{\eqref{eq:E_tau_G2}}{\leq}& G^2\cR_{\max}\sum_{t=0}^\infty\gamma^t(t+1) \nonumber \\ 
&=& \frac{G^2\cR_{\max}}{(1-\gamma)^2}, \label{eq:temp1bnd}
\end{eqnarray}
where the second line is obtained by using $\left|\cR(s_t, a_t)\right| \leq \cR_{\max}$ and the last line is obtained by Lemma~\ref{lem:sum_of_gamma}.

Finally,
\begin{eqnarray*}
\norm{\nabla^2J(\theta)} &\overset{\eqref{eq:1+2}+\eqref{eq:temp1bnd}+\eqref{eq:temp2}}{\leq}& \frac{\cR_{\max}}{(1-\gamma)^2}(G^2+F).
\end{eqnarray*}
\end{proof}

\subsection{Proof of Lemma~\ref{lem:trunc}}

\begin{proof}
From~\eqref{eq:GD2*}, we have
\begin{eqnarray} 
\norm{\nabla J(\theta) - \nabla J_H(\theta)}^2 &=& \norm{\EE{\tau}{\sum_{t=H}^\infty\gamma^t\cR(s_t, a_t)\left(\sum_{k=0}^t\nabla_{\theta}\log\pi_{\theta}(a_k \mid s_k)\right)}}^2 \nonumber \\
&\leq& \EE{\tau}{\norm{\sum_{t=H}^\infty\gamma^{t/2}\cR(s_t, a_t)\gamma^{t/2}\left(\sum_{k=0}^t\nabla_{\theta}\log\pi_{\theta}(a_k \mid s_k)\right)}^2} \nonumber \\
&\leq& \EE{\tau}{\left(\sum_{t=H}^\infty\gamma^t\cR(s_t, a_t)^2\right)\left(\sum_{k=H}^\infty\gamma^k\norm{\sum_{k'=0}^k\nabla_{\theta}\log\pi_{\theta}(a_{k'} \mid s_{k'})}^2\right)} \nonumber \\
&\leq& \frac{\cR_{\max}^2\gamma^H}{1-\gamma}\EE{\tau}{\sum_{k=H}^\infty\gamma^k\norm{\sum_{k'=0}^k\nabla_{\theta}\log\pi_{\theta}(a_{k'} \mid s_{k'})}^2} \nonumber \\
&\overset{\eqref{eq:Etnorm2}}{=}& \frac{\cR_{\max}^2\gamma^H}{1-\gamma}\sum_{k=H}^\infty\gamma^k\sum_{k'=0}^k\EE{\tau}{\norm{\nabla_{\theta}\log\pi_{\theta}(a_{k'} \mid s_{k'})}^2} \nonumber \\
&\overset{\eqref{eq:E_tau_G2}}{\leq}& \frac{G^2\cR_{\max}^2\gamma^H}{1-\gamma}\sum_{k=H}^\infty\gamma^k(k+1) \nonumber \\
&=& \frac{G^2\cR_{\max}^2\gamma^{2H}}{1-\gamma}\sum_{k=0}^\infty \gamma^k(k+1+H) \nonumber \\ 
&=& \left(\frac{1}{1-\gamma} + H\right)\frac{G^2\cR_{\max}^2\gamma^{2H}}{(1-\gamma)^2}, \label{eq:D'}
\end{eqnarray}
where the second and third lines are obtained by Jensen and Cauchy-Schwarz inequality respectively, the fourth line is obtained by using $\left|\cR(s_t, a_t)\right| \leq \cR_{\max}$ and the last line is obtained by Lemma~\ref{lem:sum_of_gamma}.

Thus
\begin{eqnarray*}
D' &\overset{\eqref{eq:D'}}{=}& \frac{G\cR_{\max}}{1-\gamma}\sqrt{\frac{1}{1-\gamma} + H}.
\end{eqnarray*}

Next, by inequality of Cauchy-Swartz we have
\begin{eqnarray}
\left|\dotprod{\nabla J_H(\theta), \nabla J_H(\theta) - \nabla J(\theta)}\right| &\leq& \norm{\nabla J_H(\theta)}\norm{\nabla J_H(\theta) - \nabla J(\theta)} \nonumber \\
&\overset{\eqref{eq:trunc2}}{\leq}& \norm{\nabla J_H(\theta)}\cdot D'\gamma^H \nonumber \\ 
&\leq& \frac{D'G\cR_{\max}}{(1-\gamma)^{3/2}}\gamma^H, \label{eq:D}
\end{eqnarray}
where the last line is obtained by Lemma~\ref{lem:lipschitzJ}~\ref{itm:J-lipschitz}. Thus
\begin{eqnarray*}
D &\overset{\eqref{eq:D}}{=}& \frac{D'G\cR_{\max}}{(1-\gamma)^{3/2}}.
\end{eqnarray*}
\end{proof}

\subsection{Lipschitz continuity of $J(\cdot)$}
\label{sec:lipschitz}

In this section, we show that $J(\cdot)$ is Lipschitz-continuous under Assumption~\ref{ass:lipschitz_smooth_policy}.

\begin{lemma} \label{lem:lipschitzJ}
If Assumption~\ref{ass:lipschitz_smooth_policy} holds, for any $m$ trajectories $\tau_i$ and $\theta\in\R^d$, we have
\begin{enumerate}[label=(\roman*)]
    \item\label{itm:g-smooth} $\hnabla_m J(\theta)$ is $L_g$-Lipschitz continuous if conditions~\eqref{eq:lipschitz_smooth_policy} hold;
    \item\label{itm:g-bounded} The norm of the gradient estimator squared in expectation is bounded, i.e. $\E{\norm{\hnabla_m J(\theta)}^2} \leq \Gamma_g^2$.
    \item\label{itm:J-lipschitz} $J(\cdot)$ is $\Gamma$-Lipschitz, namely $\norm{\nabla J(\theta)} \leq \Gamma$  with $\Gamma = \frac{G\cR_{\max}}{(1-\gamma)^{3/2}}$. Similarly, we have $\norm{\nabla J_H(\theta)} \leq \Gamma$ for the exact policy gradient of the truncated function $J_H(\cdot)$ for any horizon $H$.
\end{enumerate}
Furthermore, if $\hnabla_m J(\theta)$ is a REINFORCE gradient estimator, then $L_g = \frac{HF\cR_{\max}}{1-\gamma}$ and $\Gamma_g = \frac{\sqrt{H}G\cR_{\max}}{1-\gamma}$; if $\hnabla_m J(\theta)$ is a GPOMDP gradient estimator, then $L_g = \frac{F\cR_{\max}}{(1-\gamma)^2}$ and $\Gamma_g = \Gamma$.
\end{lemma}

\paragraph*{Remark.}
The Lipschitzness constant proposed in Lemma~\ref{lem:lipschitzJ}~\ref{itm:J-lipschitz} is novel. See Section~\ref{sec:better_cst} for more details.
%It is tighter as compared to equation (35) in the proof of Lemma 6 in~\citep{papini2019smoothing} under Assumption~(\nameref{ass:lipschitz_smooth_policy}). Compared to their bound, our result shows that when $\gamma$ is close to 1, the Lipschitzness constant $\Gamma$ depends on $(1-\gamma)^{-3/2}$ instead of $(1-\gamma)^{-2}$ derived in~\citep{papini2019smoothing}. 

The results in Lemma~\ref{lem:lipschitzJ}~\ref{itm:g-bounded}
% was already provided in Lemma 17 and 18 in~\citep{papini2019smoothing} under the same Assumption~(\nameref{ass:lipschitz_smooth_policy}). It is also 
match the special case of Lemma~\ref{lem:ABC} when the mini-batch size $m = 1$. It also implies Assumption~\eqref{eq:ABC} but with a looser upper bound, which is independent to the batch size $m$. We include a proof for completeness of the properties of a general vanilla policy gradient estimator. Notice that the bound of $\E{\norm{\hnabla_m J(\theta)}^2}$ with GPOMDP gradient estimator is a factor of $1 - \gamma$ tighter as compared to Proposition~4.2~(3) in~\citep{xu2020sample} and equation (17) in~\citep{yuan2020stochastic} under more restrictive assumptions~\eqref{eq:lipschitz_smooth_policy}.

The result with GPOMDP gradient estimator in Lemma~\ref{lem:lipschitzJ}~\ref{itm:g-smooth}
%and the Lipschitz continuity of $J(\cdot)$
was already proposed in Proposition~4.2 in~\citep{xu2020sample}, but not with REINFORCE gradient estimator. We include a proof for both gradient estimators for the completeness.

%\rob{These  comments are very similar/repeated to the comments in Section A.2. How about move up to section A and merge with comments there? }
%\rui{The first paragraphe is already explained in Section A.2. So I shortened it. The others are different compared to Section A.2.}

\begin{proof}
To prove~\ref{itm:g-smooth}, let $\hnabla_m J(\theta)$ be a REINFORCE gradient estimator. From~\eqref{eq:REINFORCE}, we have
\begin{eqnarray}
\norm{\nabla\left(\hnabla_m J(\theta)\right)} &=& \norm{\frac{1}{m}\sum_{i=1}^m\sum_{t=0}^{H-1}\left(\sum_{t'=0}^{H-1}\gamma^{t'}\cR(s_{t'}^i, a_{t'}^i)\right)\nabla_{\theta}^2\log\pi_{\theta}(a_t^i \mid s_t^i)} \nonumber \\
&\leq& \frac{1}{m}\sum_{i=1}^m\left(\sum_{t'=0}^{H-1}\gamma^{t'}\left|\cR(s_{t'}^i, a_{t'}^i)\right|\right)\sum_{t=0}^{H-1}\norm{\nabla_{\theta}^2\log\pi_{\theta}(a_t^i \mid s_t^i)} \nonumber \\
&\leq& \frac{\cR_{\max}}{1-\gamma}\cdot\frac{1}{m}\sum_{i=1}^m\sum_{t=0}^{H-1}\norm{\nabla_{\theta}^2\log\pi_{\theta}(a_t^i \mid s_t^i)} \nonumber \\
&\overset{\eqref{eq:lipschitz_smooth_policy}}{\leq}& \frac{HF\cR_{\max}}{1-\gamma},
\end{eqnarray}
where the third line is obtained by using $\left|\cR(s_{t'}^i, a_{t'}^i)\right| \leq \cR_{\max}$. In this case, $L_g = \frac{HF\cR_{\max}}{1-\gamma}$.

Let $\hnabla_m J(\theta)$ be a GPOMDP gradient estimator. From~\eqref{eq:GPOMDP}, we have
\begin{eqnarray}
\norm{\nabla \left(\hnabla_m J(\theta)\right)} &=& \norm{\frac{1}{m}\sum_{i=1}^m\sum_{t=0}^{H-1}\gamma^t\cR(s_t^i, a_t^i)\left(\sum_{k=0}^t\nabla_{\theta}^2\log\pi_{\theta}(a_k^i \mid s_k^i)\right)} \nonumber \\
&\leq& \frac{1}{m}\sum_{i=1}^m\sum_{t=0}^{H-1}\gamma^t\left|\cR(s_t^i, a_t^i)\right|\left(\sum_{k=0}^t\norm{\nabla_{\theta}^2\log\pi_{\theta}(a_k^i \mid s_k^i)}\right) \nonumber \\
&\leq& \frac{\cR_{\max}}{m}\sum_{i=1}^m\sum_{t=0}^{H-1}\gamma^t\left(\sum_{k=0}^t\norm{\nabla_{\theta}^2\log\pi_{\theta}(a_k^i \mid s_k^i)}\right) \nonumber \\
&\overset{\eqref{eq:lipschitz_smooth_policy}}{\leq}& F\cR_{\max}\sum_{t=0}^{H-1}\gamma^t(t+1) \nonumber \\
&\overset{\mbox{Lemma~\ref{lem:sum_of_gamma}}}{\leq}& \frac{F\cR_{\max}}{(1-\gamma)^2},
\end{eqnarray}
where similarly, the third line is obtained by using $\left|\cR(s_t^i, a_t^i)\right| \leq \cR_{\max}$. In this case, $L_g = \frac{F\cR_{\max}}{(1-\gamma)^2}$.

To prove~\ref{itm:g-bounded}, let $g(\tau\mid\theta)$ be a stochastic gradient estimator of one single sampled trajectory $\tau$. Thus $\hnabla_m J(\theta) = \frac{1}{m}\sum_{i=1}^m g(\tau_i\mid\theta)$. Both $\hnabla_m J(\theta)$ and $g(\tau\mid\theta)$ are unbiased estimators of $J_H(\theta)$. We have
\begin{eqnarray*}
\E{\norm{\hnabla_m J(\theta)}^2} &\leq& \EE{\tau}{\norm{g(\tau\mid\theta)}^2}.
\end{eqnarray*}

If $\hnabla_m J(\theta)$ is a REINFORCE gradient estimator, from~\eqref{eq:tempREINFORCE}, we have $\Gamma_g = \frac{\sqrt{H}G\cR_{\max}}{1-\gamma}$. If $\hnabla_m J(\theta)$ is a GPOMDP gradient estimator, from~\eqref{eq:tempGPOMDP}, we have $\Gamma_g = \frac{G\cR_{\max}}{(1-\gamma)^{3/2}}$.

%We simply replace $\nabla \left(\hnabla_m J(\theta)\right)$ by $\hnabla_m J(\theta)$, $\nabla_{\theta}^2\log\pi_{\theta}(a_t^i \mid s_t^i)$ by $\nabla_{\theta}\log\pi_{\theta}(a_t^i \mid s_t^i)$ and $F$ by $G$. If $\hnabla_m J(\theta)$ is a REINFORCE gradient estimator, we have $\Gamma_g = \frac{HG\cR_{\max}}{1-\gamma}$; if $g(\tau\mid\cdot)$ is a GPOMDP gradient estimator, then $\Gamma_g = \frac{G\cR_{\max}}{(1-\gamma)^2}$.

To prove~\ref{itm:J-lipschitz}, we have
\begin{eqnarray} 
\norm{\nabla J(\theta)}^2 &\overset{\eqref{eq:GD2*}}{=}& \norm{\EE{\tau}{\sum_{t=0}^\infty\gamma^t\cR(s_t, a_t)\left(\sum_{k=0}^t\nabla_{\theta}\log\pi_{\theta}(a_k \mid s_k)\right)}}^2 \nonumber \\
&\leq& \EE{\tau}{\norm{\sum_{t=0}^\infty\gamma^{t/2}\cR(s_t, a_t)\gamma^{t/2}\left(\sum_{k=0}^t\nabla_{\theta}\log\pi_{\theta}(a_k \mid s_k)\right)}^2} \nonumber \\
&\leq& \EE{\tau}{\left(\sum_{t=0}^\infty\gamma^t\cR(s_t, a_t)^2\right)\left(\sum_{k=0}^\infty\gamma^k\norm{\sum_{k'=0}^k\nabla_{\theta}\log\pi_{\theta}(a_{k'} \mid s_{k'})}^2\right)} \nonumber \\
&\leq& \frac{\cR_{\max}^2}{1-\gamma}\EE{\tau}{\sum_{k=0}^\infty\gamma^k\norm{\sum_{k'=0}^k\nabla_{\theta}\log\pi_{\theta}(a_{k'} \mid s_{k'})}^2} \nonumber \\
&\overset{\eqref{eq:Etnorm2}}{=}& \frac{\cR_{\max}^2}{1-\gamma}\sum_{k=0}^\infty\gamma^k\sum_{k'=0}^k\EE{\tau}{\norm{\nabla_{\theta}\log\pi_{\theta}(a_{k'} \mid s_{k'})}^2} \nonumber \\
&\overset{\eqref{eq:E_tau_G2}}{\leq}& \frac{G^2\cR_{\max}^2}{1-\gamma}\sum_{k=0}^\infty\gamma^k(k+1) \nonumber \\
&=& \frac{G^2\cR_{\max}^2}{(1-\gamma)^3}, 
\end{eqnarray}
where the second and third lines are obtained by Jensen and Cauchy-Schwarz inequality respectively, the fourth line is obtained by using $\left|\cR(s_t, a_t)\right| \leq \cR_{\max}$ and the last line is obtained by Lemma~\ref{lem:sum_of_gamma}.

Thus,
\begin{eqnarray*}
\norm{\nabla J(\theta)} &\leq& \Gamma \quad \mbox{ with } \quad \Gamma \; = \; \frac{G\cR_{\max}}{(1-\gamma)^{3/2}}.
\end{eqnarray*}
Similarly, we also have
\begin{eqnarray*}
\norm{\nabla J_H(\theta)} &\leq& \Gamma \quad \mbox{ with } \quad \Gamma \; = \; \frac{G\cR_{\max}}{(1-\gamma)^{3/2}}
\end{eqnarray*}
for the exact policy gradient of the truncated function $J(\cdot)$ for any horizon $H$.
\end{proof}

\subsection{Proof of Corollary~\ref{cor:expected}}

\begin{proof}
From Lemma~\ref{lem:smoothJ}, we know that $J$ is $L$-smooth. 
Consider policy gradient with a mini-batch sampling of size $m$. From Lemma~\ref{lem:ABC}, we have Assumption~\ref{ass:ABC} holds with $A=0$, $B=1-\frac{1}{m}$ and $C=\nu / m$. Assumption~\ref{ass:trunc} is verified as well by Lemma~\ref{lem:trunc} with appropriate $D$ and $D'$. By Theorem~\ref{pro:ABC}, plugging $A=0$, $B=1-\frac{1}{m}$ and $C=\nu /m$ in~\eqref{eq:A=0} yields the corollary's claim with step size $\eta\in\left(0,\frac{2}{L\left(1-\frac{1}{m}\right)}\right)$.
%As for the value of $\Gamma_g^2$ in REINFOCE or G(PO)MDP without baselines, this follows from the claim in Lemma~\ref{lem:lipschitzJ}.
\end{proof}

\subsection{Proof of Corollary~\ref{cor:sample_complexity}}

\begin{proof}

Consider vanilla policy gradient with step size $\eta\in\left(0,\frac{1}{L\left(1-\frac{1}{m}\right)}\right)$ and a mini-batch sampling of size $m$. We have
\begin{eqnarray*}
\E{\norm{\nabla J(\theta_U)}^2} & \overset{\eqref{eq:A=0B=1C=g/m}}{\leq} & \frac{2\delta_0}{\eta T\left(2-L\eta\left(1-\frac{1}{m}\right)\right)} + \frac{L\nu\eta}{m\left(2-L\eta\left(1-\frac{1}{m}\right)\right)} \nonumber \\ 
&\quad& \ + \left(\frac{2D\left(3-L\eta\left(1-\frac{1}{m}\right)\right)}{2-L\eta\left(1-\frac{1}{m}\right)} + D'^2\gamma^{H}\right)\gamma^H \nonumber \\ 
& \leq & \frac{2\delta_0}{\eta T} + \frac{L\nu\eta}{m} + \left(6D + D'^2\gamma^{H}\right)\gamma^H,
\end{eqnarray*}
where the second inequality is obtained by $\frac{1}{2-L\eta\left(1-\frac{1}{m}\right)} \leq 1$ with $\eta\in\left(0,\frac{1}{L\left(1-\frac{1}{m}\right)}\right)$. 

To get $\E{\norm{\nabla J(\theta_U)}^2} = \cO(\epsilon^2)$, it suffices to have
\begin{eqnarray} \label{eq:mT}
\cO(\epsilon^2) & \geq & \frac{2\delta_0}{\eta T} + \frac{L\nu\eta}{m}
\end{eqnarray}
and
\begin{eqnarray} \label{eq:gammaH}
\cO(\epsilon^2) &\geq& \left(6D + D'^2\gamma^{H}\right)\gamma^H
\end{eqnarray}
respectively.
To make the right hand side of~\eqref{eq:gammaH} smaller than $\epsilon^2$, we need $H\gamma^H = \cO(\epsilon^2)$. Thus, we require
%\rob{Maybe you are plugging in a bound on $D$, but shouldnt it be ``we need $D\gamma^H = \cO(\epsilon^2)$.''}
%\rui{Yes, I already plugged in a bound on $D$ and $D'^2$. The dominant term for $D$ and $D'^2$ is $H$.}
\[H = \cO\left(\log\left(\frac{1}{\epsilon}\right)/\log\left(\frac{1}{\gamma}\right)\right).\]
To make the right hand side of~\eqref{eq:mT} smaller than $\epsilon^2$, we require
\begin{eqnarray} \label{eq:eta<}
\frac{L\nu\eta}{m} \; \leq \; \frac{\epsilon^2}{2} &\Longleftrightarrow& \eta \; \leq \; \frac{\epsilon^2m}{2L\nu}.
\end{eqnarray}
Similarly, for the first term of the right hand side of~\eqref{eq:mT}, we require
\begin{eqnarray} \label{eq:eta>}
\frac{2\delta_0}{\eta T} \; \leq \; \frac{\epsilon^2}{2} &\Longleftrightarrow& \frac{4\delta_0}{\epsilon^2T} \; \leq \; \eta.
\end{eqnarray}
Combining the above two inequalities gives
\begin{eqnarray} \label{eq:eta}
\frac{4\delta_0}{\epsilon^2T} \; \leq \; \eta \; \leq \; \frac{\epsilon^2m}{2L\nu}.
\end{eqnarray}
This implies
\begin{eqnarray} \label{eq:Tm8}
Tm &\geq& \frac{8\delta_0L\nu}{\epsilon^4}.
\end{eqnarray}
The condition on the step size $\eta \in \left(0, \frac{1}{L\left(1-\frac{1}{m}\right)}\right)$ requires that the mini-batch size satisfies
\[\frac{\epsilon^2m}{2L\nu} \; \leq \; \frac{1}{L\left(1-\frac{1}{m}\right)} \; \Longrightarrow \; m \; \leq \; \frac{2\nu}{\epsilon^2}.\]
To conclude, it suffices to choose the step size $\eta = \frac{4\delta_0}{\epsilon^2T} = \frac{\epsilon^2m}{2L\nu}$, a mini-batch size $m$ between $1$ and $\frac{2\nu}{\epsilon^2}$, the number of iterations $T = \frac{8\delta_0L\nu}{m\epsilon^4}$ and the fixed Horizon $H = \cO\left(\log\left(\frac{1}{\epsilon}\right)/\log\left(\frac{1}{\gamma}\right)\right)$ so that the inequalities~\eqref{eq:gammaH},~\eqref{eq:eta<},~\eqref{eq:eta>},~\eqref{eq:eta} and~\eqref{eq:Tm8} hold, which guarantee $\E{\norm{\nabla J(\theta_U)}^2} = \cO(\epsilon^2)$.

Thus, the total sample complexity is 
\begin{eqnarray*}
Tm \times H \; = \; \frac{8\delta_0L\nu\log\left(\frac{1}{\epsilon}\right)}{\log\left(\frac{1}{\gamma}\right)\epsilon^4} \; = \; \widetilde{\cO}(\epsilon^{-4}).
\end{eqnarray*}

More precisely, from Lemma~\ref{lem:smoothJ}, $L = \frac{\cR_{\max}}{(1-\gamma)^2}(G^2+F)$. When using REINFORCE gradient estimator~\eqref{eq:REINFORCE}, from Lemma~\ref{lem:ABC}, $\nu = \frac{HG^2\cR_{\max}^2}{(1-\gamma)^2}$. Thus, when $\gamma$ is close to $1$, the sample complexity is
\begin{align} \label{eq:sample_complexity_REINFORCE}
\frac{8\delta_0H^2G^2\cR_{\max}^3(G^2+F)}{(1-\gamma)^4\epsilon^4} \; = \; \frac{8\delta_0G^2\cR_{\max}^3(G^2+F)\left(\log\left(\frac{1}{\epsilon}\right)\right)^2}{\left(\log\left(\frac{1}{\gamma}\right)\right)^2(1-\gamma)^4\epsilon^4} \; = \; \cO\left(\left(\log\left(\frac{1}{\epsilon}\right)\right)^2(1-\gamma)^{-6}\epsilon^{-4}\right).
\end{align}
In this case, we can choose the mini-batch size $m \in \left[1; \frac{2\nu}{\epsilon^2}\right]$, i.e. from $1$ to $\cO\left(H(1-\gamma)^{-2}\epsilon^{-2}\right)$ and the constant step size $\eta = \frac{\epsilon^2m}{2L\nu}$ varies from $\cO\left((1-\gamma)^2\right)$ to $\cO\left(H^{-1}(1-\gamma)^4\epsilon^2\right)$ accordingly.

When using GPOMDP gradient estimator~\eqref{eq:GPOMDP}, from Lemma~\ref{lem:ABC}, $\nu = \frac{G^2\cR_{\max}^2}{(1-\gamma)^3}$. Thus, when $\gamma$ is close to $1$, the sample complexity is
\begin{align} \label{eq:sample_complexity_GPOMDP}
\frac{8\delta_0HG^2\cR_{\max}^3(G^2+F)}{(1-\gamma)^5\epsilon^4} \; = \; \frac{8\delta_0G^2\cR_{\max}^3(G^2+F)\log\left(\frac{1}{\epsilon}\right)}{\log\left(\frac{1}{\gamma}\right)(1-\gamma)^5\epsilon^4} \; = \; \cO\left(\log\left(\frac{1}{\epsilon}\right)(1-\gamma)^{-6}\epsilon^{-4}\right).
\end{align}
In this case, we can choose the mini-batch size $m \in \left[1; \frac{2\nu}{\epsilon^2}\right]$, i.e. from $1$ to $\cO\left((1-\gamma)^{-3}\epsilon^{-2}\right)$ and the constant step size $\eta = \frac{\epsilon^2m}{2L\nu}$ varies from $\cO\left((1-\gamma)^2\right)$ to $\cO\left((1-\gamma)^5\epsilon^2\right)$ accordingly.
\end{proof}

\paragraph*{Remark.} Comparing~\eqref{eq:sample_complexity_GPOMDP} to~\eqref{eq:sample_complexity_REINFORCE}, we have that the sample complexity of GPOMDP is a factor of $\log\left(1/\epsilon\right)$ smaller than that of REINFORCE.

% !TEX root = summary.tex

\section{Proof of Section~\ref{sec:softmax}}

In this section, $\theta \in \R^{|\cS| |\cA|}$ and denote $\theta_s \equiv (\theta_{s,a})_{a \in \cA} \in \R^{|\cA|}$. We also use the following notations
\[\pi_{s,a}(\theta) \; \eqdef \; \pi_\theta(a \mid s) \quad \quad \mbox{ and } \quad \quad \pi_s(\theta) \; \eqdef \; \pi_\theta(\cdot \mid s) \; \in \; \Delta(\cA) \; \in \; \R^{|\cA|}.\]

\subsection{Preliminaries for the softmax tabular policy}

Recall the softmax tabular policy given by
\begin{eqnarray} \label{eq:softmax}
\pi_{s,a}(\theta) \; \eqdef \; \frac{\exp(\theta_{s,a})}{\sum_{a'\in\cA}\exp(\theta_{s,a'})}.
\end{eqnarray}

From~\eqref{eq:softmax}, for any $(s, a, a') \in \cS \times \cA \times \cA$ with $a' \neq a$, we have immediately the following partial derivatives for the softmax tabular policy
\begin{align}
\frac{\partial \pi_{s,a}(\theta)}{\partial \theta_{s,a}} &= \pi_{s,a}(\theta)(1-\pi_{s,a}(\theta)), \label{eq:softmax_sa} \\ 
\frac{\partial \pi_{s,a}(\theta)}{\partial \theta_{s,a'}} &= -\pi_{s,a}(\theta)\pi_{s,a'}(\theta). \label{eq:softmax_sa'}
\end{align}
Notice that for $s' \in \cS$ with $s' \neq s$, we have $\frac{\partial \pi_{s,a}(\theta)}{\partial \theta_{s',a}} = 0$.
From~\eqref{eq:softmax_sa} and~\eqref{eq:softmax_sa'}, we obtain respectively the gradient of $\pi_{s,a}(\theta)$ and the Jacobian of $\pi_s(\theta)$ w.r.t. $\theta_s$
\begin{align}
\frac{\partial \pi_{s,a}(\theta)}{\partial \theta_{s}} &= \left(\frac{\partial \pi_{s}(\theta)}{\partial \theta_{s,a}}\right)^\top = \pi_{s,a}(\theta)({\bf 1}_a - \pi_{s}(\theta)), \label{eq:softmax_gradient} \\ 
\frac{\partial \pi_{s}(\theta)}{\partial \theta_{s}} &= \Diag{\pi_s(\theta)} - \pi_s(\theta)\pi_s(\theta)^\top \eqdef \mH(\pi_s(\theta)),
\label{eq:softmax_Jacobian}
\end{align}
where ${\bf 1}_a \in \R^{|\cA|}$ is a vector with zero entries except one non-zero entry $1$ corresponding to the action $a$. Now from~\eqref{eq:softmax_gradient} and~\eqref{eq:softmax_Jacobian}, we obtain respectively the gradient and the Hessian of $\log\pi_{s,a}(\theta)$ w.r.t. $\theta_s$ given by
\begin{align}
\frac{\partial \log\pi_{s,a}(\theta)}{\partial \theta_{s}} &= {\bf 1}_a - \pi_s(\theta), \label{eq:softmax_log_gradient} \\ 
\frac{\partial^2 \log\pi_{s,a}(\theta)}{\partial \theta_{s}^2} &= -\mH(\pi_s(\theta)). \label{eq:softmax_log_hessian}
\end{align}

\subsection{Stationary point convergence of the softmax tabular policy}
\label{sec:fosp:softmax}

First we provide the proof of Lemma~\ref{lem:softmax_expected}.

\begin{proof}
For any state $s \in \cS$ and any $\theta \in \R^{|\cS| |\cA|}$, from~\eqref{eq:softmax_log_gradient}, we have
\begin{eqnarray} \label{eq:G2_LS}
\EE{a \sim \pi_\theta(\cdot \mid s)}{\norm{\nabla_\theta \log\pi_\theta(a \mid s)}^2} &=& \EE{a \sim \pi_\theta(\cdot \mid s)}{1 + \norm{\pi_s(\theta)}^2 - 2 \pi_{s,a}(\theta)} \nonumber \\ 
&=& 1 + \norm{\pi_s(\theta)}^2 - 2 \sum_{a \in \cA} \pi_{s,a}(\theta)^2 \nonumber \\ 
&=& 1 - \norm{\pi_s(\theta)}^2 \nonumber \\ 
&\leq& 1 - \frac{1}{|\cA|}, 
\end{eqnarray}
where the last line is obtained by using Cauchy-Schwarz inequality in the following
\[\norm{\pi_s(\theta)}^2 \; = \; \sum_{a \in \cA} \pi_{s,a}(\theta)^2 \; \geq \; \frac{1}{|\cA|}\left(\sum_{a \in \cA} \pi_{s,a}(\theta)\right)^2 \; = \; \frac{1}{|\cA|}.\]
Thus we have $G^2 = 1 - \frac{1}{|\cA|}.$

Besides, from Lemma 22 in~\citet{mei2020ontheglobal}, we have $\norm{\mH(\pi_s(\theta))} \leq 1$. Thus from~\eqref{eq:softmax_log_hessian}, we have $\norm{\nabla_\theta^2 \log\pi_\theta(a \mid s)} \leq 1$. Taking expectation over action, we have
\[\EE{a \sim \pi_\theta(\cdot \mid s)}{\norm{\nabla_\theta^2 \log\pi_\theta(a \mid s)}} \; \leq \; 1.\]
Thus we have $F = 1.$
\end{proof}

\paragraph*{Remark.} Without expectation, for any $(s,a) \in \cS \times \cA$,~\eqref{eq:G2_LS} becomes
\begin{eqnarray} \label{eq:softmax_LS_nabla_log}
\norm{\nabla_\theta \log\pi_\theta(a \mid s)}^2 \; = \; 1 + \norm{\pi_s(\theta)}^2 - 2 \pi_{s,a}(\theta) \; \leq \; 2,
\end{eqnarray}
where the inequality is obtained by
\begin{eqnarray} \label{eq:pi_s_1}
\norm{\pi_s(\theta)}^2 \; = \; \sum_{a \in \cA} \pi_{s,a}(\theta)^2 \; \leq \; \sum_{a \in \cA} \pi_{s,a}(\theta) \; = \; 1
\end{eqnarray} 
with $\pi_{s,a}(\theta) \in [0, 1]$.
This means, the softmax tabular policy satisfies~\eqref{eq:lipschitz_smooth_policy} condition with a bigger constant $G^2 = 2$ instead of $1 - \frac{1}{|\cA|}$ and $F = 1$.

Lemma~\ref{lem:softmax_expected} immediately implies that $J(\cdot)$ with the softmax tabular policy is smooth and Lipschitz as following.

\begin{lemma} \label{lem:softmax_smoothness}
$J(\cdot)$ with the softmax tabular policy is $\frac{\cR_{\max}}{(1-\gamma)^2}\left(2 - \frac{1}{|\cA|}\right)$-smooth and $\frac{\cR_{\max}}{(1-\gamma)^{3/2}}\sqrt{1 - \frac{1}{|\cA|}}$-Lipschitz.
\end{lemma}

\begin{proof}
From Lemma~\ref{lem:softmax_expected}, we know that Assumption~\ref{ass:lipschitz_smooth_policy} is satisfied with $G^2 = 1 - \frac{1}{|\cA|}$ and $F = 1$. Thus, $J(\cdot)$ with the softmax tabular policy is smooth and Lipschitz.

Indeed, from Lemma~\ref{lem:smoothJ}, we obtain the smoothness constant $\frac{\cR_{\max}}{(1-\gamma)^2}\left(2 - \frac{1}{|\cA|}\right)$ for $J(\cdot)$; and from Lemma~\ref{lem:lipschitzJ}~\ref{itm:J-lipschitz}, we obtain the Lipschitzness constant $\frac{\cR_{\max}}{(1-\gamma)^{3/2}}\sqrt{1 - \frac{1}{|\cA|}}$ for $J(\cdot)$.
\end{proof}

Now we can provide the formal statement of Corollary~\ref{cor:fosp_softmax}.

\begin{corollary}[Formal] \label{cor:fosp_softmax_formal}
For any accuracy level $\epsilon$, if we choose the mini-batch size $m$ such that $1 \leq m \leq \frac{2\nu}{\epsilon^2}$, the step size $\eta = \frac{\epsilon^2m}{2L\nu}$ with $L = \frac{\cR_{\max}}{(1-\gamma)^2}\left(2-\frac{1}{|\cA|}\right)$ and
\[
\nu \; = \;
			\begin{cases}
			\frac{H\left(1-\frac{1}{|\cA|}\right)\cR_{\max}^2}{(1-\gamma)^2} \quad &\mbox{for REINFORCE} \\
			\frac{\left(1-\frac{1}{|\cA|}\right)\cR_{\max}^2}{(1-\gamma)^3}  \quad &\mbox{for GPOMDP}
			\end{cases},
\]
the number of iterations $T$ such that
\begin{align}
    Tm \; \geq \; 
    \begin{cases}
    \frac{8\delta_0\cR_{\max}^3\left(1-\frac{1}{|\cA|}\right)\left(2-\frac{1}{|\cA|}\right)}{(1-\gamma)^4\epsilon^4} \cdot H \quad &\mbox{for REINFORCE} \\ 
    \frac{8\delta_0\cR_{\max}^3\left(1-\frac{1}{|\cA|}\right)\left(2-\frac{1}{|\cA|}\right)}{(1-\gamma)^5\epsilon^4} \quad &\mbox{for GPOMDP}
    \end{cases},
\end{align}
and the horizon $H = \cO\left((1-\gamma)^{-1}\log\left(1/\epsilon\right)\right)$, then $\E{\norm{\nabla J(\theta_U)}^2} = \cO(\epsilon^2)$. 
\end{corollary}

\begin{proof}
From Lemma~\ref{lem:softmax_smoothness}, we know that $L = \frac{\cR_{\max}}{(1-\gamma^2)}\left(2-\frac{1}{|\cA|}\right)$.

From Lemma~\ref{lem:ABC} and~\ref{lem:softmax_expected}, we know that
\[
\nu \; = \;
			\begin{cases}
			\frac{H\left(1-\frac{1}{|\cA|}\right)\cR_{\max}^2}{(1-\gamma)^2} \quad &\mbox{for REINFORCE} \\
			\frac{\left(1-\frac{1}{|\cA|}\right)\cR_{\max}^2}{(1-\gamma)^3}  \quad &\mbox{for GPOMDP}
			\end{cases}.
\]
Plugging in $L$ and $\nu$ in Corollary~\ref{cor:sample_complexity} yields the corollary's claim.
\end{proof}

\subsection{Stationary point convergence of the softmax tabular policy with log barrier regularization}
\label{sec:fosp:log}

First we provide the proof of Lemma~\ref{lem:ABC:log}.

\begin{proof}
Let $g(\tau\mid\theta)$ be a stochastic gradient estimator of one single sampled trajectory $\tau$. Thus $\hnabla_m J(\theta) = \frac{1}{m}\sum_{i=1}^m g(\tau_i\mid\theta)$. Both $\hnabla_m J(\theta)$ and $g(\tau\mid\theta)$ are unbiased estimators of $J_H(\theta)$.

From~\eqref{eq:barrier_trunc}, we have the following gradient estimator
\begin{align} \label{eq:barrier-nabla_temp1}
\hnabla_m L_{\lambda}(\theta) \; = \; \hnabla_m J(\theta) 
+ \frac{\lambda}{|\cA||\cS|}\sum_{s,a}\nabla_\theta \log \pi_{s,a}(\theta).
\end{align}
For a state $s \in \cS$, from~\eqref{eq:softmax_log_gradient}, we have
\begin{eqnarray} \label{eq:barrier-nabla_temp2}
\frac{\lambda}{|\cA||\cS|}\sum_{a \in \cA}\frac{\partial \log\pi_{s,a}(\theta)}{\partial \theta_{s}} &=& \frac{\lambda}{|\cA||\cS|}\sum_{a \in \cA} ({\bf 1}_a - \pi_s(\theta))  \nonumber \\ 
&=& \frac{\lambda {\bf 1}_{|\cA|}}{|\cA||\cS|} - \frac{\lambda \pi_s(\theta)}{|\cS|} \nonumber \\ 
&=& \frac{\lambda}{|\cS|}\left(\frac{{\bf 1}_{|\cA|}}{|\cA|} - \pi_s(\theta)\right),
\end{eqnarray}
where ${\bf 1}_{|\cA|} \in \R^{|\cA|}$ is a vector of all ones. Thus we have
\begin{eqnarray} \label{eq:barrier-nabla}
\hnabla_m L_{\lambda}(\theta) 
&\overset{\eqref{eq:barrier-nabla_temp1}+\eqref{eq:barrier-nabla_temp2}}{=}& \hnabla_m J(\theta) + \frac{\lambda}{|\cS|}\left(\frac{\ones}{|\cA|} - \colvec{\pi_{s}(\theta)}_{s \in \cS} \right),
\end{eqnarray}
where ${\bf 1} \in \R^{|\cS||\cA|}$ and 
\[
\colvec{\pi_{s}(\theta)}_{s \in \cS} \; = \; \left[ \pi_{s_1}(\theta) \ ; \ \cdots \ ; \ \pi_{s_{|\cS|}}(\theta)\right] \; \in \; \R^{|\cS||\cA|}
\]
is the stacking\footnote{Here vectors are columns by default, and given $x_1, \cdots, x_{|\cS|} \in \mathbb{R}^{|\cA|}$ we note $[x_1 \ ; \ \dots \ ; \ x_{|\cS|}]\in \R^{|\cS||\cA|}$ the (column) vector stacking the $x_i$'s on top of each other.} of the vectors $\pi_{s_i}(\theta)$.

Next, taking expectation on the trajectories, we have
\begin{eqnarray} \label{eq:barrier_ABC_temp1}
\E{\norm{\hnabla_m L_{\lambda}(\theta)}^2} &\overset{\eqref{eq:barrier-nabla}}{=}& \E{\norm{\hnabla_m J(\theta) 
+ \frac{\lambda}{|\cS|}\left(\frac{\ones}{|\cA|} - \colvec{\pi_{s}(\theta)}_{s \in \cS} \right)}^2} \nonumber \\ 
&=& \E{\norm{\nabla J_H(\theta) + \frac{\lambda}{|\cS|}\left(\frac{\ones}{|\cA|} - \colvec{\pi_{s}(\theta)}_{s \in \cS} \right) 
+ \hnabla_m J(\theta) - \nabla J_H(\theta)}^2} \nonumber \\ 
&=& \norm{\nabla L_{\lambda, H}(\theta)}^2 + \E{\norm{\hnabla_m J(\theta) - \nabla J_H(\theta)}^2} \nonumber \\ 
&\overset{\eqref{eq:stochgradbnd}}{=}& \norm{\nabla L_{\lambda, H}(\theta)}^2 
+ \frac{\E{\norm{g(\tau_1 \mid \theta) - \nabla J_H(\theta)}^2}}{m} \nonumber \\ 
&=& \norm{\nabla L_{\lambda, H}(\theta)}^2 \nonumber \\ 
&\quad& \ + \frac{\E{\norm{g(\tau_1 \mid \theta) + \frac{\lambda}{|\cS|}\left(\frac{\ones}{|\cA|} - \colvec{\pi_{s}(\theta)}_{s \in \cS} \right) 
- \nabla J_H(\theta) - \frac{\lambda}{|\cS|}\left(\frac{\ones}{|\cA|} - \colvec{\pi_{s}(\theta)}_{s \in \cS} \right)}^2}}{m} \nonumber \\
&=& \left(1-\frac{1}{m}\right)\norm{\nabla L_{\lambda, H}(\theta)}^2 + \frac{\E{\norm{g(\tau_1 \mid \theta) 
+ \frac{\lambda}{|\cS|}\left(\frac{\ones}{|\cA|} - \colvec{\pi_{s}(\theta)}_{s \in \cS} \right)}^2}}{m} \nonumber \\ 
&\leq& \left(1-\frac{1}{m}\right)\norm{\nabla L_{\lambda, H}(\theta)}^2 + \frac{2\E{\norm{g(\tau_1 \mid \theta)}^2} 
+ 2\norm{\frac{\lambda}{|\cS|}\left(\frac{\ones}{|\cA|} - \colvec{\pi_{s}(\theta)}_{s \in \cS} \right)}^2}{m}.
\end{eqnarray}
In particular, we have
\begin{eqnarray} \label{eq:barrier_ABC_temp2}
\norm{\frac{\lambda}{|\cS|}\left(\frac{\ones}{|\cA|} - \colvec{\pi_{s}(\theta)}_{s \in \cS} \right)}^2 
\; \leq \; \frac{\lambda^2}{|\cS|^2}
\left(\frac{|\cS||\cA|}{|\cA|^2} - 2\frac{|\cS|}{|\cA|} + |\cS|\right)
\; = \; \frac{\lambda^2}{|\cS|}\left(1 - \frac{1}{|\cA|}\right),
\end{eqnarray}
where the inequality is obtained by using $\norm{\pi_s(\theta)}^2 \leq 1$ in~\eqref{eq:pi_s_1}.

As for $\E{\norm{g(\tau_1 \mid \theta)}^2}$, if $\hnabla_m J(\theta)$ is a REINFORCE gradient estimator, from~\eqref{eq:tempREINFORCE}, we have
\begin{align} \label{eq:barrier_ABC_temp3}
\E{\norm{g(\tau_1 \mid \theta)}^2} \; \leq \; \frac{HG^2\cR_{\max}^2}{(1-\gamma)^2} \; = \; \frac{H\cR_{\max}^2\left(1 - \frac{1}{|\cA|}\right)}{(1-\gamma)^2},
\end{align}
where the equality is obtained by Lemma~\ref{lem:softmax_expected} with $G^2 = \left(1 - \frac{1}{|\cA|}\right)$.

Combining~\eqref{eq:barrier_ABC_temp1},~\eqref{eq:barrier_ABC_temp2} and~\eqref{eq:barrier_ABC_temp3}, we have that the REINFORCE gradient estimator $\hnabla_m L_\lambda(\theta)$ satisfies~\eqref{eq:ABC} assumption with
\[\E{\norm{\hnabla_m L_\lambda(\theta)}^2} \; \leq \; \left(1-\frac{1}{m}\right)\norm{\nabla L_{\lambda, H}(\theta)}^2 + \frac{2}{m}\left(1 - \frac{1}{|\cA|}\right)\left(\frac{H\cR_{\max}^2}{(1-\gamma)^2} + \frac{\lambda^2}{|\cS|}\right).\]
If $\hnabla_m J(\theta)$ is a GPOMDP gradient estimator, from~\eqref{eq:tempGPOMDP}, we have
\begin{align} \label{eq:barrier_ABC_temp4}
\E{\norm{g(\tau_1 \mid \theta)}^2} \; \leq \; \frac{G^2\cR_{\max}^2}{(1-\gamma)^3} \; = \; \frac{\cR_{\max}^2\left(1 - \frac{1}{|\cA|}\right)}{(1-\gamma)^3}.
\end{align}
Combining~\eqref{eq:barrier_ABC_temp1},~\eqref{eq:barrier_ABC_temp2} and~\eqref{eq:barrier_ABC_temp4}, we have that the GPOMDP gradient estimator $\hnabla_m L_\lambda(\theta)$ satisfies~\eqref{eq:ABC} assumption with
\[\E{\norm{\hnabla_m L_\lambda(\theta)}^2} \; \leq \; \left(1-\frac{1}{m}\right)\norm{\nabla L_{\lambda, H}(\theta)}^2 + \frac{2}{m}\left(1 - \frac{1}{|\cA|}\right)\left(\frac{\cR_{\max}^2}{(1-\gamma)^3} + \frac{\lambda^2}{|\cS|}\right).\]
Thus $\hnabla_m L_\lambda(\theta)$ satisfies the~\eqref{eq:ABC} assumption for both REINFORCE and GPOMDP gradient estimators, which concludes the proof.
\end{proof}

We also verify that $L_{\lambda}(\cdot)$ is smooth and Lipschitz in the following lemma.

\begin{lemma} \label{lem:smooth:log}
$L_{\lambda}(\cdot)$ is $\left(\frac{\cR_{\max}}{(1-\gamma)^2}\left(2 - \frac{1}{|\cA|}\right) + \frac{\lambda}{|\cS|}\right)$-smooth and $\sqrt{2\left(1-\frac{1}{|\cA|}\right)\left(\frac{\cR_{\max}^2}{(1-\gamma)^3} + \frac{\lambda^2}{|\cS|}\right)}$-Lipschitz.
\end{lemma}

\begin{proof}
For the smoothness constant, first, from Lemma~\ref{lem:softmax_smoothness}, we know that $J(\cdot)$ is $\frac{\cR_{\max}}{(1-\gamma)^2}\left(2 - \frac{1}{|\cA|}\right)$-smooth.

It remains to show the regularizer $R(\theta) \eqdef \frac{\lambda}{|\cA||\cS|}\sum_{s,a}\log\pi_\theta(a \mid s)$ is $\frac{\lambda}{|\cS|}$-smooth. From~\eqref{eq:barrier-nabla}, we have
\[\nabla R(\theta) \; = \; \frac{\lambda}{|\cS|}\left(\frac{\ones}{|\cA|} - \colvec{\pi_{s}(\theta)}_{s \in \cS} \right).\]
From~\eqref{eq:softmax_Jacobian}, we have
\[\norm{\frac{\partial^2 R(\theta)}{\partial \theta_s^2}} \; = \; \norm{- \frac{\lambda}{|\cS|}\mH(\pi_s(\theta))} \; \leq \; \frac{\lambda}{|\cS|},\]
where the inequality is obtained by using $\norm{\mH(\pi_s(\theta))} \leq 1$ from Lemma 22 in~\citet{mei2020ontheglobal}.

Since $\frac{\partial^2 R(\theta)}{\partial \theta_s \partial \theta_{s'}} = 0$ for $s \neq s'$, we have that $\norm{\nabla^2 R(\theta)} \leq \frac{\lambda}{|\cS|}$, which yields the smoothness constant of $L_\lambda(\cdot)$.

For the Lipschitzness constant, from~\eqref{eq:barrier-nabla}, we know that
\begin{eqnarray}
\norm{\nabla L_\lambda(\theta)}^2 &=& \norm{\nabla J(\theta) + \frac{\lambda}{|\cS|}\left(\frac{\ones}{|\cA|} - \colvec{\pi_{s}(\theta)}_{s \in \cS} \right)}^2 \nonumber \\ 
&\leq& 2\norm{\nabla J(\theta)}^2 + 2\norm{\frac{\lambda}{|\cS|}\left(\frac{\ones}{|\cA|} - \colvec{\pi_{s}(\theta)}_{s \in \cS} \right)}^2 \nonumber \\ 
&\overset{\mbox{Lemma~\ref{lem:softmax_smoothness}}}{\leq}& 2\left(1-\frac{1}{|\cA|}\right)\frac{\cR_{\max}^2}{(1-\gamma)^3} 
+ 2\norm{\frac{\lambda}{|\cS|}\left(\frac{\ones}{|\cA|} - \colvec{\pi_{s}(\theta)}_{s \in \cS} \right)}^2 \nonumber \\ 
&\overset{\eqref{eq:barrier_ABC_temp2}}{\leq}& 2\left(1-\frac{1}{|\cA|}\right)\frac{\cR_{\max}^2}{(1-\gamma)^3} 
+ \frac{2\lambda^2}{|\cS|}\left(1-\frac{1}{|\cA|}\right) \nonumber \\
&=& 2\left(1-\frac{1}{|\cA|}\right)\left(\frac{\cR_{\max}^2}{(1-\gamma)^3} + \frac{\lambda^2}{|\cS|}\right).
\end{eqnarray}
Thus,
\[\norm{\nabla L_\lambda(\theta)} \; \leq \; \sqrt{2\left(1-\frac{1}{|\cA|}\right)\left(\frac{\cR_{\max}^2}{(1-\gamma)^3} + \frac{\lambda^2}{|\cS|}\right)}.\]
\end{proof}

\paragraph*{The truncated gradient assumption in the case of $L_{\lambda,H}(\cdot)$.}
As $L_\lambda(\theta)$ and $L_{\lambda, H}(\theta)$ use the same regularizer, the bias due to the truncation does not affect the regularization. Besides, from Lemma~\ref{lem:softmax_expected}, we have that Assumption~(\nameref{E-LS}) holds. Thus, from Lemma~\ref{lem:trunc}, Assumption~\ref{ass:trunc} holds for $L_\lambda(\theta)$ and $L_{\lambda, H}(\theta)$ with the same constant $D$ and $D'$ in Lemma~\ref{lem:trunc} and the constant $G$ in Lemma~\ref{lem:softmax_expected}. That is,
\begin{eqnarray}
\left| \dotprod{\nabla L_{\lambda, H}(\theta), L_{\lambda, H}(\theta) - L_\lambda(\theta)} \right| &\leq& D\gamma^H, \\
\norm{\nabla L_{\lambda, H}(\theta) - L_\lambda(\theta)} &\leq& D'\gamma^H,
\end{eqnarray}
with
\begin{eqnarray}
D &=& \frac{D'\cR_{\max}}{(1-\gamma)^{3/2}}\sqrt{1 - \frac{1}{|\cA|}}, \\
D' &=& \frac{\cR_{\max}}{1-\gamma}\sqrt{\left(\frac{1}{1-\gamma} + H\right)\left(1 - \frac{1}{|\cA|}\right)}.
\end{eqnarray}

Similar to Corollary~\ref{cor:fosp_softmax_formal}, now we can provide the FOSP convergence of $L_\lambda(\theta)$.

\begin{corollary} \label{cor:fosp:log}
Consider the vanilla PG (either REINFORCE or GPOMDP) applied in $L_\lambda(\cdot)$. Let $\delta_0 \eqdef L_\lambda^* - L_\lambda(\theta_0)$ with $L_\lambda^* \eqdef \max_{\theta \in \R^d} L_\lambda(\theta)$.
For any accuracy level $\epsilon$, if we choose the mini-batch size $m$ such that  $1 \leq m \leq \frac{2\nu}{\epsilon^2}$, the step size $\eta = \frac{\epsilon^2m}{2L\nu}$ with $L = \frac{\cR_{\max}}{(1-\gamma)^2}\left(2-\frac{1}{|\cA|}\right) + \frac{\lambda}{|\cS|}$ and
\begin{align} \label{eq:log:nu}
\nu \; = \;
 	\begin{cases} 
 	2\left(1-\frac{1}{|\cA|}\right)\left(\frac{H\cR_{\max}^2}{(1-\gamma)^2} + \frac{\lambda^2}{|\cS|}\right) \quad &\mbox{when using REINFORCE}  \\
 	2\left(1-\frac{1}{|\cA|}\right)\left(\frac{\cR_{\max}^2}{(1-\gamma)^3} + \frac{\lambda^2}{|\cS|}\right)  \quad &\mbox{when using GPOMDP}
 	\end{cases},
\end{align}
the number of iterations $T$ such that
\begin{align}
    Tm \; \geq \; \frac{8\delta_0L\nu}{\epsilon^4} \; = \; \cO((1-\gamma)^{-5}\epsilon^{-4}),
\end{align}
and the horizon $H = \cO\left((1-\gamma)^{-1}\log\left(1/\epsilon\right)\right)$, then $\E{\norm{\nabla L_\lambda(\theta_U)}^2} = \cO(\epsilon^2)$. 
\end{corollary}

\begin{proof}
From Lemma~\ref{lem:smooth:log}, we know that $L = \frac{\cR_{\max}}{(1-\gamma)^2}\left(2-\frac{1}{|\cA|}\right) + \frac{\lambda}{|\cS|}$.

From Lemma~\ref{lem:ABC:log}, we know that
\[
\nu \; = \;
 	\begin{cases} 
 	2\left(1-\frac{1}{|\cA|}\right)\left(\frac{H\cR_{\max}^2}{(1-\gamma)^2} + \frac{\lambda^2}{|\cS|}\right) \quad &\mbox{when using REINFORCE}  \\
 	2\left(1-\frac{1}{|\cA|}\right)\left(\frac{\cR_{\max}^2}{(1-\gamma)^3} + \frac{\lambda^2}{|\cS|}\right)  \quad &\mbox{when using GPOMDP}
 	\end{cases}.
\]
Plugging in $L$ and $\nu$ in Corollary~\ref{cor:sample_complexity} yields the corollary's claim.
\end{proof}

\subsection{Sample complexity of high probability global optimum convergence for the softmax tabular policy with log barrier regularization}
\label{sec:high_proba}

In this section, we provide the sample complexity to reach a global optimum convergence of the expected return $J(\cdot)$ in  high probability for the softmax tabular policy with log barrier regularization.

Before the results, we introduce the stationary distribution 
\[d_{\rho, s}(\pi^*) \; \eqdef \; \EE{s_0 \sim \rho(\cdot), \tau \sim p(\cdot \mid \pi^*)}{(1-\gamma)\sum_{t=0}^\infty\gamma^t\mathbb{P}\left(s_t = s\right)},\]
where $\pi^*$ is the optimal policy. We refer to $\norm{\frac{d_\rho(\pi^*)}{\rho}}_{\infty} \eqdef \max_{s \in \cS} \frac{d_{\rho, s}(\pi^*)}{\rho(s)}$ as the distribution mismatch coefficient of $\pi$ under $\rho$~\citep{agarwal2021theory}\footnote{For simplicity, we assume that the sampling for the initial state distribution is the same as the initial state distribution appeared in the expected return $J(\cdot)$. There is no difference, compared to our results, to impose a different initial state distribution $\mu \neq \rho$ for the stochastic vanilla PG. In this case, the distribution mismatch coefficient will be $\norm{\frac{d_\rho(\pi^*)}{\mu}}_{\infty}$.}. We assume that the initial state distribution $\rho$ satisfies $\min_s\rho(s) > 0$. This assumption was adapted by~\citet{agarwal2021theory} to ensure that the distribution mismatch coefficient is finite.

\begin{corollary} \label{cor:log:concentration}
For any accuracy level $\epsilon > 0$, any probability accuracy level $\delta \in (0, 1)$ and any starting state distribution $\rho$, 
consider the vanilla PG (either REINFORCE or GPOMDP) applied to $L_\lambda(\cdot)$. If we chose the horizon $H = \cO\left((1-\gamma)^{-1}\log\left(1/\epsilon_{opt}\right)\log\left(1/\delta\right)\right)$, the batch size  $1 \leq m \leq \frac{2\nu}{\delta\epsilon_{opt}^2}$ and the number of iterations $T$ such that $Tm \geq \frac{8(L_\lambda^* - L_\lambda(\theta_0))L\nu}{\delta^2\epsilon_{opt}^4}$, the regularization parameter $\lambda = \frac{(1-\gamma)\epsilon}{2\norm{\frac{d_\rho(\pi^*)}{\rho}}_{\infty}}$ and
\begin{eqnarray}
\epsilon_{opt} \; = \; \frac{\lambda}{2|\cS||\cA|} \; = \; \frac{(1-\gamma)\epsilon}{4|\cS||\cA|\norm{\frac{d_\rho(\theta^*)}{\rho}}_{\infty}}
\end{eqnarray}
with $L, \nu$ in the setting of Corollary~\ref{cor:fosp:log}, then 
we have an upper bound of the sample complexity
\begin{eqnarray}
Tm \times H = \cO\left(\frac{|\cS|^4|\cA|^4\norm{\frac{d_\rho(\theta^*)}{\rho}}_{\infty}^4}{\delta^2\epsilon^4(1-\gamma)^{10}} \cdot \log\left(1/\epsilon\right)\log\left(1/\delta\right)\right)
\end{eqnarray}
 guarantees that $J^* - J(\theta_T) \leq \epsilon$ with probability at least $1-\delta$.
%Let
%\begin{align}
%C(\delta, \epsilon) \; = \; 
%\min_{T, m \geq 0}\left\{Tm \; \big| \; \mathbb{P}\left(J^* - J(\theta_T) \geq \epsilon\right) \leq \delta\right\}.
%\end{align}
%Then we have the following sample complexity
%$
%C(\delta, \epsilon) %\; \leq \; \frac{(J^* - J(\theta_0))L\nu|\cS|^4|\cA|^4}{\delta^2 \cdot \lambda^4} 
%\; = \; \cO\left(\frac{|\cS|^4|\cA|^4\norm{\frac{d_\rho(\theta^*)}{\rho}}_{\infty}^4}{\delta^2\epsilon^4(1-\gamma)^9}\right).
%$
\end{corollary}
The above high probability global optimum  sample complexity holds with a wide range of parameters (e.g. batch size $m$ and step size $\eta$) thanks to Corollary~\ref{cor:fosp:log}.

We need the following result to link the stationary point convergence of $L_\lambda(\cdot)$ to the suboptimality gap convergence $J^* - J(\cdot)$ when the norm of the gradient of a stationary point and the regularization parameter $\lambda$ are sufficiently small.

\begin{proposition}[Theorem 5.2 in~\citet{agarwal2021theory}] \label{pro:agarwal}
Suppose $\theta$ is such that $\norm{\nabla L_{\lambda}(\theta)} \leq \frac{\lambda}{2|\cS||\cA|}$, then for every initial distribution $\rho$, we have
\begin{eqnarray}
J^* - J(\theta) \; \leq \; \frac{2\lambda}{1-\gamma}\norm{\frac{d_\rho(\theta^*)}{\rho}}_{\infty}.
\end{eqnarray}
\end{proposition}

By leveraging Proposition~\ref{pro:agarwal}, we now derive the proof for Corollary~\ref{cor:log:concentration}.

\begin{proof}
From Corollary~\ref{cor:fosp:log} we have that $\E{\norm{\nabla L_\lambda(\theta_U)}^2} \leq \delta\epsilon_{opt}^2$,

Thus, there exists $t_0 \in \{0, \cdots, T-1\}$ s.t. $\E{\norm{\nabla L_\lambda(\theta_{t_o})}^2} \leq \E{\norm{\nabla L_\lambda(\theta_U)}^2} \leq \delta\epsilon_{opt}^2$.

From Proposition~\ref{pro:agarwal}, we know that if $\norm{\nabla L_\lambda(\theta_{t_o})} \leq \epsilon_{opt}$, we have
\[J^* - J(\theta_{t_0}) \; \leq \; \frac{2\lambda}{1-\gamma}\norm{\frac{d_\rho(\theta^*)}{\rho}}_{\infty} \; = \; \epsilon.\]
Thus, we have
\begin{eqnarray} \label{eq:proba:J*-J}
\mathbb{P}(J^* - J(\theta_{t_0}) \leq \epsilon) \; \geq \; \mathbb{P}\left(\norm{\nabla L_\lambda(\theta_{t_o})} \leq \epsilon_{opt}\right).
\end{eqnarray}
Consequently, we have
\begin{eqnarray}
\mathbb{P}(J^* - J(\theta_{t_0}) \geq \epsilon) &=& 1 - \mathbb{P}(J^* - J(\theta_{t_0}) \leq \epsilon) \nonumber \\ 
&\overset{\eqref{eq:proba:J*-J}}{\leq}& 1 - \mathbb{P}\left(\norm{\nabla L_\lambda(\theta_{t_o})} \leq \epsilon_{opt}\right) \nonumber \\
&=& \mathbb{P}\left(\norm{\nabla L_\lambda(\theta_{t_o})} \geq \epsilon_{opt}\right) \nonumber \\ 
&=& \mathbb{P}\left(\norm{\nabla L_\lambda(\theta_{t_o})}^2 \geq \epsilon_{opt}^2\right) \nonumber \\ 
&\leq& \frac{\E{\norm{\nabla L_\lambda(\theta_{t_o})}^2}}{\epsilon_{opt}^2} \quad \quad \mbox{(by Markov's inequality)} \nonumber \\ 
&\leq& \delta.
\end{eqnarray}
Since $t_0 m \leq Tm$, we conclude that the upper bound of the sample complexity is
\[
Tm \times H \; \geq \; \frac{8(J^* - J(\theta_0))L\nu}{\delta^2\epsilon_{opt}^4} \times H \; = \; \cO\left(\frac{|\cS|^4|\cA|^4\norm{\frac{d_\rho(\theta^*)}{\rho}}_{\infty}^4}{\delta^2\epsilon^4(1-\gamma)^{10}}
\cdot \log\left(1/\epsilon\right)\log\left(1/\delta\right)\right).
\]
\end{proof}

\paragraph*{Remark.}
Following the proof of Corollary~\ref{cor:log:concentration}, we can also deduce the iteration complexity of the exact full gradient updates for the global optimum convergence.

Indeed, from Lemma~\ref{lem:smooth:log}, $L_\lambda(\cdot)$ is smooth. From Theorem~\ref{pro:ABC}, we know that with the number of iterations
\begin{align} \label{eq:log_iteration_complexity}
T \; \geq \; \frac{12\delta_0L}{\epsilon_{opt}^2} \; = \; \cO\left(\frac{\delta_0}{(1-\gamma)^4\epsilon^2}\right),
\end{align}
we have $\min_{0 \leq t \leq T-1} \norm{\nabla L_\lambda(\theta_t)}^2 \leq \epsilon_{opt}^2$ for the exact full gradient updates.

From Proposition~\ref{pro:agarwal}, we have $\min_{0 \leq t \leq T-1} J^* - J(\theta_t) \leq \epsilon$. 

Compared to the iteration complexity in Corollary 5.1 in~\citet{agarwal2021theory}, ours~\eqref{eq:log_iteration_complexity} is improved by a factor of $1-\gamma$ thanks to an improved analysis of the smoothness constant in Lemma~\ref{lem:smooth:log}.

\subsection{Sample complexity of the average regret convergence for softmax tabular policy with log barrier regularization}

By leveraging Proposition~\ref{pro:agarwal}, we now derive the proof for Corollary~\ref{cor:regret_log}.

\begin{proof}
We define the following set of "bad" iterates based on a technique developed by~\citet{zhang2020sample}
\begin{eqnarray}
I^+ &\eqdef& \left\{t \in \{0, \cdots, T-1\} \ \bigg| \ \norm{\nabla L_\lambda(\theta_t)} \; \geq \; \frac{\lambda}{2|\cS||\cA|} \right\}
\end{eqnarray}
with
\begin{eqnarray} \label{eq:lambda}
\lambda &=& \frac{(1-\gamma)\epsilon}{2\norm{\frac{d_\rho(\theta^*)}{\mu}}_{\infty}}.
\end{eqnarray}

We have
\begin{eqnarray}
J^* - \frac{1}{T}\sum_{t=0}^{T-1}J(\theta_t) &=& \frac{1}{T}\sum_{t \in I^+} J^* - J(\theta_t) + \frac{1}{T}\sum_{t \notin I^+} J^* - J(\theta_t) \nonumber \\ 
&\leq& \frac{|I^+|}{T} \cdot \frac{2\cR_{\max}}{1-\gamma} + \frac{1}{T}\sum_{t \notin I^+} J^* - J(\theta_t) \nonumber \\ 
&\leq& \frac{|I^+|}{T} \cdot \frac{2\cR_{\max}}{1-\gamma} + \frac{T - |I^+|}{T} \cdot \frac{2\lambda}{1-\gamma}\norm{\frac{d_\rho(\theta^*)}{\rho}}_{\infty} \nonumber \\ 
&\leq& \frac{|I^+|}{T} \cdot \frac{2\cR_{\max}}{1-\gamma} + \frac{2\lambda}{1-\gamma}\norm{\frac{d_\rho(\theta^*)}{\rho}}_{\infty} \nonumber \\ 
&\overset{\eqref{eq:lambda}}{=}& \frac{|I^+|}{T} \cdot \frac{2\cR_{\max}}{1-\gamma} + \epsilon. \label{eq:J*-1/TJ}
\end{eqnarray}
where the second line is obtained as $|J(\cdot)| \leq \frac{\cR_{\max}}{1-\gamma}$ and the third line is obtained by Proposition~\ref{pro:agarwal}.

It remains to bound $|I^+|$. In fact,
\begin{eqnarray}
\sum_{t=0}^{T-1}\norm{\nabla L_\lambda(\theta_t)}^2 &\geq& \sum_{t \in I^+}\norm{\nabla L_\lambda(\theta_t)}^2 \nonumber \\
&\geq& \frac{|I^+|\lambda^2}{4|\cS|^2|\cA|^2}.
\end{eqnarray}
Thus, we have
\begin{eqnarray}
\frac{|I^+|}{T} &\leq& \frac{4|\cS|^2|\cA|^2}{\lambda^2} \cdot \frac{1}{T}\sum_{t=0}^{T-1}\norm{\nabla L_\lambda(\theta_t)}^2 \nonumber \\ 
&\overset{\eqref{eq:lambda}}{=}& \frac{16\norm{\frac{d_\rho(\theta^*)}{\mu}}_{\infty}^2|\cS|^2|\cA|^2}{(1-\gamma)^2\epsilon^2} \cdot \frac{1}{T}\sum_{t=0}^{T-1}\norm{\nabla L_\lambda(\theta_t)}^2. \label{eq:I+}
\end{eqnarray}
Thus, we have
\begin{eqnarray}
J^* - \frac{1}{T}\sum_{t=0}^{T-1}J(\theta_t) \; \overset{\eqref{eq:J*-1/TJ}+\eqref{eq:I+}}{\leq} \; \frac{32\cR_{\max}\norm{\frac{d_\rho(\theta^*)}{\rho}}_{\infty}^2|\cS|^2|\cA|^2}{(1-\gamma)^3\epsilon^2} \cdot \frac{1}{T}\sum_{t=0}^{T-1}\norm{\nabla L_\lambda(\theta_t)}^2 + \epsilon.
\end{eqnarray}
Taking expectation over the iterations on both side, we have
\begin{eqnarray}
J^* - \frac{1}{T}\sum_{t=0}^{T-1}\E{J(\theta_t)} \; \overset{\eqref{eq:J*-1/TJ}+\eqref{eq:I+}}{\leq} \; \frac{32\cR_{\max}\norm{\frac{d_\rho(\theta^*)}{\rho}}_{\infty}^2|\cS|^2|\cA|^2}{(1-\gamma)^3\epsilon^2} \cdot \frac{1}{T}\sum_{t=0}^{T-1}\E{\norm{\nabla L_\lambda(\theta_t)}^2} + \epsilon.
\end{eqnarray}
It suffices to have $\frac{1}{T}\sum_{t=0}^{T-1}\E{\norm{\nabla L_\lambda(\theta_t)}^2} \leq (1-\gamma)^3\epsilon^3$ to guarantee that $J^* - \frac{1}{T}\sum_{t=0}^{T-1}\E{J(\theta_t)} \leq \cO(\epsilon)$.

From Corollary~\ref{cor:sample_complexity}, consider the batch size $m$ such that $1 \leq m \leq \frac{2\nu}{(1-\gamma)^3\epsilon^3} = \cO\left(\frac{1}{(1-\gamma)^6\epsilon^3}\right)$, the step size $\cO(\epsilon^3) \leq \eta = \frac{(1-\gamma)^3\epsilon^3m}{2L\nu} \leq \cO(1)$ with $L, \nu$ in the setting of Corollary~\ref{cor:fosp:log}
%, the horizon $H = \cO\left(\frac{\log(1/\epsilon)}{1-\gamma}\right)$ and the number of iterations $T$ such that
. If the horizon $H = \cO\left(\frac{\log(1/\epsilon)}{1-\gamma}\right)$ and the number of iterations $T$ is such that
\[
%Tm \; \geq \; \cO\left(\frac{|\cS|^4|\cA|^4\norm{\frac{d_\rho(\theta^*)}{\rho}}_{\infty}^4}{(1-\gamma)^{11}\epsilon^6}\right),
Tm \times H \; \geq \; \frac{8(J^* - J(\theta_0))L\nu}{(1-\gamma)^6\epsilon^6} \times H \; = \; \widetilde{\cO}\left(\frac{1}{(1-\gamma)^{12}\epsilon^6}\right),
\]
we have $\frac{1}{T}\sum_{t=0}^{T-1}\E{\norm{\nabla L_\lambda(\theta_t)}^2} \leq (1-\gamma)^3\epsilon^3$, which conclude the proof.
\end{proof}

\section{Proof of Section~\ref{sec:FI}}
\label{sec:proof_FI}

First, we give the definition of the advantage function $A^{\pi_\theta}$ induced by the policy $\pi_\theta$ appeared in the transferred compatible function approximation error in Assumption~\ref{ass:compatible}. To do this, given a policy $\pi$, we define the state-action value function $Q^\pi: \cS \times \cA \rightarrow \R$ as
\[ Q^\pi(s,a) \eqdef \EE{a_t \sim \pi(\cdot \mid s_t), s_{t+1} \sim \cP(\cdot \mid s_t, a_t)}{\sum_{t=0}^\infty\gamma^t\cR(s_t, a_t) \ \bigg | \ s_0 = s, a_0 = a }.\]
From this, the state-value function $V^\pi: \cS \rightarrow \R$ and the advantage function $A^\pi: \cS \times \cA \rightarrow \R$, under the policy $\pi$, can be defined as
\begin{align*}
V^\pi(s) &\eqdef \EE{a \sim \pi(\cdot \mid s)}{Q^\pi(s,a)}, \\ 
A^\pi(s,a) &\eqdef Q^\pi(s,a) - V^\pi(s).
\end{align*}
Before presenting the sample complexity of the average regret convergence and the proof of Corollary~\ref{cor:FI} for Fisher-non-degenerate parametrized policy, we need the following result to show that Fisher-non-degenerate parametrized policy satisfies the relaxed weak gradient domination assumption.
\begin{proposition}[Lemma~4.7 in~\citet{ding2021global}] \label{pro:ding}
If the policy $\pi_\theta$ satisfies Assumption~\ref{ass:lipschitz_smooth_policy},~\ref{ass:FI} and~\ref{ass:compatible}, then
\begin{align}
\frac{\mu_F \sqrt{\epsilon_{bias}}}{(1-\gamma)G} + \norm{\nabla J_H(\theta)} \geq \frac{\mu_F}{G}(J^* - J(\theta)).
\end{align}
\end{proposition}
\paragraph*{Remark.} Here we use the weaker assumption~(\nameref{E-LS}) instead of~\eqref{eq:lipschitz_smooth_policy} compared to the original Lemma~4.7 in~\citet{ding2021global}. The relaxed weak gradient domination property still holds. The proof essentially follows the same arguments and thus is omitted here.

\subsection{Sample complexity of the average regret convergence for Fisher-non-degenerate policy}
\label{sec:average}

Consequently, it is straightforward to obtain the average regret to the global optimum convergence under the setting of Corollary~\ref{cor:sample_complexity} for Fisher-non-degenerate parametrized policy.

\begin{corollary} \label{cor:FI_regret}
%Suppose that Asm.~\ref{ass:lipschitz_smooth_policy} is satisfied.
Assume that the policy $\pi_\theta$ satisfies Asm.~\ref{ass:lipschitz_smooth_policy},~\ref{ass:FI} and~\ref{ass:compatible}.
 For a given $\epsilon>0$, by choosing the mini-batch size $m$ such that $1 \leq m \leq \frac{2\nu}{\epsilon^2}$, the step size $\eta = \frac{\epsilon^2m}{2L\nu}$, the number of iterations $T$ such that
\begin{align}
    Tm \geq \frac{8\delta_0L\nu}{\epsilon^4} =  
    \begin{cases}
    \cO\left(\frac{H}{(1-\gamma)^4\epsilon^4}\right) \quad \mbox{for REINFORCE} \\ 
    \cO\left(\frac{1}{(1-\gamma)^5\epsilon^4}\right) \quad \mbox{for GPOMDP}
    \end{cases}
\end{align}
and the horizon $H = \cO\left((1-\gamma)^{-1}\log\left(1/\epsilon\right)\right)$, then $J^* - \frac{1}{T}\sum_{t=0}^{T-1} \E{J(\theta_t)} = \cO(\epsilon) + \cO(\sqrt{\epsilon_{bias}})$. 
\end{corollary}

\paragraph{Remark.} The sample complexity $\widetilde{\cO}(\epsilon^{-4})$ of the average regret is also shown in Theorem 4.6 in~\citet{liu2020animproved}. However,~\citet{liu2020animproved} use the more restrictive assumption~\eqref{eq:lipschitz_smooth_policy} and require large batch size $m = \cO(\epsilon^{-2})$. We improve upon them by using weaker assumption~\nameref{E-LS},
allowing much wider range of choices for the batch size $m \in \left[1;\frac{2\nu}{\epsilon^2}\right]$ and the constant step size $\eta$ to achieve the same optimal sample complexity $\widetilde{\cO}\left(\epsilon^{-4}\right)$.

\begin{proof}
From Corollary~\ref{cor:sample_complexity}, we know that $\E{\norm{\nabla J(\theta_U)}^2} = \cO(\epsilon^2)$. However, from Proposition~\ref{pro:ding}, we know that Assumption~\ref{ass:weak} is satisfied. Thus, by doing a similar analysis as in Corollary~\ref{cor:regret}, we conclude the proof.
\end{proof}

\subsection{Proof of Corollary~\ref{cor:FI}}

Now we provide the proof of Corollary~\ref{cor:FI}.

\begin{proof}
From Proposition~\ref{pro:ding}, we have that Assumption~\ref{ass:weak} holds. Also because of Assumption~(\nameref{E-LS}), we have Lemmas~\ref{lem:ABC},~\ref{lem:smoothJ} and~\ref{lem:trunc} hold. Finally, by Corollary~\ref{cor:weak}, this directly concludes the proof.
\end{proof}

\section{FOSP convergence analysis for the softmax with entropy regularization.}
\label{sec:entropy}

In this section, we study stochastic gradient ascent on the softmax tabular policy with entropy regularization, which is
\begin{align} \label{eq:entropy}
\tilde{J}(\theta) \; \eqdef \; J(\theta) + \mathbb{H}(\theta)
\end{align}
where $\mathbb{H}(\theta)$ is the ``discounted entropy'' defined as
\[\mathbb{H}(\theta) \; \eqdef \; \EE{\tau \sim p(\cdot \mid \theta)}{\sum_{t=0}^\infty - \gamma^t \lambda \log\pi_{s_t,a_t}(\theta)}.\]

Using the same technique to derive the full gradient of the expected return~\eqref{eq:GD}, we have
\begin{eqnarray} \label{eq:entropy-nabla}
\nabla\tilde{J}(\theta) &=& \nabla J(\theta) - \lambda \EE{\tau}{\nabla \log p(\tau \mid \theta) \sum_{t=0}^\infty \gamma^t \log \pi_{s_t, a_t}(\theta)} - \lambda \EE{\tau}{\sum_{t=0}^\infty \gamma^t \nabla_\theta \log \pi_{s_t, a_t}(\theta)} \nonumber \\
&\overset{\eqref{eq:p}}{=}& \nabla J(\theta) - \lambda \EE{\tau}{\sum_{k=0}^\infty \nabla_\theta \log \pi_{s_k, a_k}(\theta) \sum_{t=0}^\infty \gamma^t \log \pi_{s_t, a_t}(\theta)} - \lambda \EE{\tau}{\sum_{t=0}^\infty \gamma^t \nabla_\theta \log \pi_{s_t, a_t}(\theta)} \nonumber \\ 
&=& \nabla J(\theta) - \lambda \EE{\tau}{\sum_{t=0}^\infty \gamma^t \log \pi_{s_t, a_t}(\theta) \left(\sum_{k=0}^t \nabla_\theta \log \pi_{s_k, a_k}(\theta)\right)} - \lambda \EE{\tau}{\sum_{t=0}^\infty \gamma^t \nabla_\theta \log \pi_{s_t, a_t}(\theta)} \nonumber \\ 
&\overset{\eqref{eq:GD2*}}{=}& \EE{\tau}{\sum_{t=0}^\infty\gamma^t \left(\left(\cR(s_t, a_t) - \lambda\log\pi_{s_t,a_t}(\theta)\right)\left(\sum_{k=0}^t\nabla_{\theta}\log\pi_{s_k,a_k}(\theta)\right) - \lambda \nabla_\theta \log \pi_{s_t,a_t}(\theta)\right)},
\end{eqnarray}
where the third line is obtained by using the fact that for any $0 \leq t < k$, we have
\begin{align} \label{eq:cross2}
\EE{\tau}{\log \pi_{s_t, a_t}(\theta) \nabla_\theta \log \pi(s_k, a_k)(\theta)} \; = \; 0.
\end{align}
Equation~\eqref{eq:cross2} is derived by following the same proof technique of Lemma~\ref{lem:cross}.

Thus, the stochastic gradient estimator of $\nabla\tilde{J}(\theta)$ with mini-batch size $m$ is
\begin{eqnarray} \label{eq:entropy-nabla-stoch}
\hnabla_m \tilde{J}(\theta) &\eqdef& \hnabla_m J(\theta) - \frac{\lambda}{m}\sum_{i=1}^m\sum_{t=0}^{H-1} \gamma^t \left( \log\pi_{s_t^i, a_t^i}(\theta)\left(\sum_{k=0}^t\nabla_{\theta}\log\pi_{s_k^i,a_k^i}(\theta)\right) + \nabla_\theta \log \pi_{s_t^i,a_t^i}(\theta) \right).
\end{eqnarray}
Notice that $\hnabla_m \tilde{J}(\cdot)$ is the unbiased gradient estimator of the truncated function
\begin{eqnarray} 
\tilde{J}_H(\theta) &\eqdef& \EE{\tau}{\sum_{t=0}^{H-1} \gamma^t \left(\cR(s_t,a_t) - \lambda \log\pi_{s_t,a_t}(\theta)\right)}.
\end{eqnarray}

We show that $\hnabla_m \tilde{J}(\cdot)$ satisfies the~\eqref{eq:ABC} assumption as following.

\begin{lemma} \label{lem:entropy:ABC}
The stochastic gradient estimator~\eqref{eq:entropy-nabla-stoch} satisfies Assumption~\eqref{eq:ABC} with
\begin{eqnarray}
\E{\norm{\hnabla_m \tilde{J}(\theta)}^2} \; \leq \; \left(1-\frac{1}{m}\right)\norm{\nabla \tilde{J}(\theta)}^2 + \frac{2\left(1 - \frac{1}{|\cA|}\right)\cR_{\max}^2}{m(1-\gamma)^3} + \frac{2\lambda^2}{m(1-\gamma^2)}\left(1 - \frac{1}{|\cA|}\right) + \frac{8H|\cA|\lambda^2}{m(1-\gamma)^3}.
\end{eqnarray}
\end{lemma}

\begin{proof}
Let $g(\tau\mid\theta)$ be a stochastic gradient estimator of one single sampled trajectory $\tau$ of $\nabla J_H(\theta)$. Thus $\hnabla_m J(\theta) = \frac{1}{m}\sum_{i=1}^m g(\tau_i\mid\theta)$. Both $\hnabla_m J(\theta)$ and $g(\tau\mid\theta)$ are unbiased estimators of $J_H(\theta)$.

Similarly, let $\tilde{g}(\tau\mid\theta)$ be a stochastic gradient estimator of one single sampled trajectory $\tau$ of $\nabla \tilde{J}_H(\theta)$. Thus $\hnabla_m \tilde{J}(\theta) = \frac{1}{m}\sum_{i=1}^m \tilde{g}(\tau_i\mid\theta)$, and $\hnabla_m \tilde{J}(\theta)$ and $\tilde{g}(\tau\mid\theta)$ are unbiased estimators of $\tilde{J}_H(\theta)$.

Similar to~\eqref{eq:stochgradbnd}, from~\eqref{eq:entropy-nabla-stoch} we have
\begin{align}
\E{\norm{\hnabla_m \tilde{J}(\theta)}^2} &= \E{\norm{\hnabla_m \tilde{J}(\theta) + \nabla \tilde{J}_H(\theta) - \nabla \tilde{J}_H(\theta)}^2} \nonumber \\ 
&= \norm{\nabla \tilde{J}_H(\theta)}^2 + \E{\norm{\hnabla_m \tilde{J}(\theta) - \nabla \tilde{J}_H(\theta)}^2} \nonumber \\ 
&= \norm{\nabla \tilde{J}_H(\theta)}^2 + \E{\norm{\frac{1}{m} \sum_{i=1}^m (\tilde{g}(\tau_i\mid\theta) - \nabla \tilde{J}_H(\theta))}^2} \nonumber \\
&= \norm{\nabla \tilde{J}_H(\theta)}^2 + \frac{1}{m}\E{\norm{\tilde{g}(\tau_1\mid\theta) - \nabla \tilde{J}_H(\theta)}^2} \nonumber \\ 
&= \left(1-\frac{1}{m}\right)\norm{\nabla \tilde{J}(\theta)}^2 + \frac{1}{m}\E{\norm{\tilde{g}(\tau_1\mid\theta)}^2}. \label{eq:entropy-nabla-stoch0}
\end{align}
It remains to show $\EE{\tau}{\norm{\tilde{g}(\tau\mid\theta)}^2}$ is bounded. From~\eqref{eq:entropy-nabla-stoch} we have

\begin{align}
\E{\norm{\tilde{g}(\tau\mid\theta)}^2} &= \EE{\tau}{\norm{g(\tau \mid \theta) - \lambda \sum_{t=0}^{H-1} \gamma^t \log \pi_{s_t,a_t}(\theta)\left(\sum_{k=0}^t\nabla_\theta\log\pi_{s_k,a_k}(\theta)\right) - \lambda\sum_{t=0}^{H-1} \gamma^t\nabla_\theta\log\pi_{s_t,a_t}(\theta)}^2} \nonumber \\ 
&\leq 2\E{\norm{g(\tau \mid \theta)}^2} + 2\lambda^2\E{\norm{\sum_{t=0}^{H-1} \gamma^t \log \pi_{s_t,a_t}(\theta)\left(\sum_{k=0}^t\nabla_\theta\log\pi_{s_k,a_k}(\theta)\right)}^2} \nonumber \\ 
&\quad \ + 2\lambda^2\E{\norm{\sum_{t=0}^{H-1} \gamma^t\nabla_\theta\log\pi_{s_t,a_t}(\theta)}^2} \nonumber \\ 
&\leq \frac{2\left(1 - \frac{1}{|\cA|}\right)\cR_{\max}^2}{(1-\gamma)^3} + 2\lambda^2\underbrace{\E{\norm{\sum_{t=0}^{H-1} \gamma^t \log \pi_{s_t,a_t}(\theta)\left(\sum_{k=0}^t\nabla_\theta\log\pi_{s_k,a_k}(\theta)\right)}^2}}_{\circled{1}} \nonumber \\ 
&\quad \ + 2\lambda^2\underbrace{\E{\norm{\sum_{t=0}^{H-1} \gamma^t\nabla_\theta\log\pi_{s_t,a_t}(\theta)}^2}}_{\circled{2}}, \label{eq:entropy-nabla-stoch2}
\end{align}
where the last inequality is obtained by Lemma~\ref{lem:ABC} with GPOMDP estimator and the constant $G^2 = 1 - \frac{1}{|\cA|}$ provided from Lemma~\ref{lem:softmax_expected}.

Now we will bound $\circled{1}$ and $\circled{2}$ separately.

From Lemma~\ref{lem:Etnorm2}, we know that
\begin{eqnarray}
\circled{2} &=& \sum_{t=0}^{H -1}\gamma^{2t}\E{\norm{\nabla_\theta\log\pi_{s_t,a_t}(\theta)}^2} \nonumber \\ 
&\overset{\mbox{Lemma~\ref{lem:softmax_expected}}}{\leq}& \left(1 - \frac{1}{|\cA|}\right)\sum_{t=0}^{H-1}\gamma^{2t} \nonumber \\ 
&\leq& \frac{1}{1 - \gamma^2}\left(1 - \frac{1}{|\cA|}\right). \label{eq:entropy-nabla-stoch3}
\end{eqnarray}

As for $\circled{1}$, we have
\begin{eqnarray}
\circled{1} &\leq& H\sum_{t=0}^{H -1}\gamma^{2t}\E{\left(\log\pi_{s_t,a_t}(\theta)\right)^2\norm{\sum_{k=0}^t\nabla_\theta\log\pi_{s_k,a_k}(\theta)}^2} \nonumber \\
&\leq& H\sum_{t=0}^{H -1}\gamma^{2t}\E{\left(\log\pi_{s_t,a_t}(\theta)\right)^2\norm{\sum_{k=0}^t\nabla_\theta\log\pi_{s_k,a_k}(\theta)}^2} \nonumber \\
&\leq& H\sum_{t=0}^{H -1}\gamma^{2t}\E{\left(\log\pi_{s_t,a_t}(\theta)\right)^2(t+1)\sum_{k=0}^t\norm{\nabla_\theta\log\pi_{s_k,a_k}(\theta)}^2} \nonumber \\ 
&\overset{\eqref{eq:softmax_LS_nabla_log}}{\leq}& 2H\sum_{t=0}^{H-1}\gamma^{2t}(t+1)^2\E{\left(\log\pi_{s_t,a_t}(\theta)\right)^2} \nonumber \\
&\leq& 2H|\cA|\sum_{t=0}^{H-1} \gamma^{2t}(t+1)^2 \label{eq:entropy-nabla-stoch1} \\ 
&\leq& \frac{4H|\cA|}{(1-\gamma)^3}, \label{eq:entropy-nabla-stoch4}
\end{eqnarray}
where~\eqref{eq:entropy-nabla-stoch1} is obtained by using
\[\E{\left(\log\pi_{s_t,a_t}(\theta)\right)^2} \; = \; \EE{s_t}{\sum_{a \in \cA}\pi_{s_t,a}(\theta)\left(\log\pi_{s_t,a}(\theta)\right)^2} \; \leq \; |\cA|,\] and the last line is obtained by $\gamma^{2t} \leq \gamma^t$ and Lemma~\ref{lem:sum_of_gamma2}.

Combining~\eqref{eq:entropy-nabla-stoch0},~\eqref{eq:entropy-nabla-stoch2},~\eqref{eq:entropy-nabla-stoch3} and~\eqref{eq:entropy-nabla-stoch4} yields the claim of the lemma.
\end{proof}

By adopting Lemma 14 in~\citet{mei2020ontheglobal}, we show that $\tilde{J}(\cdot)$ is smooth as following.

\begin{lemma} \label{lem:entropy:smooth}
$\tilde{J}(\cdot)$ is $\left(\frac{\cR_{\max}}{(1-\gamma)^2}\left(2 - \frac{1}{|\cA|}\right) + \frac{\lambda(4 + 8\log|\cA|)}{(1-\gamma)^3}\right)$-smooth.
\end{lemma}

\begin{proof}
From~\eqref{eq:entropy}, we have
\[\tilde{J}(\theta) \; = \; J(\theta) - \lambda\EE{\tau}{\sum_{t=0}^\infty \gamma^t \log\pi_{s_t,a_t}(\theta)}.\]
From Lemma~\ref{lem:softmax_smoothness}, we know that $J(\cdot)$ is $\left(\frac{\cR_{\max}}{(1-\gamma)^2}\left(2 - \frac{1}{|\cA|}\right)\right)$-smooth.

From Lemma 14 in~\citet{mei2020ontheglobal}, we know that $\EE{\tau}{\sum_{t=0}^\infty \gamma^t \log\pi_{s_t,a_t}(\theta)}$ is $\left(\frac{\lambda(4 + 8\log|\cA|)}{(1-\gamma)^3}\right))$-smooth.

Combining the two smoothness constants yields the claim of the lemma.
\end{proof}

From Lemma~\ref{lem:entropy:ABC} and Lemma~\ref{lem:entropy:smooth} we can also establish a similar FOSP convergence as for Corollary~\ref{cor:sample_complexity}.

\begin{corollary}
%Suppose that Asm.~\ref{ass:lipschitz_smooth_policy} is satisfied.
Consider the vanilla PG updates~\eqref{eq:entropy-nabla-stoch} for the softmax with entropy regularization~\eqref{eq:entropy} .
 For a given $\epsilon>0$, by choosing the mini-batch size $m$ such that $1 \leq m \leq \frac{2\nu}{\epsilon^2}$, the step size $\eta = \frac{\epsilon^2m}{2L\nu}$, the horizon $H = \cO\left((1-\gamma)^{-1}\log\left(1/\epsilon\right)\right)$ and the number of iterations $T$ such that
\begin{align} 
    Tm \geq \frac{8\delta_0L\nu}{\epsilon^4} = \cO((1-\gamma)^{-6}\epsilon^{-4})
\end{align}
with 
\[L = \left(\frac{\cR_{\max}}{(1-\gamma)^2}\left(2 - \frac{1}{|\cA|}\right) + \frac{\lambda(4 + 8\log|\cA|)}{(1-\gamma)^3}\right)\]
and
\[\nu = \frac{2\left(1 - \frac{1}{|\cA|}\right)\cR_{\max}^2}{(1-\gamma)^3} + \frac{2\lambda^2}{(1-\gamma^2)}\left(1 - \frac{1}{|\cA|}\right) + \frac{8H|\cA|\lambda^2}{(1-\gamma)^3},\]
then $\E{\norm{\nabla \tilde{J}(\theta_U)}^2} = \cO(\epsilon^2)$. 
\end{corollary}

\paragraph*{Remark.}
The sample complexity $Tm \times H$ is $\cO((1-\gamma)^{-8}\epsilon^{-4})$ instead of $\cO((1-\gamma)^{-6}\epsilon^{-4})$ as in Corollary~\ref{cor:sample_complexity} due to the $(1-\gamma)^{-3}$ dependency on the smoothness constant $L$ and the $(1-\gamma)^{-4}$ dependency on the bounded variance constant $\nu$.

\begin{proof}
From Lemma~\ref{lem:entropy:smooth}, we know that \[L = \left(\frac{\cR_{\max}}{(1-\gamma)^2}\left(2 - \frac{1}{|\cA|}\right) + \frac{\lambda(4 + 8\log|\cA|)}{(1-\gamma)^3}\right).\]

From Lemma~\ref{lem:entropy:ABC}, we know that
\[
\nu = \frac{2\left(1 - \frac{1}{|\cA|}\right)\cR_{\max}^2}{(1-\gamma)^3} + \frac{2\lambda^2}{(1-\gamma^2)}\left(1 - \frac{1}{|\cA|}\right) + \frac{8H|\cA|\lambda^2}{(1-\gamma)^3}.
\]
Plugging in $L$ and $\nu$ in Corollary~\ref{cor:sample_complexity} yields the corollary's claim.
\end{proof}

% !TEX root = summary.tex

\section{Global optimum convergence under the gradient domination assumption}
\label{sec:PL}

As~\citet{fazel2018global,mei2020ontheglobal} did for the exact policy gradient update, relying on the following gradient domination assumption, we establish a global optimum convergence guarantee and the sample complexity analysis for the stochastic vanilla PG.

\begin{assumption}[Gradient domination] \label{ass:PL}
 We say that a differentiable function $J$ satisfies the gradient domination condition if for all $\theta\in\R^d$, there exists $\mu > 0$ such that
 \begin{align} \label{eq:PL}
 \frac{1}{2}\norm{\nabla J_H(\theta)}^2 \; \geq \; \mu\left(J^* - J(\theta)\right). \tag{PL}
 \end{align}
\end{assumption}

The gradient domination condition is also known as the Polyak-Lojasiewicz (PL) condition. The PL condition was originally discovered independently in the seminal works of B. Polyak and S. Łojasiewicz \citep{polyak1963gradient,lojasiewicz1963une,lojasiewicz1959probleme}. Equipped with this additional assumption, we can adapt Theorem 3 in~\citet{khaled2020better} and obtain the following global optimum convergence guarantee.

\begin{theorem} \label{pro:PL}
Suppose that Assumptions~\ref{ass:smooth},~\ref{ass:trunc},~\ref{ass:ABC} and~\ref{ass:PL} hold. Suppose that PG defined in~\eqref{eq:GA} (Alg.~\ref{alg:pg}) is run for $T>0$ iterations with step size $(\eta_t)_t$ chosen as
\begin{eqnarray}
\eta_t &=&
\begin{cases}
\frac{1}{b} \quad & \mbox{if } T \leq \frac{b}{\mu} \; \mbox{ or } \; t \leq t_0 \\
\frac{2}{2b + \mu(t-t_0)} \quad & \mbox{if } T \geq \frac{b}{\mu} \; \mbox{ and } \; t > t_0
\end{cases}
\end{eqnarray}
with $t_0 = \left[\frac{T}{2}\right]$ and $b = \max\{2AL/\mu, 2BL, \mu\}$. Then
\begin{eqnarray} \label{eq:PL_bound}
J^* - \E{J(\theta_T)} \leq 16\exp\left(-\frac{\mu (T - 1)}{2\max\{\frac{2AL}{\mu}, 2BL, \mu\}}\right)\left(J^* - J(\theta_0)\right) + \frac{12LC}{\mu^2T} + \frac{26D\gamma^H}{\mu}.
\end{eqnarray}
\end{theorem}
%\rui{A proposition for the convergence rate and the sample complexity of the average regret with the global optimum.}

\paragraph*{Remark.} Notice that for the exact full gradient update, we have Assumption~\ref{ass:trunc} and~\ref{ass:ABC} hold with $A=C=D=0$ and $B=1$. Thus under the smoothness assumption and the~\eqref{eq:PL} condition , we establish a linear convergence rate for the number of iterations to the global optimal. We recover the linear convergence rate for the softmax with entropy regularization in Theorem 6 in~\citet{mei2020ontheglobal} where the smoothness assumption holds and the~\eqref{eq:PL} condition holds under the path of the iterations in the exact case.

As for the stochastic vanilla PG, the dominant term in~\eqref{eq:PL_bound} is $\frac{12LC}{\mu^2T}$. This implies that the sample complexity is $T \times H = \widetilde{\cO}(\epsilon^{-1})$ with $T = \cO(\epsilon^{-1})$ and $H = \log \epsilon^{-1}$.

\begin{proof}
Using the $L$-smoothness of $J$ from Assumption~\ref{ass:smooth},
\begin{eqnarray}
J^* - J(\theta_{t+1}) &\leq& J^* - J(\theta_t) - \dotprod{\nabla J(\theta_t), \theta_{t+1}-\theta_t} + \frac{L}{2}\norm{\theta_{t+1}-\theta_t}^2 \nonumber \\ 
&=& J^* - J(\theta_t) - \eta_t\dotprod{\nabla J(\theta_t), \hnabla_m J(\theta_t)} + \frac{L\eta_t^2}{2}\norm{\hnabla_m J(\theta_t)}^2.
\end{eqnarray}
Taking expectation conditioned on $\theta_t$ and using Assumption~\ref{ass:ABC} and~\ref{ass:PL},
\begin{eqnarray}
\EE{t}{J^* - J(\theta_{t+1})} &\leq& J^* - J(\theta_t) - \eta_t\dotprod{\nabla J(\theta_t), \nabla J_H(\theta_t)} + \frac{L\eta_t^2}{2}\EE{t}{\norm{\hnabla_m J(\theta_t)}^2} \nonumber \\ 
&\overset{\eqref{eq:ABC}}{\leq}& J^* - J(\theta_t) - \eta_t\dotprod{\nabla J_H(\theta_t) + (\nabla J(\theta_t) - \nabla J_H(\theta_t)), \nabla J_H(\theta_t)} + \nonumber \\ 
&\quad& \ + \frac{L\eta_t^2}{2}\left(2A(J^* - J(\theta_t)) + B\norm{\nabla J_H(\theta_t)}^2 +C\right) \nonumber \\ 
&=& (1+L\eta_t^2A)(J^* - J(\theta_t)) - \eta_t\left(1-\frac{LB\eta_t}{2}\right)\norm{\nabla J_H(\theta_t)}^2 + \frac{L\eta_t^2C}{2} \nonumber \\ 
&\quad& \ - \eta_t\dotprod{\nabla J(\theta_t) - \nabla J_H(\theta_t), \nabla J_H(\theta_t)} \nonumber \\ 
&\overset{\eqref{eq:PL}}{\leq}& \left(1 - 2\eta_t\mu\left(1-\frac{LB\eta_t}{2}\right) + L\eta_t^2A\right)(J^* - J(\theta_t)) + \frac{L\eta_t^2C}{2} \nonumber \\ 
&\quad& \ - \eta_t\dotprod{\nabla J(\theta_t) - \nabla J_H(\theta_t), \nabla J_H(\theta_t)} \nonumber \\ 
&\leq& \left(1-\frac{3\eta_t\mu}{2}+L\eta_t^2A\right)(J^* - J(\theta_t)) + \frac{L\eta_t^2C}{2} \nonumber \\ 
&\quad& \ - \eta_t\dotprod{\nabla J(\theta_t) - \nabla J_H(\theta_t), \nabla J_H(\theta_t)} \label{eq:3/2} \\ 
&\overset{\eqref{eq:trunc}}{\leq}& \left(1-\frac{3\eta_t\mu}{2}+L\eta_t^2A\right)(J^* - J(\theta_t)) + \frac{L\eta_t^2C}{2} + \eta_tD\gamma^H \nonumber \\ 
&\leq& (1-\eta_t\mu)(J^* - J(\theta_t)) + \frac{L\eta_t^2C}{2} + \eta_tD\gamma^H, \label{eq:eta_mu}
\end{eqnarray}
where~\eqref{eq:3/2} is obtained by the inequality $1 - \frac{LB\eta_t}{2} \geq \frac{3}{4}$, and~\eqref{eq:eta_mu} is obtained by the inequality $L\eta_t A \leq \frac{\mu}{2}$, due to the choice of step size $\eta_t \leq \frac{1}{b}$ for all $t \geq 0$ with $b \geq 2BL, 2AL/\mu$, respectively. Here, $1 - \eta_t\mu \geq 0$ as $\eta_t \leq \frac{1}{b}$ and $b \geq \mu$.

Taking total expectation and letting $r_t \eqdef \E{J^* - J(\theta_t)}$ on~\eqref{eq:eta_mu}, we have
\begin{eqnarray} \label{eq:recurse}
r_{t+1} &\leq& (1-\eta_t\mu)r_t + \frac{L\eta_t^2C}{2} + \eta_tD\gamma^H.
\end{eqnarray}

If $T \leq \frac{b}{\mu}$, we have $\eta_t = \frac{1}{b}$. Recursing the above inequality, we get
\begin{eqnarray}
r_T &\leq& \left(1-\frac{\mu}{b}\right)r_{T-1} + \frac{LC}{2b^2} + \frac{D\gamma^H}{b} \nonumber \\ 
&\overset{\eqref{eq:recurse}}{\leq}& \left(1-\frac{\mu}{b}\right)^Tr_0 + \left(\frac{LC}{2b^2} + \frac{D\gamma^H}{b}\right)\sum_{i=0}^{T-1} \left(1-\frac{\mu}{b}\right)^i \nonumber \\ 
&\leq& \exp\left(-\frac{\mu T}{b}\right)r_0 + \frac{LC}{2\mu b} + \frac{D\gamma^H}{\mu} \label{eq:T<} \\ 
&\overset{T \leq \frac{b}{\mu}}{\leq}& \exp\left(-\frac{\mu T}{b}\right)r_0 + \frac{LC}{2\mu^2 T} + \frac{D\gamma^H}{\mu}. \label{eq:1/T}
\end{eqnarray}

If $T \geq \frac{b}{\mu}$, as $\eta_t = \frac{1}{b}$ when $t \leq t_0$, from~\eqref{eq:T<}, we have
\begin{eqnarray}
r_{t_0} &\leq& \exp\left(-\frac{\mu t_0}{b}\right)r_0 + \frac{LC}{2\mu b} + \frac{D\gamma^H}{\mu} \nonumber \\ 
&\leq& \exp\left(-\frac{\mu (T-1)}{2b}\right)r_0 + \frac{LC}{2\mu b} + \frac{D\gamma^H}{\mu}, \label{eq:t0}
\end{eqnarray}
where the last line is obtained by $t_0 = \left[\frac{T}{2}\right] \geq \frac{T-1}{2}$.

For $t > t_0$,
\[\eta_t = \frac{2}{\mu\left(\frac{2b}{\mu}+t-t_0\right)}.\] 
From~\eqref{eq:recurse}, we have
\begin{eqnarray}
r_t &\leq& (1-\eta_t\mu)r_{t-1} + \frac{L\eta_t^2C}{2} + \eta_tD\gamma^H \nonumber \\ 
&=& \frac{\frac{2b}{\mu}+t-t_0 - 2}{\frac{2b}{\mu}+t-t_0} r_{t-1} + \frac{2LC}{\mu^2\left(\frac{2b}{\mu}+t-t_0\right)^2} + \frac{2D\gamma^H}{\mu \left(\frac{2b}{\mu}+t-t_0\right)}.
\end{eqnarray}
Multiplying both sides by $\left(\frac{2b}{\mu}+t-t_0\right)^2$, we have
\begin{align}
\left(\frac{2b}{\mu}+t-t_0\right)^2r_t &\leq \left(\frac{2b}{\mu}+t-t_0\right)\left(\frac{2b}{\mu}+t-t_0-2\right)r_{t-1} + \frac{2LC}{\mu^2} + \frac{2D\gamma^H}{\mu}\left(\frac{2b}{\mu}+t-t_0\right) \nonumber \\ 
&\leq \left(\frac{2b}{\mu}+t-t_0-1\right)^2 r_{t-1} + \frac{2LC}{\mu^2} + \frac{2D\gamma^H}{\mu}\left(\frac{2b}{\mu}+t-t_0\right).
\end{align}
Let $w_t \eqdef \left(\frac{2b}{\mu}+t-t_0\right)^2$. Then,
\begin{eqnarray}
w_tr_t &\leq& w_{t-1}r_{t-1} + \frac{2LC}{\mu^2} + \frac{2D\gamma^H}{\mu}\left(\frac{2b}{\mu}+t-t_0\right).
\end{eqnarray}
Summing up for $t = t_0+1, \cdots, T$ and telescoping, we get,
\begin{align}
w_Tr_T &\leq w_{t_0}r_{t_0} + \frac{2LC(T-t_0)}{\mu^2} + \frac{2D\gamma^H}{\mu}\sum_{t=t_0+1}^T\left(\frac{2b}{\mu}+t-t_0\right) \nonumber \\ 
&= \frac{4b^2}{\mu^2}r_{t_0} + \frac{2LC(T-t_0)}{\mu^2} + \frac{4bD(T-t_0)\gamma^H}{\mu^2} + \frac{D\gamma^H}{\mu}(T-t_0)(T-t_0+1).
\end{align}
Dividing both sides by $w_T$ and using that since
\[w_T \; = \; \left(\frac{2b}{\mu}+T-t_0\right)^2 \; \geq \; (T-t_0)^2,\]
we have
\begin{eqnarray}
r_T &\leq& \frac{4b^2}{\mu^2w_T}r_{t_0} + \frac{2LC(T-t_0)}{\mu^2w_T} + \frac{4bD(T-t_0)\gamma^H}{\mu^2w_T} + \frac{D\gamma^H}{\mu w_T}(T-t_0)(T-t_0+1) \nonumber \\ 
&\leq& \frac{4b^2}{\mu^2(T-t_0)^2}r_{t_0} + \frac{2LC}{\mu^2(T-t_0)} + \frac{4bD\gamma^H}{\mu^2(T-t_0)} + \frac{2D\gamma^H}{\mu}.
\end{eqnarray}
By the definition of $t_0$, we have $T-t_0 \geq \frac{T}{2}$. Plugging this estimate, we have
\begin{eqnarray}
r_T &\leq& \frac{16b^2}{\mu^2T^2}r_{t_0} + \frac{4LC+8bD\gamma^H}{\mu^2T} + \frac{2D\gamma^H}{\mu} \nonumber \\ 
&\overset{T\geq\frac{b}{\mu}}{\leq}& \frac{16b^2}{\mu^2T^2}r_{t_0} + \frac{4LC}{\mu^2T} + \frac{10D\gamma^H}{\mu} \nonumber \\ 
&\overset{\eqref{eq:t0}}{\leq}& \frac{16b^2}{\mu^2T^2}\left(\exp\left(-\frac{\mu (T-1)}{2b}\right)r_0 + \frac{LC}{2\mu b} + \frac{D\gamma^H}{\mu}\right) + \frac{4LC}{\mu^2T} + \frac{10D\gamma^H}{\mu} \nonumber \\ 
&\overset{T\geq\frac{b}{\mu}}{\leq}& 16\exp\left(-\frac{\mu (T-1)}{2b}\right)r_0 + \frac{8LC}{\mu^2 T} + \frac{16D\gamma^H}{\mu} + \frac{4LC}{\mu^2T} + \frac{10D\gamma^H}{\mu} \nonumber \\ 
&=& 16\exp\left(-\frac{\mu (T-1)}{2b}\right)r_0 + \frac{12LC}{\mu^2 T} + \frac{26D\gamma^H}{\mu}. \label{eq:>t0}
\end{eqnarray}
It remains to take the maximum of the two bounds~\eqref{eq:1/T} and~\eqref{eq:>t0} with $b = \max\{2AL/\mu, 2BL, \mu\}$.
\end{proof}

%\end{appendices}

\end{document}